\documentclass{article}

\usepackage{amstext,pict2e,amssymb,latexsym}
\usepackage{graphicx,tabularx} 
\usepackage{amsmath,colonequals,nicefrac}  
\usepackage{array,color} 

\usepackage{fancyhdr}           
\usepackage{ifthen}
\usepackage[T1]{fontenc}
\fontencoding{T1}\selectfont

\newcommand{\etal}{{\em et al.}}
\newcommand\settitle[2][]{%
 \title{#2}
 \ifthenelse{\equal{#1}{}}%
  {\fancyhead[RO]{\nouppercase #2 \qquad \thepage}}%
  {\fancyhead[RO]{\nouppercase #1 \qquad \thepage}}%
}

\newcommand{\eqr}[1]{\w{\upshape (#1)}}
\newcommand{\cp}[1]{\ws{$\lambda_{CP}(#1)$}}
\newcommand{\rest}{\mathord{\upharpoonright}}
\newcommand{\net}[1]{\w{${\cal #1}$}}
\newcommand{\att}[1]{\rightarrow}
\newcommand{\tuple}[1]{\w{$\langle #1 \rangle$}}
\newcommand{\eq}[1]{\w{\boldmath$#1$\unboldmath}}
\newcommand{\eqmax}{\w{$Eq_{\w{max}}$}}

\newcommand{\eqinv}{\w{$Eq_{\w{inv}}$}}

\newcommand{\Sol}{\w{\boldmath$f$}}
\newcommand{\fg}[1]{\w{$V(#1)$}}
\newcommand{\fsol}{\w{$V$}}
\newcommand{\f}[1]{\w{$V_e(#1)$}}
\newcommand{\feq}{\w{$V_e$}}
\newcommand{\fzero}{\w{$V_0$}}
\newcommand{\fgt}[2]{\w{$V^{#2}(#1)$}}
\newcommand{\imp}{\w{$\rightarrow$}}
\newcommand{\inc}{\w{{\bf in}}}
\newcommand{\exc}{\w{{\bf out}}}
\newcommand{\und}{\w{{\bf und}}}

\newcommand{\bN}{\w{\boldmath$N$}}
\newcommand{\bP}{\w{\boldmath$P$}}
\newcommand{\bQ}{\w{\boldmath$Q$}}

\newcommand{\insertnumequation}[2]{
\medskip

\noindent\parbox[t]{3cm}{(#2)}#1

\medskip
}

\def\keywordsname{Keywords.}

\newtheorem{definition}{Definition}[section]
\newtheorem{theorem}{Theorem}[section]
\newtheorem{remark}{Remark}[section]
\newtheorem{corollary}{Corollary}[section]
\newtheorem{proposition}{Proposition}[section]
\newtheorem{lemma}{Lemma}[section]
\newtheorem{example}{Example}[section]
\newenvironment{proof}{
\noindent{\bf Proof.}
}

\let\oldmarginpar\marginpar
\renewcommand\marginpar[1]{\oldmarginpar[\raggedleft\footnotesize #1]%
{\raggedright\footnotesize #1}}

\newcommand{\w}[1]{\text{#1}}
\newcommand{\ws}[1]{~\text{#1}~}
\newcommand{\wR}[1]{\text{#1}~}

\ifx\text\undefined
  \newcommand{\text}[1]{\relax
    \ifmmode\mathchoice
      {\hbox{\the\textfont0\relax#1}}%
      {\hbox{\the\textfont0\relax#1}}%
      {\hbox{\the\scriptfont0\relax#1}}%
      {\hbox{\the\scriptscriptfont0\relax#1}}%
    \else{\relax#1}\fi}
  \fi

\ifx\text\undefined
  \newcommand{\text}[1]{\relax
    \ifmmode\mathchoice
      {\hbox{\the\textfont0\relax#1}}%
      {\hbox{\the\textfont0\relax#1}}%
      {\hbox{\the\scriptfont0\relax#1}}%
      {\hbox{\the\scriptscriptfont0\relax#1}}%
    \else{\relax#1}\fi}
  \fi

\newcommand{\hf}{\w{$\nicefrac{1}{2}$}}
\newcommand{\df}{argu\-men\-tation-friendly}
\renewcommand{\c}[1]{\w{${\cal #1}$}}

\begin{document}
\title{Equilibrium States in Numerical Argumentation Networks}
\date{}

\author{D. Gabbay\\
King's College London,\\ Department of Informatics,\\ 
The Strand,\\ London, WC2R 2LS, UK\\
{\tt dov.gabbay@kcl.ac.uk}
\and
O. Rodrigues\\
King's College London,\\ Department of Informatics,\\ 
The Strand,\\ London, WC2R 2LS, UK\\
{\tt odinaldo.rodrigues@kcl.ac.uk}\\
}
\maketitle

\begin{abstract}
Given an argumentation network with initial values to the 
arguments, we look for algorithms which can yield extensions compatible 
with such initial values. We find that the best way of tackling this problem is
to offer an iteration formula that takes the initial values and
the attack relation and iterates a sequence of intermediate values that
eventually converges leading to an extension. The properties 
surrounding the application of the iteration formula and its connection 
with other numerical and non-numerical techniques proposed by others are 
thoroughly investigated in  this paper.
\end{abstract}

\maketitle    


\section{Orientation and Background\label{sec:intro}}

\subsection{Orientation}
A finite  system \tuple{S,R}, with $R$ a binary relation on $S$, can be 
viewed in many different ways; among them are
\begin{enumerate}
\item As an abstract argumentation framework \cite{dung-aaf}, and 
\item As a generator of equations \cite{dov-eq-short:2011,dov-eq:12}
\end{enumerate}

When viewed as an abstract argumentation framework, the basic concepts studied are those of {\em extensions} (being certain subsets of $S$)  and different {\em semantics} (being sets of extensions). When studied as generators of equations, one can generate equations in such a way that the solutions $\Sol$ to the equations correspond to (complete) extensions and sets of such solutions correspond to semantics.

This paper offers an iteration schema for finding specific solutions to the equations responding to initial requirements and shows what these solutions correspond to in the abstract argumentation sense.

We now explain the role iteration formulas play in general in the equational context.

When we have a system of equations designed to model an application 
area\footnote{For example, equations of fluid flow in hydrodynamics or equations of particle motion in mechanics, or equations modelling argumentation networks according
to the equational approach (to be explained later), or equations modelling a 
biological system of predator-prey  ecology, or some polynomial equation 
arising in macroeconomics.} we face two problems: 1) find any solution to 
the system of equations, which will have a meaning in the application 
area giving rise to the equations; 2) given boundary conditions and/or 
other requirements not necessarily mathematical which are meaningful in 
the application area,\footnote{For example, initial conditions in the case of 
particle mechanics, or initial size of population in the ecology, or 
arguments that we would like to be accepted.} we would like to find a solution 
to the system of equations that is compatible/respects the initial 
conditions/requirements.

These two problems are distinct. The first one of finding any solution 
is a numerical analysis problem. There are various iteration methods in
numerical analysis to find solutions, of which one of the most known is 
Newton's method.\footnote{This method starts with an initial guess of a
possible solution and uses various iteration formulae hoping that it will
converge to a solution (for an introduction on numerical analysis see
\cite{Suli-Mayers:2003}).}
The second problem is totally different. It calls for an understanding
of the requirements coming from the application area and possibly the
design of a specialised iteration formula which respects the type of
requirements involved. 

This paper provides the {\em Gabbay-Rodrigues Iteration Schema}, for the
case of the equational approach to argumentation, seeking solutions (which
we shall see will correspond to complete extensions) respecting as much as
possible initial demands and restrictions of what arguments are in
or out of the extension. We compare what our iteration schema does with 
Caminada and Pigozzi's down-admissible and up-complete constructions 
\cite{Caminada-Pigozzi:11}.
Because we are dealing with iteration formulas (involving limits) and we are comparing with set theoretical operations (as in Caminada and Pigozzi's paper) we have to be detailed and precise and despite it being conceptually clear and simple, the proofs 
turn out to be mathematically involved, and require some patience from our readers.  However, once we establish the properties of our iteration schema, its use and 
application are straightforward and computationally simple, especially in the 
context of such tools as MATHEMATICA and others like it. The reader may wish to 
just glance at the technical proofs and concentrate on the examples and discussions.
Note the iteration schema idea is very general and applies to other systems of equations possibly using other iteration formulas.

The actual technical development of the paper will start in
Section~\ref{sec:initial-values}. In Appendix~\ref{app:examples}
we emphasise the distinction between the above two problems with two 
detailed examples, the first modelling the dynamics of predator-prey 
interactions and the second about merging/voting in argumentation networks. 
We shall see that Newton's method does not work in these 
scenarios and that there is the need for a new type of iteration schema. Thus 
this paper is not just incremental to the equational approach but constitutes 
a serious and necessary conceptual extension. 


\subsection{Background\label{sec:background}}
 
\begin{figure}[htb]
\begin{center}
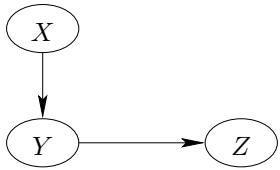\\
\end{center}
\caption{A sample argumentation network.
         \label{sample-network}}
\end{figure}

An abstract argumentation framework is a formalism proposed by Dung \cite{dung-aaf} 
and defined in terms of a tuple \tuple{S,R}, where $S$ is a non-empty set of arguments 
and $R \subseteq S \times S$ is a binary {\em attack relation}.  We will refer to an 
abstract argumentation framework \tuple{S,R} simply as an {\em argumentation network}. 
If $(X,Y) \in R$, we say that the argument $X$ attacks the argument $Y$. 
\tuple{S,R} can be seen as a directed graph (see 
Figure~\ref{sample-network}). As informally introduced in 
Section~\ref{sec:intro}, $Att(X)$ will be used to denote the 
set $\{Y \in S\; | \; (Y,X) \in R\}$, i.e., the set of arguments attacking 
the argument $X$. 
Following graph theory convention, if $X$ has no attackers (i.e., 
$Att(X)=\varnothing$), we say that $X$ is a {\em source node} in 
\tuple{S,R}. Given a set $E\subseteq S$, we write $E \rightarrow X$ 
as a shorthand for $\exists Y \in E$, such that $(Y,X) \in R$. 
Furthermore, following \cite{Caminada:07}, we use $E^+$ to denote 
the set $\{ Y \in S \; | \; E \rightarrow Y\}$.

Given an argumentation network, one usually wants to reason about the 
{\em status} of its arguments, i.e., whether an argument persists or is defeated
by other arguments. It should be clear that arguments that have no attacks 
on them always persist. However, an attack from $X$ to $Y$ may not in itself 
be sufficient to defeat $Y$, because $X$ may be defeated by some 
argument that attacks it, and thus one needs an evaluation process to determine 
the status of all arguments systematically. In Dung's original formulation, 
this was done through an {\em acceptability semantics} defining conditions 
for the acceptability of an argument. The semantics can be defined in terms
of {\em extensions} --- subsets of $S$ with special properties. These
subsets are based on two fundamental notions which are explained next.

A set $E \subseteq S$ is said to be {\em conflict-free} if for all 
elements $X,Y \in E$, we have that $(X,Y) \not \in R$. Intuitively, 
arguments of a conflict-free set do not attack each other. However, this 
does not necessarily mean that all arguments in the set are properly 
supported. Well supported sets satisfy special {\em admissibility} criteria. 
We say that an argument $X\in S$ is {\em acceptable with respect to $E\subseteq 
S$}, if for all $Y \in S$, such that $(Y,X) \in R$, there is an element $Z\in E$, 
such that $(Z,Y) \in R$. A set $E\subseteq S$ is {\em admissible} if it 
is conflict-free and all of its elements are acceptable with respect 
to itself. An admissible set $E$ is a {\em complete extension} if and 
only if $E$ contains all arguments which are acceptable with respect
to itself.
$E$ is called a {\em preferred extension} of $S$, if and only if
$E$ is maximal with respect to set inclusion amongst all complete
extensions of $S$. Similarly, $E$ is called a {\em stable extension} of
$S$ if and only if $E$ is conflict-free and for every $X \in S\backslash E$,
there is an element $Y \in E$, such that $(Y,X) \in R$.

\begin{figure}[hbt]

\medskip

\begin{center}
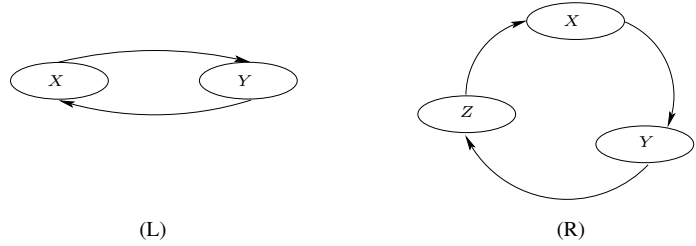
\end{center}

\medskip

\vspace*{-5mm}
\caption{Sample argumentation networks.\label{figure_example}}
\end{figure}

Now consider the argumentation networks (L) and (R) depicted in 
Figure~\ref{figure_example}. According to the semantics given above, 
the network (L) has three extensions $E_0=\varnothing$, 
$E_1=\{X\}$ and $E_2=\{Y\}$. Both $E_1$ and $E_2$ are preferred and 
stable extensions. The network (R) only has only one extension, 
which is empty, and hence this is also its only preferred extension. 
This extension is however not stable.

Besides Dung's acceptability semantics, it is also possible to give 
meaning to these networks through Caminada's {\em labelling semantics} 
\cite{caminada-gabbay:2009,caminada:2011} and through Gabbay's {\em 
equational approach} \cite{dov-eq-short:2011,dov-eq:12}. These are 
explained next.

\medskip
\noindent{\em The labelling semantics.}

  The labelling semantics uses labelling functions $\lambda:S 
  \longrightarrow \{\inc,\exc,\und\}$ satisfying certain conditions
  tailored so as to obtain a complete correspondence 
  with Dung's semantics. 

  The labelling of an argument in disagreement with Dung's semantics is 
  said to be ``illegal''. This is explained further as follows.

\begin{definition}[Illegal labelling of an argument \cite{Caminada-Pigozzi:11}]
Let \tuple{S,R} be an argumentation network and $\lambda$ a labelling
function for $S$.
\label{def:illegal-label}
\begin{enumerate}
\item An argument $X \in S$ is illegally labelled \inc\ by $\lambda$ if 
$\lambda(X) =\inc$ and there exists $Y \in Att(X)$ such that $\lambda(Y)
\neq \exc$.
\item An argument $X \in S$ is illegally labelled \exc\ by $\lambda$ if 
$\lambda(X)=\exc$ and there is no $Y \in Att(X)$ such that $\lambda(Y)= \inc$.
\item An argument $X \in S$ is illegally labelled \und\ by $\lambda$ if 
$\lambda(X)=\und$ and either for all $Y \in Att(X)$, $\lambda(Y)= out$ or
there exists $Y \in Att(X)$, such that $\lambda(Y)=\inc$.
\end{enumerate}
\end{definition} 

A legal (complete) labelling is a labelling in which no argument is illegally 
labelled.

It is possible to have more than one legal labelling function for the
same argumentation network. Each labelling function will correspond to
an extension in Dung's semantics.
For   example, for network  (L), we have the three 
  functions $\lambda_1$, $\lambda_2$ and $\lambda_0$ below.

\medskip

\begin{tabularx}{11cm}{X|X|X}
$\lambda_1 \Leftrightarrow E_1=\{X\}$ & $\lambda_2 \Leftrightarrow E_2=
\{Y\}$ & $\lambda_0 \Leftrightarrow E_0=\varnothing$ \\ \hline 
$\lambda_1(X)=$ \inc & $\lambda_2(X)=$ \exc & $\lambda_0(X)=$ \und \\
$\lambda_1(Y)=$ \exc & $\lambda_2(Y)=$ \inc & $\lambda_0(Y)=$ \und \\
\end{tabularx}

\medskip

For the network (R), we have only the function $\lambda$ such that 
$\lambda(X)=\lambda(Y)=\lambda(Z)=\und$. This gives the empty extension.

\medskip

\noindent{\em The equational approach.}

  The equational approach views an 
  argumentation network $\tuple{S,R}$ as a mathematical graph generating 
  equations for
  functions in the unit interval $U=[0,1]$.  Any solution $\Sol$ to
  these equations conceptually corresponds to an extension. Of course,
  the end result depends on how the equations are generated and we can
  get different solutions for different equations. Once the equations
  are fixed, the totality of the solutions to the system of equations is 
  viewed as the totality of extensions via an appropriate mapping. One 
  equation schema we can possibly use for generating equations is the 
  \eqmax\ below, where \fg{X} is the value of a node $X\in S$:
  \insertnumequation{$\textstyle\fg{X}=1-\max_{Y_i \in Att(X)}\{\fg{Y_i}\}$}%
                    {\eqmax}

  Another possibility is \eqinv:
  \insertnumequation{$\textstyle\fg{X}=\prod_{Y_i\in Att(X)}(1-\fg{Y_i})$}{\eqinv}

It is easy to see that according to \eqmax\ the value of any source 
argument will be $1$ (since they have no attackers) and the value of any 
argument with an attacker with value $1$ will be $0$. The situation is more 
complex with nodes participating in cycles. Consider the network (L) again, 
with equations
\[
\begin{array}{c}
\fg{X}=1-\fg{Y}\\
\fg{Y}=1-\fg{X}
\end{array}
\]
If values are taken from the unit interval, this system of equations 
will accept any solution $\fsol$ such that $\fg{X}+\fg{Y}=1$. We can divide 
these solutions between three classes: $\fgt{X}{1}=1$, $\fgt{Y}{1}=0$; 
$\fgt{X}{2}=0$, $\fgt{Y}{2}=1$ and $0<\fgt{X}{0} <1$, $0<\fgt{Y}{0} <1$ 
with $\fgt{X}{0}+\fgt{Y}{0}=1$. These again correspond to the three 
extensions $E_1$, $E_2$ and $E_0$ given before.

In fact, Gabbay has shown that in the case of $Eq_{\w{max}}$ the totality 
of solutions to the system of equations corresponds to the totality of 
extensions in Dung's sense \cite{dov-eq:12}. The correspondence is best 
explained in terms of the labelling semantics, using the following 
correspondence:
\begin{center}
\begin{tabular}{lcl}
$\fg{X}=1$ & $\coloncolon$ & $\lambda(X)=$ \inc \\
$\fg{X}=0$ & $\coloncolon$ & $\lambda(X)=$ \exc \\
$0 < \fg{X} < 1$ & $\coloncolon$ & $\lambda(X)=$ \und \\
\end{tabular}
\end{center}

The advantage of the equational approach is that it allows us to think
of an argumentation network as a numeric system in which nodes are
given certain values depending on specific rules governing their 
interaction with their neighbours. A rule may for instance require 
the value of a node to be $0$ if the value of any attacking node is 
$1$. Another rule may force the value of a node to be $1$ if it has
no attacking nodes. The schema $Eq_{\w{max}}$ and $Eq_{\w{inv}}$ embed these 
rules, and they agree with Dung's semantics.
A solution to the system of equations is any combination of values of 
nodes satisfying the equations. Of course, since the node values are 
no longer discrete we have more freedom to design rules which are 
appropriate for a given application. Part of the objective of this 
paper is to explore the nature of these rules.

\begin{figure}[hbt]

\medskip

\begin{center}
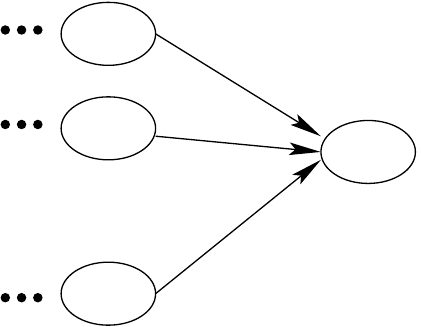
\end{center}

\medskip

\vspace*{-5mm}
\caption{Multiple attacks on a node.\label{fig:multiple-attacks}}
\end{figure}

We start by generalising some concepts a bit further.
Consider the network in Figure~\ref{fig:multiple-attacks} in which
$Att(X)=\{ Y_1,Y_2,\ldots,Y_k\}$.  To agree with Dung's semantics, if the 
value of any attacker of $X$ is $1$, we want the value of $X$ to be $0$. 
If all of the attackers of $X$ have value $0$, we want the value of $X$ 
to be $1$. For any other combination of values of the attackers we want 
the value of $X$ to be anything other than $0$ or $1$. So within the 
traditional semantics but taking the extended set of values of the 
unit interval, we can think of a single attack by a node with value
$v$ as the order-reversing operation which returns the value $1-v$.
This is a kind of {\em negation}.\footnote{If we make \und\ equals 
$\frac{1}{2}$, then an attack by a single undecided node will have 
value $\frac{1}{2}$.}
Since a node can have multiple attacks, we also need an operation
to combine the values of the attackers. We can think of this as a type
of {\em conjunction}, which numerically can be obtained through several 
operations. For instance, in fuzzy logic, the standard
semantics of (weak) conjunction is given by the operation $\min$.

Therefore, the value of a node $X$ can be defined as 
\[\fg{X}=\textstyle\min\limits_{Y \in Att(X)}
\{1-\fg{Y}\}\]
which is equivalent to 
\[\fg{X}=1-\max\limits_{Y \in Att(X)}\{\fg{Y}\}\] 
obtained by our now familiar schema \eqmax. Note that the conjunction 
operation in the schema \eqinv\ is {\em product}. The operations $\min$ 
and product are two examples of t-norms. They are two instances of functions 
that are particularly suitable for argumentation semantics. The following
definition elaborates on this further.

\begin{definition}\label{def:arg-friendly} A function $g$ with domain being 
the family of all finite sequences of elements from $U$ and range $U$
is {\em \df} if $g$ satisfies the following conditions.
\begin{itemize}
\item [(T1)] $g(\varnothing)=1$
\item [(T2)] $g(1;\Delta)=g(\Delta)$\footnote{The values of $g$ for any sequence
containing the value $1$ is the same as the value of $g$ for the subsequence
without the $1$.}
\item [(T3)] $g(\tuple{x_1,\ldots,x,\ldots,y,\ldots,x_n})=g(\tuple{x_1,\ldots,y,\ldots,x,\ldots,x_n})$
\item [(T4)] $g(\Delta)=0$ if and only if $0 \in \Delta$
\item [(T5)] $g(\Delta)=1$ if and only if $x=1$ for every $x \in \Delta$
\item [(T6)] $g$ is continuous as a multi-variable function\footnote{In fact, this condition 
is only needed to guarantee the existence of solutions to the equations.}
\end{itemize}
\end{definition}

\medskip
\begin{example} Below are some examples of \df\ functions:
\label{ex:1.1}
\begin{enumerate}
\item $g(\Delta)=
\left\{\begin{array}{ll}
1, & \w{if $\Delta=\varnothing$}\\
\min\{x_i\}, & \w{if $\Delta=\tuple{x_1,\ldots,x_n}$}
\end{array}\right.$
\item $g(\Delta)=
\left\{\begin{array}{ll}
1, & \w{if $\Delta=\varnothing$}\\
\Pi_{1}^n(1-x_i), & \w{if $\Delta=\tuple{x_1,\ldots,x_n}$}
\end{array}\right.$
\item $g_{\lambda}(\Delta)=(1-\lambda)\min\{\frac{1}{2},g(\Delta)\}+\lambda\max\{\frac{1}{2},g(\Delta)\}$, for any $g$ satisfying (T1)--(T6).
\end{enumerate}
\end{example}

Later on, we will see that 
argumentation-friendly functions will be used both to calculate aggregation 
of attacks as well as for combining the value of attacks with initial values.
However, as we mentioned attack is a type of negation and hence when 
operating on the attack of a node with value $v$, we will consider
the complement of $v$ to $1$, i.e., $(1-v)$.

Notice that t-norms satisfy conditions (T1)--(T4) above.

\newcommand{\dom}[1]{\w{dom~$#1$}}

\begin{definition}\label{def:from-v-to-in-out}
For any assignment of values $v:S \longmapsto U$ define
the sets $in(v)=\{X \in \dom{v} \; | \; v(X)=1\}$ and 
$out(v)=\{X \in \dom{v} \; | \; v(X)= 0\}$. 
\end{definition}


\begin{theorem}\label{eq:comp-ext}
Let $\net{N}=\tuple{S,R}$ be a network, $g$ an \df\ function, and \eq{T} a system of 
equations written for
\net{N}, where for each node $X$, $\fg{X}=g_{Y\in Att(X)}
(\{1-\fg{Y}\})$. Take any solution $V$ to \eq{T}, it follows that
$in(V)$ is a complete extension.

\begin{proof} 
Suppose that $in(V)$ is not conflict-free. Then there
are $X,Y \in in(V)$, such that $(X,Y) \in R$. Since $Y \in in(V)$, then
$\fg{Y}=1=g_{W\in Att(Y)} (\{1-\fg{W}\})$. But $X\in Att(Y)$ and $X \in in(V)$, 
and hence $\fg{X}=1$. It then follows by (T4) that $g(\tuple{\ldots,0,\ldots})=0$
and hence $1\neq 0$, a contradiction. 

Now suppose that $X \in in(V)$. We show that for all $Y \in Att(X)$
there exists $Z \in in(V)$, such that $(Z,Y) \in R$. If $\fg{X}=1$, then 
$g_{Y\in Att(X)} (\{1-\fg{Y}\})=1$ and then by (T5) it follows that
$1-\fg{Y}=1$, for all $Y \in Att(X)$ and hence $\fg{Y}=0$ for all
$Y \in Att(X)$. Take any such $Y$. Since $\fg{Y}=0$, we have  
by (T4) that for some $W \in Att(Y)$, $\fg{W}=1$. It then follows that 
$W \in in(V)$.
\end{proof}
\end{theorem} 

\begin{theorem}
\label{th:existsSol}
Let $\net{N}=\tuple{S,R}$ be a network, $g$ an \df\ function, and \eq{T} a system
of equations written for \net{N}, where for each node $X$,
$\fg{X}=g_{Y\in Att(X)} \{1-\fg{Y}\}$. Then for every preferred
extension $E_{\net{N}}$ of $\net{N}$, there exists a solution $V$ to \eq{T}
such that
\begin{itemize}
\item[(C1)] If $X \in E_{\net{N}}$, then $\fg{X} = 1$
\item[(C2)] If $E_{\net{N}} \rightarrow X$, then $\fg{X} = 0$
\item[(C3)] If $X \not \in E_{\net{N}} \ws{and} E_{\net{N}} \not\rightarrow X$, then 
$0 < \fg{X} < 1$
\end{itemize}

\begin{proof} 
Let us start by partitioning the set $S$ using $E_{\net{N}}$ into three sets
$\Delta_1=E_{\net{N}}$, $\Delta_0=\{X \in S \; | \; E_{\net{N}} \rightarrow X\}$, and 
$\Delta_u=S \backslash (\Delta_0 \cup \Delta_1)$. Note that the elements of
$\Delta_u$ are the undecided elements in $S$ with respect to $E_{\net{N}}$.
Each element of $\Delta_u$ is not attacked by any element  of $\Delta_1$ and its
attackers cannot all come from $\Delta_0$, i.e., at least one attacker comes from
$\Delta_u$ itself. Consider the argumentation network \tuple{\Delta_u,R\rest \Delta_u}.
Write a system of equations \eq{T_u} using $g$ for \tuple{\Delta_u,R\rest \Delta_u}.
For each $X \in \Delta_u$, the equation is
\[V_u(X)=g_{Y\in\Delta_u\ws{s.t.}(Y,X)\in R\rest \Delta_u} \{1-V_u(Y)\}\]
By Brouwer's theorem, the above equations have a solution $V_u$.\footnote{The Euclidean version of the theorem states that if $g$ is a real-valued function, defined and continuous on a bounded closed interval $I$ of the real line where $g(x) \in I$, for all $x\in I$, then $g$ has a fixed-point. In our case, there are $n=|S|$ variables in the network \tuple{S,R}, which we can associate with the vector $\overrightarrow{X}$. We can then see each equation as $\overrightarrow{X}=\overrightarrow{g}(\overrightarrow{X})$, where $\overrightarrow{g}$ is a continuous function on the $n$-dimensional space $[0,1]^n$ (see Theorem~1.2 in \cite{Suli-Mayers:2003}).} To be
clear $V_u$ is defined on $\Delta_u$, giving values $V_u(X)$, such that
for every $X\in \Delta_u$, $V_u(X)=g_{Y\in\Delta_u\ws{s.t.}(Y,X)\in R\rest \Delta_u} 
\{1-V_u(Y)\}$

We are seeking however a solution $V$ defined for all of $S=\Delta_0\cup\Delta_1
\cup \Delta_u$, which satisfies the system of equations \eq{T} for \tuple{S,R}:
\[\fg{X}=g_{Y\in Att(X)} \{1-\fg{Y}\}\]
Furthermore, we want $V$ to be such that $V(X)=1$ for $X \in \Delta_1$, $V(X)=0$, 
for $X\in \Delta_0$ and $V(X)\in(0,1)$ for $X \in \Delta_u$. We now define such a
solution $V$. Let 

\begin{tabular}{lcl}
$V(X)=1$, & for all $X\in \Delta_1$\\
$V(X)=0$, & for all $X\in \Delta_0$\\
$V(X)=V_u(X)$, & for all $X\in \Delta_u$
\end{tabular}

We have to show now that $V$ indeed solves the system of equations \eq{T}
for \tuple{S,R}. Take $X\in S$:

\medskip

\noindent{\bf Case 1:} $X\in \Delta_1$. We defined $V(X)=1$. We need to show that $1=
g_{Y\in Att(X)}\allowbreak \{1-\fg{Y}\}$. Since $X \in E_{\net{N}}$, then all of its attackers are
in $\Delta_0$, and then $V(Y)=0$ (by definition), for all $Y \in Att(X)$. Therefore,
$g_{Y\in Att(X)} \{1-\fg{Y}\}=1$, by (T5).

\medskip
\noindent{\bf Case 2:} $X\in \Delta_0$. We defined $V(X)=0$. We need to show that $0=
g_{Y\in Att(X)}\allowbreak \{1-\fg{Y}\}$. Since $E_{\net{N}}\rightarrow X$, then there exists
$Y \in Att(X)$, such that $Y\in\Delta_1$. By definition, $V(Y)=1$, and then  
$g_{Y\in Att(X)} \{1-\fg{Y}\}=0$, by (T4).

\medskip
\noindent{\bf case 3:} $X\in \Delta_u$. We defined $V(X)=V_u(X)=g_{Y\in\Delta_u\ws{s.t.}
(Y,X)\in R\rest \Delta_u} \{1-V_u(Y)\}$. We need to show that $g_{Y\in Att(X)} \{1-V(Y)\}
= g_{Y\in\allowbreak\Delta_u\ws{s.t.} (Y,X)\in R\rest \Delta_u}\allowbreak \{1-V_u(Y)\}$. We noted above, 
that $X\in \Delta_u$ implies that none of its attackers belong to $\Delta_1$ and
therefore any remaining attackers $Z$ not in $\Delta_u$ must be in $\Delta_0$.
By definition, $V(Z)=0$, therefore $1-0=1$ and by (T2), such values can be safely 
deleted in the calculation of $g_{Y\in Att(X)} \{1-\fg{Y}\}$. Therefore, deleting all
such values will show that $g_{Y\in\Delta_u\ws{s.t.} (Y,X)\in R\rest \Delta_u} \{1-V_u(Y)\}= g_{Y\in\Delta_u \cup \Delta_0 \ws{s.t.} (Y,X)\in R} \{1-V_u(Y)\}$.

Having shown that $V$ above solves the system of equations \eq{T}, we can use
Theorem~\ref{eq:comp-ext} to show that $in(V)$ is a complete extension. We now ask whether any of the values $V_u(X)$, for $X\in\Delta_u$ can be $0$
or $1$. The answer is no, for if $V_u(X)=1$ for any $X \in \Delta_u$, then 
$V(X)=1$  and then $X\in in(V)\backslash E_{\net{N}}$, which is impossible, since
$E_{\net{N}}$ is a {\em preferred} extension. Analogously, we can only get $V(X)=0$
for some $X\in Delta_u$, if for some of its attackers $Z\in \Delta_u$, $V(Z)=1$,
which as we mentioned is impossible. This completed the proof.
\end{proof}
\end{theorem} 

The condition of preferred extension of the Theorem~\ref{th:existsSol} is 
necessary, as shown in the example below.

\begin{example} Consider the complete extension $E=\{X\}$ of the network 
below. $E$ is not preferred, since $E$ is a proper subset of $\{X,W\}$.

\medskip

\begin{center}
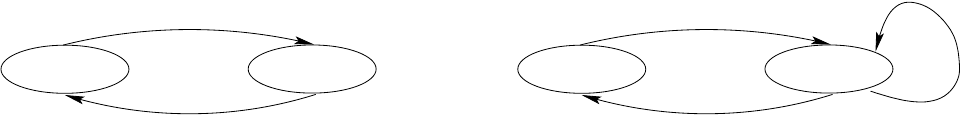
\end{center}

\medskip

The network generates the following equations.
\begin{eqnarray}
\fg{X} & = & 1-\fg{Y} \label{eq:fx}\\
\fg{Y} & = & 1-\fg{X} \label{eq:fy}\\
\fg{W} & = & 1-\fg{Z} \label{eq:fw}\\
\fg{Z} & = & g(\{1-\fg{W},1-\fg{Z}\}) \label{eq:fz}
\end{eqnarray}
Since $\fg{X}=1$, we get that $\fg{Y}=0$ and these values satisfy 
equations~(\ref{eq:fx}) and (\ref{eq:fy}) above.
However, replacing (3) in (4) gives us
\[\fg{Z} = g(\fg{Z},1-\fg{Z})\]
If $g$ is product, this gives us $\fg{Z} = \fg{Z}\cdot(1-\fg{Z}))$, and hence
$1=1-\fg{Z} \therefore \fg{Z}=0$, and hence $\fg{W}=1$, and therefore 
no solution corresponding to $E$ using $g$ exists. Note that the two 
preferred extensions $\{X,W\}$ and $\{Y,W\}$ include $W$. No 
extension can include $Z$.

However, with $g$ as $\min$, we have that (4) becomes
\[\fg{Z} = \min(\{1-\fg{W},1-\fg{Z}\})\]
and for this set of equations, the values $\fg{X}=1$, $\fg{Y}=0$,
$\fg{W}=\fg{Z}=\frac{1}{2}$ form a solution corresponding to $E$.
\end{example}

The loop in the example above is quite elucidating. Let us analyse it in
some more detail.

\begin{example} Consider the network with a single self-referencing loop 
below.

\medskip

\begin{center}
\begin{picture}(0,0)%
\includegraphics{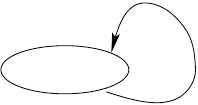}%
\end{picture}%
\setlength{\unitlength}{2072sp}%
\begingroup\makeatletter\ifx\SetFigFont\undefined%
\gdef\SetFigFont#1#2#3#4#5{%
  \reset@font\fontsize{#1}{#2pt}%
  \fontfamily{#3}\fontseries{#4}\fontshape{#5}%
  \selectfont}%
\fi\endgroup%
\begin{picture}(1799,937)(1728,-1973)
\put(2321,-1719){\makebox(0,0)[b]{\smash{{\SetFigFont{6}{7.2}{\familydefault}{\mddefault}{\updefault}{\color[rgb]{0,0,0}$X$  }%
}}}}
\end{picture}%

\end{center}

\medskip

The network generates the equation:
\begin{eqnarray*}
\fg{X} & = & g(\{1 - \fg{X}\})
\end{eqnarray*}
Notice that $g(\{1 - \fg{X}\})=1-\fg{X}$ and hence we have that
$\fg{X}=1-\fg{X} \therefore \fg{X}=\frac{1}{2}$, whatever the function $g$ is,
as long as it satisfies (T1)--(T5).
\end{example}

Note that $\min$ satisfies (T1)--(T4). As a result, we have that:

\begin{corollary}
Let $\net{N}=\tuple{S,R}$ be a network and \eq{T} a system of equations written for
\net{N}, where for each node $X$, $\fg{X}=\min_{Y\in Att(X)}
(\{1-\fg{Y}\})$. Take any solution $V$ to \eq{T}. It follows that
$in(V)$ is a complete extension.
\end{corollary}

This follows from Theorem~\ref{eq:comp-ext}. What it means is that any 
solution to the system of equations defined in
terms of \eqmax\ can be translated into a complete extension simply by
defining that extension as the set containing the nodes whose
solution values are $1$. Obviously, different solutions will 
give rise to different extensions. 

\begin{proposition}
\label{cor:complete-extensions}
Let $\net{N}=\tuple{S,R}$ be a network and \eq{T} a system
of equations written for \net{N}, where for each node $X$,
$\fg{X}=\min_{Y\in Att(X)}(\{1-Y\})$. Then for every {\em complete}
extension $E$ of $\net{N}$, there exists a solution $V$
to \eq{T} satisfying:
\begin{itemize}
\item[(C1)] If $X \in E$, then $\fg{X} = 1$.
\item[(C2)] If $E \rightarrow X$, then $\fg{X} = 0$.
\item[(C3)] If $X \not \in E \ws{and} E \not\rightarrow X$, then 
$0 < \fg{X} < 1$.
\end{itemize}
\begin{proof}
Let $E$ be a complete extension. Consider the following assignment of
values to the nodes in $S$:
\begin{itemize}
\item if $X\in E$, then $\fg{X} = 1$
\item if $E \rightarrow X$, then $\fg{X} = 0$
\item $\fg{X}=\frac{1}{2}$, otherwise
\end{itemize}
We now show that the values above form a solution to the system of
equations \eq{T}. As in Theorem~\ref{th:existsSol}, replacing the above values 
in the original system of equations will reduce them  to the following
types.
\begin{itemize}
\item[(1)] $1 = \min(\Delta_1)$
\item[(2)] $0 = \min(\Delta_2)$
\item[(3)] $\frac{1}{2}= \min(\Delta_3)$ 
\end{itemize}
We have seen that $\Delta_1=\{1\}$ and since $1=\min(\{1\})$, (1) is satisfied. 
Similarly, $0 \in \Delta_2$ and since $\min(\{0,\ldots\})=0$, so is (2). Notice 
that the image of $V$ is $\{0,\nicefrac{1}{2},1\}$.  All values
in $\Delta_3$ are greater than $0$, but at least one of them is
$\frac{1}{2}$, therefore $\min(\Delta_3)=\frac{1}{2}$, and hence
the above assignment solves the equations.
\end{proof}
\end{proposition}

So far, we have shown the basics of the equational numerical approach to
abstract argumentation frameworks. In the next section we consider two 
additional developments that follow naturally. Firstly, we know
that solutions do exist to the system of equations, but can we 
find them using some numerical method? For example, by applying iterations 
given some initial guess?\footnote{As can be done to find the square root 
of numbers using Newton's method.} Secondly, we would like to apply our 
methodology to questions of merging, voting, or any other application 
where a set of initial values emerges and needs to be transformed to 
the ``closest'' extension. How can we do that? The following
section provides a method to answer these questions.

\section{The Gabbay-Rodrigues Iteration Schema\label{sec:initial-values}}

Suppose we are given initial values which do not correspond to any
extension in the way that we presented them in the previous
section. These values may come attached to the nodes for different
reasons. For instance, the arguments themselves may be expressed as some proof 
in a fuzzy logic and the initial values can represent the values of the 
conclusions of the proofs, or they can be obtained as the result of the merging 
of some networks, or they may come from some voting mechanism, etc.
Whatever the reason, the initial values may or may not correspond to a complete
extension in Dung's sense and we seek a mechanism that would allow us to find 
the ``best'' possible extension corresponding to them. 

Consider the equation \eqmax:
  \insertnumequation{$\textstyle\fg{X}=1-\max_{Y_i \in Att(X)}\{\fg{Y_i}\}$}%
                    {\eqmax}
\eqmax\ is satisfied when the value of the node $X$ is {\em legal} (in
Caminada and Pigozzi's terminology \cite{Caminada-Pigozzi:11}). That is, 
if the value of $X$ is $1$ and the value of all of $X$'s attackers are $0$; or
if the value of $X$ is $0$ and at least of one $X$'s attackers has value $1$; or
if the value of $X \in (0,1)$ and at least one of $X$'s attackers has value 
in $(0,1)$ and no attacker of $X$ has value $1$.
If we aim to correct the values of the nodes in a network iteratively, we need
a mechanism that leaves legal \inc, \exc\ and \und\ node values intact, 
changing illegal \inc\ or \exc\ values into \und.\footnote{We will come to
the correction of illegal \und\ nodes later.} To make a distinction
between these classes of values, we will call the values in $\{0,1\}$ {\em crisp} and the
values in $(0,1)$ {\em undecided}. 

Now consider the following averaging function:
\[
   \textstyle(1-X) \cdot \min\left\{\frac{1}{2},1 - \max_{Y\in Att(X)} Y\right\} +
   X \cdot \max\left\{\frac{1}{2},1 - \max_{Y\in Att(X)} Y\right\}
\]
For legal assignments of values, we have three cases to consider:
\begin{itemize}
\item[\eqr{L1}] $X$ is legally \inc. In this case $X=1$ and all of its attackers
have value $0$. We want the value of $X$ to remain $1$. We have that:
\begin{eqnarray*}
   (1-X) \cdot\min\left\{\frac{1}{2},1 - \displaystyle\max_{Y\in Att(X)} Y\right\} +
   X \cdot \max\left\{\frac{1}{2},1 - \max_{Y\in Att(X)} Y\right\} & \hspace*{-3mm}= \hspace*{-3mm}& \\
   1 \cdot \max\left\{\frac{1}{2},1\right\} & = & \\
                                 & = & 1
\end{eqnarray*}

\item[\eqr{L2}] $X$ is legally \exc. In this case $X=0$ and at least one of its
attackers has value $1$. We want the value of $X$ to remain $0$.
We have that:
\begin{eqnarray*}
  1 \cdot \min\left\{\frac{1}{2},1 - \max_{Y\in Att(X)} Y\right\} +
  X \cdot \max\left\{\frac{1}{2},1 - \max_{Y\in Att(X)} Y\right\} & = & \\
  1 \cdot \min\left\{\frac{1}{2},0\right\} + 0 \cdot  \max\left\{\frac{1}{2},0\right\}& = & \\
                                 & = & 0
\end{eqnarray*}
\item[\eqr{L3}]$X$ is legally \und. In this case $0<X<1$, none of its
attackers has value $1$ and at least one of its attackers has value greater 
than $0$. This means that $0 < \max_{Y\in Att(X)} Y < 1$ and therefore $0 < (1 - 
\max_{Y\in Att(X)} Y) < 1$. Let $\alpha_1= \min\left\{\frac{1}{2},1 - \max_{Y\in Att(X)} Y\right\}$
and $\alpha_2=\max\left\{\frac{1}{2},1 - \max_{Y\in Att(X)} Y\right\}$. It follows that
$0 < \alpha_1 < 1$ and $0 < \alpha_2 < 1$. We want the value of $X$ to remain
undecided, although we are prepared to accept changes to its initial value
as long as its final value remains in the interval $(0,1)$. We have that:
\begin{eqnarray*}
   (1-X) \cdot \min\left\{\frac{1}{2},1 - \max_{Y\in Att(X)} Y\right\} +
   X \cdot \max\left\{\frac{1}{2},1 - \max_{Y\in Att(X)} Y\right\} & = & \\
   (1-X) \cdot \alpha_1 + X \cdot  \alpha_2 & = & \\
   \alpha_1 - X \cdot  \alpha_1 +  X \cdot  \alpha_2 & = & \\
   \alpha_1 - X \cdot  (\alpha_1 - \alpha_2) & = & \kappa\\
\end{eqnarray*}
Notice that $\alpha_1 \leq \frac{1}{2}$ and $\alpha_2 \geq \frac{1}{2}$,
therefore $\alpha_2 \not < \alpha_1$. If $\alpha_1 = \alpha_2$, then $\kappa = 
\alpha_1$ and hence $0 < \kappa \leq  \frac{1}{2}$. If $\alpha_1 < \alpha_2$, 
then $0< \alpha_1 < \frac{1}{2}$ and $\alpha_2=\frac{1}{2}$. Therefore,
$-\frac{1}{2} < (\alpha_1 - \alpha_2) < 0$. It then follows that
$0 < \alpha_1 \leq \kappa < \frac{1}{2}$ and therefore the value of 
$X$ remains in $(0,1)$.
\end{itemize}

What \eqr{L1}--\eqr{L3} above give us is that legal labellings are preserved.%
\footnote{Legal undecided values may change, although they remain in the undecided
range (by \eqr{L3}).} Later on, we shall see that our iteration schema also
eventually {\em corrects} all illegal values. It does so in two stages. In the first stage,
all illegal crisp values are turned into undecided (this is done in $t \leq |S|$
iterations). In the second stage, all remaining illegal undecided values 
converge to whatever legal crisp values they should be, so that in the limit,
all of the values in the sequence are legal. Therefore, the {\em Gabbay-Rodrigues 
Iteration Schema} introduced below provides a numerical iterative method to turn 
any initial illegal assignment of values to arguments into its closest legal 
assignment.\footnote{The precise definition of ``closest'' will be made clear 
in Theorem~\ref{th:main-theorem}.}

\begin{definition}
Let $\net{N}=\tuple{S,R}$ be an argumentation network and $V_0$ be an
assignment of values to the nodes in $S$. The {\em Gabbay-Rodrigues 
Iteration Schema} is defined by the following system of equations \eq{T}, 
where for each node $X \in S$, the value $V_{i+1}(X)$ is defined in
terms of the values of the nodes in $V_i$ as follows:

\medskip
\parbox[t]{10cm}{\begin{tabular}{rcl}
$V_{i+1}(X)$ & $=$ & $(1-V_i(X)) \cdot \min\left\{\frac{1}{2}, 1 - \max_{Y\in Att(X)}V_i(Y)\right\} 
  \; +$\\[1ex]
     &  & $V_i(X) \cdot \max\left\{\frac{1}{2},1 - \max_{Y\in Att(X)}V_i(Y)\right\}$
\end{tabular}}\flushright{\raisebox{5mm}[-16mm]{\eqr{T}}}
\end{definition}

We call the system of equations for \net{N} using the above iteration schema its
{\em GR system of equations}.

We ask whether we can regard the iteration schema above as an 
{\em equation} schema
as in the previous section, i.e.,
\[ X = (1-X) \cdot \min\left\{\frac{1}{2},1 - \max_{Y\in Att(X)}Y\right\} +
   X \cdot \max\left\{\frac{1}{2},1 - \max_{Y\in Att(X)}Y\right\} \eqno \eqr{GR}\]
   
To further
clarify this point, let us take an equation written with an \df\ function $g$ for a node $X$ in terms of its attackers. The equation would be
\[ X = g(\cup_{Y\in Att(X)}\{1-Y\})\]
It is clear that if one of the attackers of $X$ is $1$, the value of
$X$ solves to $0$, and if all the attackers of $X$ are $0$, the value
of $X$ will solve to $1$. This follows from the properties (T1)--(T5)
of an \df\ function. Now let us compare and see what happens when we
use the formula above. If the value of one of the
attackers of $X$ is $1$, the first component of the sum will be $0$,
whereas the second component will be $\frac{1}{2}$, because the
equation is implicit, we have the equation
\[ X = \frac{X}{2}\]
which solves to $X=0$, which is correct. If the values of all
attackers of $X$ are $0$, then we get the equation
\[X=\frac{(1-X)}{2}  + X\]
which solves to $X=1$, which again gives a correct result. Otherwise,
assume that the values of all attackers are either $0$ or
$\frac{1}{2}$, with at least one of them being $\frac{1}{2}$. We get
the equation
\[X=\frac{(1-X)}{2}  + \frac{X}{2}\]
which again solves to the correct value of $X=\frac{1}{2}$. By correct
we mean that the results are exactly compatible with the Caminada
labelling mentioned in Section~\ref{sec:intro}, where $X=1$ means $X$ is
\inc, $X=0$ means $X$ is \exc\ and $X=\frac{1}{2}$ means $X$ is \und. 

Therefore, the Gabbay-Rodrigues schema remains faithful to the spirit of Dung's 
semantics captured through the legal Caminada labellings just as \eqmax\ does. Its 
advantage over \eqmax\ is that it can be used iteratively as we will show in the rest 
of this section. \footnote{As an {\em equation}, we can regard the expression ~\eqr{GR} 
just as another type of $g$, a special $eq_{GR}$.}

We start by showing some properties of the schema. The first
one ensures that the values of all nodes remain in the unit interval in
all iterations.

\begin{proposition}\label{prop:between-0-and-1}
Let $\net{N}=\tuple{S,R}$ be an argumentation network and $V_0:S\longmapsto U$
an assignment of initial values to the nodes in $S$. Let each assignment $V_i$, $i>0$, be calculated by the Gabbay-Rodrigues Iteration Schema for \net{N}. It follows that $V_{i}(X)\in U$, for all $i \geq 0$ and all $X \in S$.

\begin{proof} The base of the induction is the initial value assignment that
holds trivially. The induction step is proven by looking at the maximum and
minimum values that the nodes can take and showing that the sum in the iterated
schema is always a number in $U$.
Now, suppose that indeed for all nodes $X\in S$, $0 \leq
V_{k}(X) \leq 1$, for a given iteration $k$. Pick any node $X$. It follows that
\begin{eqnarray*}
V_{k+1}(X) & = &
   (1-V_{k}(X)) \cdot \min\left\{\nicefrac{1}{2},1 - \max_{Y\in Att(X)}V_{k}(Y)\right\} +\\
      &   &
   V_{k}(X) \cdot \max\left\{\nicefrac{1}{2},1 - \max_{Y\in Att(X)}V_{k}(Y)\right\}
\end{eqnarray*}
So we have that $V_{k+1}(X)=(1-\alpha)\cdot x + \alpha \cdot y$, where
$0 \leq \alpha \leq 1$, $0 \leq (1-\alpha) \leq 1$, $0 \leq x \leq
\nicefrac{1}{2}$, and $\nicefrac{1}{2} \leq y \leq 1$.

The lowest value for $V_{k+1}(X)$ is obtained with the lowest values for $x$
and $y$, when we get that $V_{k+1}(X)=\frac{\alpha}{2}$. If $\alpha=0$, then
$V_{k+1}(X)=0 \geq 0$. If $\alpha=1$, then we get $V_{k+1}(X)=\nicefrac{1}{2}
\leq 1$. The highest value for $V_{k+1}(X)$ is obtained with the highest
values for $x$ and $y$, when we get that $V_{k+1}(X)=\frac{(1-\alpha)}{2}
+ \alpha$. If $\alpha=0$, then $V_{k+1}(X)=\nicefrac{1}{2} \leq 1$. If
$\alpha=1$, then we get $V_{k+1}(X)=1 \leq 1$. In all cases, $0 \leq
V_{k+1}(X) \leq 1$.
\end{proof}
\end{proposition}

We now show that a given ``legal'' set of initial values for the nodes
in $S$ satisfies the equations and hence the values remain unchanged.

\begin{proposition}
\label{prop:admissible-extensions}
Let $\net{N}=\tuple{S,R}$ be a network and \eq{T} its GR system
of equations.
Then for every complete extension $E$ of $\net{N}$ and all
$X\in S$, if $V_0$ is defined using $E$ by the clauses (C1)--(C3)
below, we have that $V_1(X)=V_0(X)$.
\begin{itemize}
\item[(C1)] If $X \in E$, then $V_0(X) = 1$
\item[(C2)] If $E \rightarrow X$, then $V_0(X) = 0$
\item[(C3)] If $X \not \in E$ and $E \not\rightarrow X$, then 
$V_0(X) = \frac{1}{2}$
\end{itemize}

\begin{proof} Let $E$ be a complete extension and suppose
$V_0(X)=1$. Then $X \in E$ and hence, {\em i)} either $Att(X)=
\varnothing$, or {\em ii)} for all $Y \in Att(X)$, $E \rightarrow 
Y$ (since $E$ is admissible). As a result, $1-\max_{Y\in Att(X)}\{\fg{Y}\}
=1$, and hence we have that
\[ V_1(X) = \max\left\{\frac{1}{2},1\right\} = 1 = V_0(X).\]
If on the other hand, $V_0(X)=0$, then $E \rightarrow X$.
Therefore, there exists some $Y \in Att(X)$, such that $Y \in E$ and hence
$V_0(Y)=1$. It follows that
\[ V_1(X) = \min\left\{\frac{1}{2},1-1\right\}=0=V_0(X).\]
Finally, if $V_0(X)=\frac{1}{2}$, then $X \not \in E$ and $E 
\not\rightarrow X$. We must have that for all $Y \in Att(X)$, 
$V_0(Y)<1$ (otherwise, we would have that $E \rightarrow X$). 
We must also have that for some $Y \in Att(X)$, $V_0(Y)>0$,
otherwise $E$ would defend $X$ and since it is complete
$X \in E$, but then $V_0(X)=1$. Therefore, $1-\max_{Y\in Att(X)}\{\fg{Y}\}
=\frac{1}{2}$, and hence we have that
\[V_1(X) = \frac{1}{2} \cdot \min\left\{\frac{1}{2},\frac{1}{2}\right\} +
   \frac{1}{2} \cdot \max\left\{\frac{1}{2},\frac{1}{2}\right\}=\frac{1}{4}+\frac{1}{4}=
\frac{1}{2}=V_0(X).\]
\end{proof}
\end{proposition}

Obviously, if for all nodes $X$, $V_1(X)=V_0(X)$ as above, then for all 
nodes $X$, $V_{i+1}(X)=V_i(X)$, for all $i\geq 0$.

Furthermore, crisp values do not ``swap'' between each other and undecided 
values do not become crisp:

\begin{theorem}
\label{th:maximality}
Let $\net{N}=\tuple{S,R}$ be an argumentation network, \eq{T} a system
of equations for \net{N} using the Gabbay-Rodrigues Iteration Schema,
and $V_0:S\longmapsto U$ an assignment of  initial values to the nodes in
$S$. Let $V_0$, $V_1$, $V_2$, \ldots\ be a sequence of value assignments
where each $V_i$, $i>0$, is generated by \eq{T}.
Then the following properties hold for all $X\in S$ and for all $k\geq 0$
\begin{enumerate}
\item If $V_k(X) = 0$, then $V_{k+1}(X) \neq 1$.
\item If $V_k(X) = 1$, then $V_{k+1}(X) \neq 0$.
\item If $0 < V_k(X) < 1$, then $0 < V_{k+1}(X) < 1$.
\end{enumerate}
\begin{proof} ~

\begin{enumerate}
\item Suppose $V_k(X) = 0$, then $V_{k+1}(X) = \min\left\{\nicefrac{1}{2},1 -
\max_{Y\in Att(X)}V_i(Y)\right\}\leq \nicefrac{1}{2}$.
\item Suppose $V_k(X) = 1$, then $V_{k+1}(X) = \max\left\{\nicefrac{1}{2},1 -
\max_{Y\in Att(X)}V_i(Y)\right\}\geq \nicefrac{1}{2}$.

\item Suppose $0 < V_k(X) < 1$. We first show that $V_{k+1}(X)>0$.
Note that $0 < (1-V_k(X)) < 1$. Therefore, we have that

\begin{eqnarray*}
V_{k+1}(X) & = &(1-V_k(X))\cdot \min\left\{\nicefrac{1}{2},1 - \max_{Y\in Att(X)}V_i(Y)\right\} +\\
  &   & V_k(X)\cdot \max\left\{\nicefrac{1}{2},1 - \max_{Y\in Att(X)}V_i(Y)\right\}
\end{eqnarray*}
It is easy to see that the first component of the above sum is greater than 
or equal to $0$, whereas the second is strictly greater than $0$, and hence
$V_{k+1}(X)>0$.

Since we start with values in $U$,
Proposition~\ref{prop:between-0-and-1}, gives us that $V_{k+1}(X) \leq 1$,
for all $X \in S$. We therefore only need to show that
$V_{k+1}(X) \neq 1$. Again we have that $V_{k+1}(X)=(1-\alpha)\cdot x +
\alpha \cdot y$, where
\begin{eqnarray*}
0 < \alpha < 1 \\
0 < (1-\alpha) < 1 \\
0 \leq x \leq \nicefrac{1}{2} \\
\nicefrac{1}{2} \leq y \leq 1 \\
\end{eqnarray*}
Suppose $V_{k+1}(X)=1$. It follows that
\begin{eqnarray*}
(1-\alpha)\cdot x + \alpha \cdot y  & = & 1 \\
x - \alpha \cdot x + \alpha \cdot y & = & 1 \\
\alpha(y - x) & = & (1 - x) \\
\alpha & = & \frac{1 - x}{y-x}
\end{eqnarray*}
Since $\alpha < 1$, we have that $1-x < y - x$, and hence $y > 1$, a
contradiction.
\end{enumerate}
\end{proof}
\end{theorem}

The above theorem shows that any changes between iterations can only 
generate new values for nodes in the interval $(0,1)$, i.e., successive 
iterations can only turn crisp values into undecided. Therefore, the 
sets of nodes with crisp values can only decrease throughout the 
iterations:

\begin{corollary}
\label{cor:inclusiveness}
Let $\net{N}=\tuple{S,R}$ be an argumentation network, $V_0:S \longmapsto U$ an 
initial assignment of values to the nodes in $S$ and \eq{T} its GR system
of equations. It follows that for all $0 \leq i\leq j$, $in(V_j) \subseteq 
in(V_i)$ and $out(V_j)\subseteq out(V_i)$.
\end{corollary}

The situation in the 
{\em limit} of the sequence of values is more complex and we will deal with 
it later. If between two successive iterations there are no changes in the
crisp values, then these values ``stabilise'':

\begin{theorem} \label{th:in-out-stability}
Let $\net{N}=\tuple{S,R}$ be a network, \eq{T} its GR system
of equations, and $V_0$ an initial assignment of values to the nodes in $S$.
Let $V_0$, $V_1$, $V_2$, \ldots\ be a sequence of value assignments
where each $V_i$, $i>0$, is generated by \eq{T}.
Assume that for some iteration $i$ and all nodes $X\in S$ such that
$V_i(X)\in \{0,1\}$, we have that $V_{i+1}(X)=V_i(X)$, then for all
$j\geq 1$, $V_{i+j}(X)=V_i(X)$.

\begin{proof} Assume that $V_i(X)\in\{0,1\}$ for some node $X$. There are
two cases to consider.

\medskip

\noindent{\bf Case 1:} $V_i(X)=0$. By assumption, we have that 
$V_{i+1}(X)=0$. We show that $V_{i+2}(X)=0$. If $V_{i+1}(X)=0$, we have that
\begin{eqnarray*}
V_{i+1}(X) & = & (1-V_i(X))\cdot \min\left\{\frac{1}{2},1 -\max_{Y\in
  Att(X)} \{V_i(Y)\}\right\} +\\ & & V_i(X)\cdot \max\left\{\frac{1}{2},1
-\max_{Y\in Att(X)} \{V_i(Y)\}\right\}\\
0 & = & \min\left\{\frac{1}{2},1 -\max_{Y\in  Att(X)} \{V_i(Y)\}\right\} 
\end{eqnarray*}
So, $\max_{Y\in  Att(X)} \{V_i(Y)\}=1$ and hence for some $Y \in
Att(X)$, $V_i(Y)=1$. By assumption $V_{i+1}(Y)=1$ and hence
$\max_{Y\in  Att(X)} \{V_{i+1}(Y)\}=1$. Therefore,
\[V_{i+2}(X)= \min\left\{\frac{1}{2},1 -\max_{Y\in Att(X)} \{V_{i+1}(Y)\}\right\}=0\] 

\noindent{\bf Case 2:} $V_i(X)=1$. By assumption, we have that 
$V_{i+1}(X)=1$. We show that $V_{i+2}(X)=1$. If $V_{i+1}(X)=1$, we have that
\begin{eqnarray*}
V_{i+1}(X) &=& (1-V_i(X))\cdot\min\left\{\frac{1}{2},1-\max_{Y\in Att(X)}\{V_i(Y)\}\right\}+\\ 
          & & V_i(X) \cdot \max\left\{\frac{1}{2},1-\max_{Y\in Att(X)} \{V_i(Y)\}\right\}\\ 
        1 &=& \max\left\{\frac{1}{2},1-\max_{Y\in Att(X)} \{V_i(Y)\}\right\}\\ 
\end{eqnarray*}
So, $\max_{Y\in Att(X)} \{V_i(Y)\}=0$, and hence for all $Y \in Att(X)$,
$V_i(Y)=0$. By assumption, $\max_{Y\in Att(X)} \{V_{i+1}(Y)\}=0$, and hence
\[V_{i+2}(X)= \max\left\{\frac{1}{2},1-\max_{Y\in Att(X)}\{V_{i+1}(Y)\}\right\}=1\]
\end{proof}
\end{theorem}

\begin{definition}\label{def:in-out-stability} Let $\net{N}=\tuple{S,R}$ be an
argumentation network and $V_0:S\longmapsto U$ an assignment of initial values 
to the nodes in $S$. A sequence of assignments $V_{i}:S \longmapsto U$
where each $i >0$ is generated by the Gabbay-Rodrigues Iteration Schema for
\net{N} becomes {\em stable at iteration $k$}, if for all nodes $X \in S$ we 
have that
\begin{enumerate}
\item If $V_k(X) \in \{0,1\}$, then $V_{k+1}(X)=V_k(X)$\label{cor:stab-two}; and
\item $k$ is the smallest value for which the condition above holds.
\end{enumerate}
\end{definition}

Note that if $V_k(X) \in (0,1)$, then $V_{k+1}(X)\in (0,1)$, for all
$k\geq 0$, by Theorem~\ref{th:maximality}.

\begin{corollary}\label{cor:in-out-stability} Consider a sequence of value
assignments $V_0$, $V_1$, $V_2$, \ldots as described in
Theorem~\ref{th:in-out-stability}. If the sequence becomes stable at iteration 
$k$, then the sequence remains stable for all iterations $k+j$, $j\geq 0$.

\begin{proof}
The first stability condition in Definition~\ref{def:in-out-stability}
follows from Theorem~\ref{th:maximality} and the second condition
follows from Theorem~\ref{th:in-out-stability}.
\end{proof}
\end{corollary}

\begin{corollary} \label{cor:max-iter}
Let $\net{N}=\tuple{S,R}$ be an argumentation network, $V_0:S\longmapsto U$ 
an assignment of initial values to the nodes in $S$ and \eq{T} its GR
system of equations. The following hold:
\begin{enumerate}
\item\label{cor:max-iter-one} If the sequence of value assignments is
  not stable at iteration $k$, then there exists $X \in S$, such that
  $V_k(X)\in \{0,1\}$ and $V_{k+1}(X)\in (0,1)$.
\item\label{cor:max-iter-two} Let $|S|=n$. Then, the sequence is
  stable for some $k \leq n$.
\end{enumerate}
\begin{proof}
(\ref{cor:max-iter-one}) follows from Theorem~\ref{th:maximality}. For
  (\ref{cor:max-iter-two}), notice that each iteration $i$ which is
  not stable causes at least one node to change value from $\{0,1\}$
  into $(0,1)$. Theorem~\ref{th:maximality} states that all values in
  $(0,1)$ remain in $(0,1)$.  Since $S$ is finite, there are only
  finitely many nodes that can change from $\{0,1\}$ into $(0,1)$ and
  the number of iterations in which this can happen is bounded by
  $|S|$.
\end{proof}
\end{corollary}

Corollary~\ref{cor:max-iter} shows that for some value $0 \leq k \leq |S|$, 
the sequence of value assignments $V_0(X), V_{1}(X), V_{2}(X), \ldots$
eventually becomes stable. That is, there exists $k \geq 0$, such that for 
all $j \geq 0$ and all nodes $X$
\begin{itemize}
\item if $V_k(X)=0$, then $V_{k+j}=0$;
\item if $V_k(X)=1$, then $V_{k+j}=1$; and
\item if $V_k(X)\in (0,1)$, then $V_{k+j}\in (0,1)$.
\end{itemize} 

\begin{remark}\label{rem:using-g-in-GR}
Given an \df\ function $g$, we can define the Gabbay-Rodrigues Iteration Schema 
for $g$, denoted by GR(g), as follows.

\begin{eqnarray*}
V_{i+1}(X) & = & 
   (1-V_i(X)) \cdot \min\left\{\frac{1}{2},g(\cup_{Y\in Att(X)}\{1-V_i(Y)\})\right\} +\\
      &   & 
   V_i(X) \cdot \max\left\{\frac{1}{2},g(\cup_{Y\in Att(X)}\{1-V_i(Y)\})\right\}
\end{eqnarray*}
If we further assume that $g$ satisfies the optional condition
\begin{itemize}
\item [(T6)] If for all $x\in\Delta$, we have that $x<1$ and for some $x\in\Delta$,
$x>0$, then $g(\Delta)\in(0,1)$.
\end{itemize}
Then the above sequence of definitions and theorems in this section still holds if we
replace $GR$ by $GR(g)$. 
\end{remark}

The above discussion laid out the properties of the Gabbay-Rodrigues
Iteration Schema. In what follows we shall apply it to the following
question. Suppose we have an argumentation network \tuple{S,R} with
associated equations and an initial assignment $f:S \longmapsto
U$. $f$ may come from a single agent who insists on giving certain
values to the arguments of $S$; or $f$ may be the result of merging
several argumentation frameworks with the nodes in $S$ (through some
well-defined process, e.g., voting); or $f$ may arise from any other
process. Our problem is to find the function $f^{\prime}$,
closest to $f$, which also corresponds to an extension of \tuple{S,R} (for
example, solves the equations generated from \tuple{S,R}). Now, what do we mean by
``closest''? Following Caminada and Pigozzi
\cite{Caminada-Pigozzi:11}, we take the view that ``closest'' means
agreeing on the maximal number of nodes with $f$-values in
$\{0,1\}$. In what follows, we show how to find such an assignment
$f^{\prime}$, through the Gabbay-Rodrigues Iteration Schema.

\begin{theorem}
\label{th:unique-maximal-admissible}
Let \tuple{S,R} be a network and $f:S \longmapsto U$ an assignment of
values to the nodes in $S$. Then there is an assignment
$h:S\longmapsto U$ such that the sets $in(h) \subseteq in(f)$ and
$out(h) \subseteq out(f)$ are maximal and for every node $X\in S$:

\begin{eqnarray}
\w{If $h(X)=1$, then $\textstyle\max_{Y\in Att(X)}\{h(Y)\}=0$; and} \label{adm1}\\
\w{If $h(X)=0$, then $\textstyle\max_{Y\in Att(X)}\{h(Y)\}=1$.} \label{adm2}
\end{eqnarray}

\begin{proof}
The proof is analogous to the proof of Theorem~5 in
\cite{Caminada-Pigozzi:11}.


Take any two assignments $g_1$ and $g_2$ such that for all  $X \in S$:
\begin{itemize}
\item $g_1(X)=0$ implies $f(X)=0$ and $g_2(X)=0$ implies $f(X)=0$; and
\item $g_1(X)=1$ implies $f(X)=1$ and $g_2(X)=1$ implies $f(X)=1$
\end{itemize}
and
\begin{eqnarray}
\w{If $g_1(X)=1$, then $\textstyle\max_{Y\in Att(X)}\{g_1(Y)\}=0$; and} \label{g1in}\\
\w{If $g_2(X)=1$, then $\textstyle\max_{Y\in Att(X)}\{g_2(Y)\}=0$; and} \label{g2in}\\
\w{If $g_1(X)=0$, then $\textstyle\max_{Y\in Att(X)}\{g_1(Y)\}=1$; and} \label{g1out}\\
\w{If $g_2(X)=0$, then $\textstyle\max_{Y\in Att(X)}\{g_2(Y)\}=1$} \label{g2out}
\end{eqnarray}


It follows that $in(g_1)\subseteq in(f)$ and $out(g_1)\subseteq out(f)$;
and $in(g_2)\subseteq in(f)$ and $out(g_2)\allowbreak \subseteq out(f)$. 

Let us construct an assignment $h:S \longmapsto U$, such that
for all $X \in S$:

\begin{eqnarray}
h(X)=1  \ws{iff}  \max(g_1(X),g_2(X))=1 \label{h1}\\
h(X)=0 \ws{iff}  \min(g_1(X),g_2(X))=0 \label{h2}\\
h(X)=\nicefrac{1}{2} \ws{iff} 0 < g_1(X) < 1 \ws{and} 0 < g_2(X) < 1\label{h3}
\end{eqnarray}

We now show that the assignment $h$ is a well-defined function and that
$in(h) \subseteq in(f)$ and that $out(h)\subseteq out(f)$. It is easy to
see that every node $X$ gets at least one value $h(X)$. We need to show
that for every node $X$, this value is unique and that the above inclusions
are satisfied. From (\ref{h3}), it is easy to see that $h(X)$ is equal to 
$\nicefrac{1}{2}$ if and only if both $g_1(X)\in (0,1)$ and $g_2(X) \in 
(0,1)$. 
To show inclusion, suppose $X \in in(h)$. Then $h(X)=1$ and hence 
$\max(g_1(X),g_2(X))=1$. Either $g_1(X)=1$ or $g_2(X)=1$ (or both), 
and hence $f(X)=1$. Therefore $X \in in(f)$. To show that
$h(X)$ is unique in this case, it is sufficient to show that 
$\min(g_1(X),g_2(X))>0$. Suppose $\min(g_1(X),g_2(X))=0$, then either 
$g_1(X)=0$ or $g_2(X)=0$, 
in which case $f(X)=0$, a contradiction, since $f$ is a function. 
Analogously, if $X \in out(h)$, then $h(X)=0$ and hence $\min(g_1(X),g_2(X))=0$.
Then either $g_1(X)=0$ or $g_2(X)=0$ (or both), and hence $f(X)=0$. Therefore, 
$X \in out(f)$. To show that $h(X)$ is also unique in this case,
it suffices to show that $\max(g_1(X),g_2(X))<1$. Suppose that
$\max(g_1(X),g_2(X))=1$, then either $g_1(X)=1$ or $g_2(X)=1$, in
which case $f(X)=1$, a contradiction, since $f$ is a function.

We now show that $h$ satisfies (\ref{adm1}) and (\ref{adm2}). 

Suppose $h(X)=1$. 
By construction, $\max(g_1(X),g_2(X))=1$. It follows that {\em i)}
either $X \in in(g_1)$, and then by~(\ref{g1in}),   $\max_{Y \in Att(X)}\{g_1(Y\}=0$. This means that for every $Y \in Att(X)$, $g_1(Y)=0$. By (\ref{h2}), for every $Y \in Att(X)$,
$h(Y)=0$, and hence $\max_{Y\in Att(X)}\{h(Y)\}=0$; or {\em ii)} $X \in in(g_2)$, and 
then by (\ref{g2in}), $\max_{Y \in Att(X)}\{g_2(Y\}=0$. By (\ref{h2}), for 
$in(g_2)$ is also admissible, $Y \in out(g_2)$, and hence for every $Y \in Att(X)$,
$h(Y)=0$, and hence again $\max_{Y\in Att(X)}\{h(Y)\}=0$. This shows that $h$
satisfies (\ref{adm1}).

As for (\ref{adm2}), suppose $h(X)=0$, then by the construction of $h$
either $g_1(X)=0$ or $g_2(X)=0$ (or both). The two cases are identical.
We consider only the case $g_1(X)=0$. By (\ref{g1out}), $\max_{Y \in Att(X)}\{g_1(Y)\}=1$,
and hence for some $Y \in Att(X)$, $g_1(Y)=1$. By (\ref{h1}),
we have that $h(Y)=1$ and then $\max_{Y \in Att(X)}\{h(Y)\}=1$.

Note that $in(g_1) \subseteq in(h)$, $out(g_1) \subseteq out(h)$,
$in(g_2) \subseteq in(h)$ and $out(g_2) \subseteq out(h)$.  Therefore,
since every $g_1$ and $g_2$ satisfying (\ref{g1in})--(\ref{g2out}) give rise
to a function $h$ as described, and the number of all such functions is
finite, then there exists one such $h$ that the sets $in(h)$ and
$out(h)$ are maximal. 
\end{proof}
\end{theorem}

\begin{corollary}
\label{cor:unique-maximal-admissible}
Let \tuple{S,R} be a network and $f:S \longmapsto U$ an assignment of
values to the nodes in $S$ and $h:S\longmapsto U$ the assignment such that 
the sets $in(h) \subseteq in(f)$ and $out(h) \subseteq out(f)$ are maximal and 
for every node $X\in S$:
\begin{eqnarray}
\w{If $h(X)=1$, then $\textstyle\max_{Y\in Att(X)}\{h(Y)\}=0$; and} \label{adm11}\\
\w{If $h(X)=0$, then $\textstyle\max_{Y\in Att(X)}\{h(Y)\}=1$.} \label{adm21}
\end{eqnarray}
as given by Theorem~\ref{th:unique-maximal-admissible}.
Then the set $in(h)$ is the largest admissible subset of $in(f)$ such that
{\em also} $out(h) \subseteq out(f)$.

\begin{proof} $in(h)$ is conflict-free: if you take $X \in in(h)$,
then $h(X)=1$ and then $\max_{Y\in Att(X)}\{h(Y)\}=0$. Therefore,
either $Att(X)=\varnothing$; or for all $Y \in Att(X)$, $h(Y)=0$, and hence
$Y \not \in in(h)$. 

To show that $in(h)$ is admissible, we just need to show
that if $X \in in(h)$ and $Y \in Att(X)$, then there exists $Z \in Att(Y)$, such
that $Z \in in(h)$. Assume that $X \in in(h)$ and $Y \in Att(X)$. By definition,
$h(X)=1$, and then by (\ref{adm11}), $\max_{W_x\in Att(X)}\{h(W_x)\}=0$, and hence $h(Y)=0$. 
By (\ref{adm21}), $\max_{W_y\in Att(Y)}\allowbreak \{h(W_y)\}=1$. Therefore, there exists $Z \in Att(Y)$, 
such that $Z \in in(h)$.

The fact that $in(h)$ is the largest subset of $in(f)$ subject to $out(h) \subseteq
out(f)$ comes directly from Theorem~\ref{th:unique-maximal-admissible}.
\end{proof}
\end{corollary}

\begin{remark}
Consider the following network.

\medskip

\begin{center}
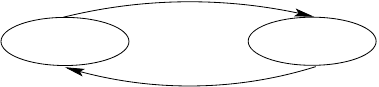
\end{center}

\medskip

There is no largest admissible subset of $E=\{X,Y\}$! There are two {\em maximal} 
admissible subsets $E_1=\{X\}$ and $E_2=\{Y\}$, so the requirement that ``no new
\exc\ nodes are generated'' is very important in 
Theorem~\ref{th:unique-maximal-admissible}. In terms of assignments (or labellings
for that matter) this was expressed as: $out(h) \subseteq out(f)$.\footnote{If we 
are given just $E$,
we may want to think of an assignment $f$ such that $in(f)=E$ and $out(f)=
\{X \; | \; E \rightarrow X\}$, leaving the nodes in $S \backslash (in(E) \cup 
out(E))$ with a value in $(0,1)$.}

If we are given an 
assignment $f(A)=1$ and $f(B)=1$, there is a class of assignments $h$ such that
the sets $in(h) \subseteq in(f)$ {\bf and} $out(h) \subseteq out(f)$ are the largest. 
For instance, $h(A)=h(B)=\frac{1}{2}$. In the example above, it is sufficient to set 
$0<h(A)<1$ and  $0<h(B)<1$  (we chose the value $\frac{1}{2}$ in 
Theorem~\ref{th:unique-maximal-admissible} simply 
because we wanted to show that one existed and because as we shall see the legal 
undecided values 
will end up converging to $\frac{1}{2}$).

Note, in particular that the assignment $f$ does not satisfy the conditions
of Theorem~\ref{th:unique-maximal-admissible} (which guarantee by 
Corollary~\ref{cor:unique-maximal-admissible} that $in(f)$ is an 
admissible set).
We could turn $f$ into an admissible assignment by just flipping one of
the values of $A$ or $B$ to $0$. However, if we did this, for instance, by
generating the assignment $f'(A)=1$ and $f'(B)=0$, then although $in(f')$ is 
admissible and $in(f') \subseteq in(f)$, we would not have that $out(f')=\{B\} 
\subseteq out(f)=\varnothing$!

This is as it should be, because an initial assignment $f$ encodes not only
which nodes we would like to be \inc, but also those that we would like to be
\exc, and we cannot decide without further information to optimise on the \inc's
in detriment of the \exc's. 
\end{remark}

\begin{theorem}
\label{th:gr-down-admissible}
Let $\net{N}=\tuple{S,R}$ be a network and \eq{T} its GR system
of equations.
If the sequence of values $V_0$, $V_1$, \ldots becomes stable at
iteration $k$, then $in(V_k)$ is the largest admissible set
such that $in(V_k) \subseteq in(V_0)$ and $out(V_k) \subseteq
out(V_0)$.

\begin{proof}
We first show that $in(V_k)$ is an admissible set.
\begin{enumerate}
\item Suppose $in(V_k)$ is not conflict-free. Therefore, there must exist
$X,Y \in in(V_k)$, such that $(Y,X) \in R$. Since $X,Y \in in(V_k)$, $V_k(X)=
V_k(Y)=1$. $V_{k+1}(X)=\max\left\{\nicefrac{1}{2},1 -\max _{Y \in Att(X)}V_k(Y)\right\}=
\nicefrac{1}{2}$, and then the sequence is not stable at $k$, a
contradiction. Therefore, $in(V_k)$ is conflict-free.
\item Suppose $in(V_k)$ is not admissible. It follows that
there exists $X \in in(V_k)$ and some $Y \in S$ with $(Y,X) \in R$, such
that $in(V_k) \not \rightarrow Y$. Since $X \in in(V_k)$, then $V_k(X)=1$
and since the sequence is stable at $k$, $V_{k+1}(X)=1=\max\left\{\nicefrac{1}{2},
1 -\max_{W\in Att(X)}V_k(W)\right\}$. Therefore,
$\max_{W\in Att(X)}V_k(W)\allowbreak =0$. In particular, $V_k(Y)=0$, and hence $V_{k+1}(Y)=
\min\left\{\nicefrac{1}{2},1-\right.$\linebreak $\left.\max_{Z\in Att(Y)}\allowbreak V_k(Z)\right\}=0$, and therefore
there exists $Z \in Att(Y)$, such that $V_k(Z)=1$, and hence $Z\in in(V_k)$,
and hence $in(V_k) \rightarrow Y$, a contradiction.  Therefore, $in(V_k)$ is
admissible.
\end{enumerate}
Now we need to show that $in(V_k)$ is indeed the maximal admissible
set such that $in(V_k) \subseteq in(V_0)$ and $out(V_k) \subseteq
out(V_0)$. By Theorem~\ref{th:unique-maximal-admissible}, there are unique
maximal sets $in(V_{max}) \subseteq in(V_0)$ and $out(V_{max}) \subseteq
out(V_0)$ such that $in(V_{max})$ is admissible. Furthermore,
$in(V_{max}) \supseteq in(V_k)$
and $out(V_{max}) \supseteq out(V_k)$. Suppose either $in(V_k)$ or
$out(V_k)$ are not maximal and let $0< j < k$ be the first index
such that there is some $X \in in(V_{max})$, such that $X \not \in in(V_j)$
or that there is some $Y \in out(V_{max})$ such that $Y \not \in out(V_j)$
(or both). We start with the first case. Since $X \in in(V_{max})$, then
$X \in in(V_{j-1})$ and hence $V_{j-1}(X)=1$. Since $X \not \in
in(V_j)$, then $V_j(X) < 1$. It follows that
$V_{j}(X)=\max\left\{\nicefrac{1}{2},1 - \max_{Y\in Att(X)}V_{j-1}(Y)\right\} < 1$.
Therefore, there exists $Y \in Att(X)$, such that $V_{j-1}(Y)>0$
and hence $Y \not \in out(V_{j-1})$. Since $in(V_{max})$ is admissible,
$Y \in out(V_{max})$ and this is a contradiction with the fact
that $j$ was the first index such that there was some $Y \in out(V_{max})$
such that $Y \not \in out(V_j)$ .

The second case is analogous.
Take $Y \in out(V_{max})$ such that $Y \not \in out(V_j)$. Since $Y \in
out(V_{max})$, then $Y \in out(V_{j-1})$ and hence $V_{j-1}(Y)=0$. Since $Y
\not \in out(V_j)$, then $V_j(Y) > 0$. It follows that $V_{j}(Y)=
\min\{\nicefrac{1}{2},1 - \max_{Z\in Att(Y)}\allowbreak V_{j-1}(Z)\} > 0$. Therefore, for
all $Z \in Att(Y)$ we have that $V_{j-1}(Z) < 1$ and hence there is no
$Z \in Att(Y)$, such that $Z \in in(V_{j-1})$. Since $Y \in
out(V_{max})$, there must be some $Z^{\prime} \in Att(Y)$, such that
$Z^{\prime} \in in(V_{max})$, but this is a contradiction since
$Z^{\prime}\not \in in(V_{j-1})$ and $j$ was the first index such
that there was some $X \in in(V_{max})$, such that $X \not \in in(V_j)$.
\end{proof}
\end{theorem}

\begin{remark}
Given an argumentation network $\net{N}=\tuple{S,R}$, an \df\ function $g$, 
a system of equations \eq{T} written for $\net{N}$ using $g$, and an assignment
$v: S\longmapsto U$, which represents initial desired values, then
if $v$ corresponds to a complete extension then the above theorems
tell us that the sequence of equations $V_0 =v$, $V_1$, $V_2$,\ldots
will become stable at some iteration $k$ and $V_k =v$. Otherwise, 
$V_k$ is the function giving the maximal possible crisp 
part $in(V_k)$ and $out(V_k)$ agreeing with $v$ such that
the set $in(V_k)$ is admissible. We 
now have the option of extending $in(V_k)$ into a complete 
extension $E_{comp}$ that is the closest extension agreeing
with $in(v)$. If this extension is also preferred, then it would correspond 
to an assignment $f^{\prime}$, which solves the original system of equations
 \eq{T} (by Theorem~\ref{th:existsSol}). If the extension is not preferred,
 then whether such an $f^{\prime}$ exists depends on the nature
 of the function $g$. Some such functions, such as $\min$ can always
 find an $f^{\prime}$ for every complete extension. Others, such as
 product, can not always find them.\footnote{Product is given in Item 2. 
 of Example~\ref{ex:1.1}. For the network $S=\{A,B\}$, $R=\{(A,B),(B,A),(B,B)\}$
 and the complete extension  ``all undecided'', there is no solution using product.}
 
We will see that with the Gabbay-Rodrigues Iteration Schema, if we continue 
iterating, in the limit of the sequence, we will get an extension.
\end{remark}

The following definition helps to translate between values in
$U$ and values in $\{\inc,\exc,\und\}$.

\begin{definition}[Caminada-Pigozzi/Gabbay-Rodrigues Translation]
\label{def:CP-GR-translation}
A labelling function $\lambda$ and a valuation function $V$ can be
inter-defined according to the table below.

\begin{center}
\begin{tabular}{ccc|ccc}
\hline
$\lambda(X)$ & $\rightarrow$ & $V_{\lambda}(X)$ &
$V(X)$ & $\rightarrow$ & $\lambda_V(X)$ \\ \hline \hline
\inc & $\rightarrow$ & $1$ & $1$ & $\rightarrow$ & \inc \\
\exc & $\rightarrow$ & $0$ & $0$ & $\rightarrow$ & \exc \\
\und & $\rightarrow$ & $\nicefrac{1}{2}$ & $(0,1)$ & $\rightarrow$ & \und \\ \hline
\end{tabular}
\end{center}
\end{definition}

The choice of the value $\nicefrac{1}{2}$ in the translation from
\und\ is arbitrary. Any value in $(0,1)$ would do, but we will see that
legal undecided values will converge to $\nicefrac{1}{2}$ in the limit, and
so $\nicefrac{1}{2}$ is the natural choice.

\begin{definition}
A {\em legal assignment $V$} is an assignment of values $V:S \longmapsto U$
such that the corresponding labelling function $\lambda_V$ defined according
to Definition~\ref{def:CP-GR-translation} is also legal.
\end{definition}

\begin{proposition} 
Let $\lambda$ be a labelling function and $V_\lambda$ its corresponding
Caminada-Pigozzi translation. If the Gabbay-Rodrigues Iteration Schema 
is employed using $V_\lambda$ as $V_0$, then for some value $k\geq 0$, the 
sequence of values $V_0$, $V_1$, \ldots will become stable and the sets 
$in(V_{k})$ and $out(V_k)$ will correspond to the down-admissible labelling 
of $\lambda$.

\begin{proof} This follows directly from Theorem~\ref{th:gr-down-admissible} 
and Corollary~\ref{cor:max-iter}.
\end{proof}
\end{proposition}

We may also arbitrarily start with $V_0(X)=1$ for all nodes $X\in S$ and see if 
this assignment satisfies the equations. At each iteration, the equations may 
force the crisp values of some nodes to turn to \und. Eventually, some
iteration $k \leq |S|$ will produce the last set of new undecided values, 
at which point we say that the sequence has stabilised. We have that
$in(V_k)$ and $out(V_k)$ correspond to the largest admissible labelling 
such that $in(V_k) \subseteq in(V_0)$ and $out(V_k)\subseteq out(V_0)$. 
$in(V_k)$ can now form the basis of a complete extension. 
The  smallest of such (complete) extensions comes from what Caminada 
and Pigozzi called the {\em up-complete labelling of} $\lambda_{V_k}$:

\begin{definition}[\cite{Caminada-Pigozzi:11}] Let $\lambda$ be an 
admissible labelling. The {\em up-complete labelling of $\lambda$} is
a complete labelling $\lambda ^{\prime}$ s.t. $in(\lambda^{\prime}) 
\supseteq in(\lambda)$ and $out(\lambda^{\prime})\supseteq out(\lambda)$ 
and $in(\lambda^{\prime})$ and $out(\lambda^{\prime})$ are the smallest
sets satisfying these conditions.
\end{definition}

If we continue with our calculations we can see what happens with the 
values $V_0$, $V_1$,\ldots,$V_i$, \ldots in the limit 
of the sequence. We cal these the {\em equilibrium values}. Formally,

\begin{definition} Let $\net{N}=\tuple{S,R}$ be an argumentation network, 
\eq{T} its GR system of equations, and $V_0$ an assignment of initial values 
to the nodes in $S$. The {\em equilibrium value} of the node $X$ is defined as
$\f{X}=\lim_{i\rightarrow \infty} V_i(X)$.
\end{definition}

The understanding of the meaning of the equilibrium values requires an 
analysis of the behaviour of the sequence. The value of a node $X$ is 
essentially determined by the values of the nodes in $Att(X)$. At the 
stable point $k$ we know that the crisp values remain crisp. The values
of the attackers of a node at the stable point $k$ can be of one
of three types:
\begin{enumerate}
\item $\max_{Y \in Att(X)} \{V_k(Y)\} =0$
\item $\max_{Y \in Att(X)} \{V_k(Y)\} =1$
\item\label{case:harder} $0 < \max_{Y \in Att(X)} \{V_k(Y)\}<1$
\end{enumerate}

If the value of a node $Y$ at the stable point $k$ is in $\{0,1\}$, 
then Theorem~\ref{th:in-out-stability} ensures that it will remain the same
in the limit $\lim_{i\rightarrow\infty} V_i(Y)$. As it turns out, if 
$\max_{Y \in Att(X)} \{V_k(Y)\} =0$, then $\lim_{i\rightarrow\infty} V_i(X)=1$. 
And if $\max_{Y \in Att(X)} \{V_k(Y)\} =1$, then $\lim_{i\rightarrow\infty} V_i(X)=0$, 
as shown by the next theorem.

\begin{theorem}
\label{th:convergence-one}
Let $\net{N}=\tuple{S,R}$ be an argumentation network and $V_0:S
\longmapsto U$ assign initial values to the nodes in $S$. Let the
sequence of value assignments $V_0$, $V_{1}$, $V_2$, \ldots where
each $V_i$, $i>0$, is generated by the Gabbay-Rodrigues Iteration
Schema be stable at iteration $k$. For every $X\in S$:

\begin{enumerate}
\item If $\max_{Y\in Att(X)} \{V_k(Y)\} =0$, then $\f{X}=1$;
and
\item If $\max_{Y\in Att(X)} \{V_k(Y)\} = 1$, then $\f{X}=0$.
\item If $V_k(X)\in\{0,1\}$, then $\f{X}=V_k(X)$;
\end{enumerate}

\begin{proof}
\begin{enumerate}
\item If $\max_{Y \in Att(X)} V_k(Y) =0$, and the sequence is stable at $k$,
then by Corollary~\ref{cor:in-out-stability}, $\max_{Y \in Att(X)} V_{k+j}(Y)
=0$, for all $j\geq 0$. We have that
\begin{align}
V_{k+1}(X) & = (1 - V_k(X))\cdot \min\left\{\frac{1}{2},1\right\} + V_k(X) \cdot
\max \left\{\frac{1}{2},1\right\}\nonumber\\
& = \frac{1}{2} -\frac{V_k(X)}{2} + V_k(X) = \frac{1}{2} +\frac{V_k(X)}{2} \nonumber\\
V_{k+2}(X) & = \frac{1}{2} + \frac{1}{4} + \frac{V_{k}(X)}{4} \nonumber\\
V_{k+j}(X) & = \sum_{k=1}^{j}\frac{1}{2^k} + \frac{V_{k}(X)}{2^j} \nonumber\\
\f{X} & = \lim_{j\rightarrow\infty} V_{k+j}(X) \nonumber\\
& = \sum_{k=1}^{\infty}\frac{1}{2^k} + \lim_{j\rightarrow\infty}\frac{V_{k}(X)}{2^j} = 1 +0 = 1\nonumber
\end{align}

So if the maximum value $m_k$ of all attackers of $X$ at iteration
$k$ is $0$, then the value of $X$ converges to $1$; and finally

\item If $\max_{Y \in Att(X)} V_k(Y) =1$, and the sequence is stable at $k$,
then by Corollary~\ref{cor:in-out-stability}, $\max_{Y \in Att(X)} V_{k+j}(Y)
=1$, for all $j\geq 0$. We have that
\begin{align}
V_{k+1}(X) & = (1 - V_k(X))\cdot \min\left\{\frac{1}{2},0\right\} + V_k(X) \cdot
\max \left\{\frac{1}{2},0\right\}\nonumber \\
& = \frac{V_k(X)}{2}\nonumber\\
V_{k+2}(X) & = \frac{V_{k}(X)}{4}\; \therefore \;
V_{k+j}(X) = \frac{V_k(X)}{2^j} \nonumber\\
\f{X} & = \lim_{j\rightarrow\infty} V_{k+j}(X) = \lim_{j\rightarrow\infty}
\frac{V_{k}(X)}{2^j} = 0\nonumber
\end{align}

So if the maximum value $m_k$ of all attackers of $X$ at iteration
$k$ is $1$, then the value of $X$ converges to $0$.
\item This follows from the fact that the sequence is stable at $k$;
\end{enumerate}
\end{proof}
\end{theorem}

\begin{figure}[htb]
\begin{center}
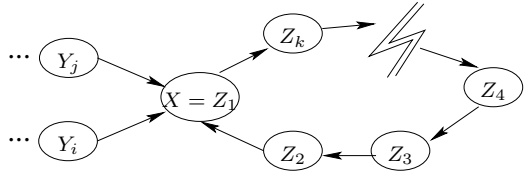  
\end{center}

\caption{A network with a cycle with $k$ nodes.\label{fig:scc-values}}
\end{figure}

The theorem above asserts self-correction for the values of nodes
whose attackers are either all $\exc$ or that have an attacker that
is $\inc$. Case~\ref{case:harder} above, in which $0 < \max_{Y \in Att(X)} 
\{V_k(Y)\}<1$, is harder and will be dealt with in stages. We start with 
the case of a cycle whose values of the nodes 
are all in $(0,1)$ (see Figure~\ref{fig:scc-values}). Such cycles may involve 
an even or odd number of nodes, so we have chains of attacks of one of the 
following types:
\begin{itemize}
\item either $X=Z_1 \leftarrow Z_2 \leftarrow \ldots \leftarrow Z_{2n}=X$ (even cycle)
\item or $X=Z_1 \leftarrow Z_2 \leftarrow \ldots \leftarrow Z_{2n+1}=X$ (odd cycle)
\end{itemize}
The next lemma shows that in either case, the value of $X$ in the limit
is $\frac{1}{2}$.

\newcommand{\fc}[1]{\w{$V^c_e(#1)$}}
\begin{theorem}
\label{th:convergence-to-half}
Let the sequence of values $V_0$, $V_1$, \ldots, be stable at iteration $k$.
Let $X$ be a point such that $V_{k+i}(X),V_{k+i+1}(X),\ldots \in (0,1)$, for all
$i\geq 0$. Our final aim is to show that $\lim_{i\rightarrow \infty} V_{k+i}(X)=\frac{1}{2}$.
As a first step towards our goal, we show that any converging subsequence
$V^{c}_{k+j}(X)$ converges to $\frac{1}{2}$ (by a subsequence $V^{c}_{k+j}(X)$ we
mean some of the elements of the  sequence $V_{k+i}(X)$, that is for every $j$
there is an $i_j$ such that $V^{c}_{k+j}(X)=V_{k+i_j}(X)$). From now on we talk 
about the subsequence $V^{c}_{k+j}(X)$, which we further assume that it converges to 
$V^{c}_e(X)$, for every such $X$, and we will show that $V^{c}_e(X)=\frac{1}{2}$,
for every $X$.

To be absolutely clear we assume for the time being that there is a sequence of values 
$s_1$, $s_2$, $s_3$, 
\ldots, such that for every $X$, the sequence $V^c_{s_i}(X)$ converges to $V^c_{e}(X)$
and we show that under these conditions $V^{c}_e(X)=\frac{1}{2}$.

Consider all possible cycles $X=Z_1 \leftarrow Z_2 \leftarrow \ldots
\leftarrow Z_{2n}=X$ (even) and $X=Z_1 \leftarrow Z_2 \leftarrow \ldots \leftarrow
Z_{2n+1}=X$ (odd) and assume that amongst them we have a cycle such that
there exists a sequence of values $r_1$, $r_2$, \ldots such that for each 
$Z_i$, $Z_{i+1}$ is the node in $Att(Z_i)$ with maximum
value and $0 < V^c_{k+r_1+r_2+\ldots+r_m}(Z_i) < 1$, for every $m \geq 0$. Then
$\fc{Z_i}=\frac{1}{2}$, for all $Z_i$.

\medskip
\begin{proof}
Since the Gabbay-Rodrigues Iteration Schema uses continuous functions, if the schema holds
for the elements of the sequence $V^c_{k+j}(X)$, for every $X \in S$, it also holds for
the limit $V^c_e(X)$.

 We get the following systems of equations
\begin{enumerate}
\item For the cycle
$X=Z_1 \leftarrow Z_2 \leftarrow \ldots \leftarrow Z_{2n}=X$:

$\fc{X}=(1-\fc{X}) \cdot \min\left\{\frac{1}{2},1-\fc{Y}\right\} + \fc{X} \cdot \max\left\{\frac{1}{2},
1 - \fc{Y}\right\}$, where $Y$ is the node in $Att(X)$ with maximum value. We have two
cases to consider. 
\begin{itemize}
\item $\fc{Y}\geq \frac{1}{2}$, then we get that
\[\fc{X}=\frac{1-\fc{Y}}{1.5 - \fc{Y}}\]
\item $\fc{Y} \leq \frac{1}{2}$, the we get that
\[ \fc{X} =\frac{1}{1 + 2\cdot \fc{Y}}\]
\end{itemize}
it is easy to see from the equations that if $\fc{Y}\geq \frac{1}{2}$, then
$\fc{X}\leq\frac{1}{2}$ and if $\fc{Y} \leq\frac{1}{2}$, then $\fc{X} \geq \frac{1}{2}$.
Therefore, if we have the cycle
$X=Z_1 \leftarrow Z_2 \leftarrow \ldots \leftarrow Z_{2n}=X$, then
we get that $\frac{1}{2} \leq Z_1 \leq \frac{1}{2}$, so all $Z_i=
\frac{1}{2}$.

\item For the cycle $X=Z_1 \leftarrow Z_2 \leftarrow \ldots \leftarrow
Z_{2n+1}=X$, we have that
\begin{itemize}
\item either $\fc{Y}\geq \frac{1}{2}$. Let us write $\fc{Y}=\frac{1}{2}+\epsilon(Y)$,
for some $0 \leq \epsilon(Y) < \frac{1}{2}$. We then get that
\begin{eqnarray*}
\fc{X} & = & \frac{1-\fc{Y}}{1.5 - \fc{Y}} \\
  & = & \frac{1 - \frac{1}{2} - \epsilon(Y)}{1.5 - (\frac{1}{2}+\epsilon(Y))}\\
  & = & \frac{\frac{1}{2} - \epsilon(Y)}{1 -\epsilon(Y)}
\end{eqnarray*}
Write $\fc{X} = \frac{1}{2}-\eta$, for some $0 < \eta < \frac{1}{2}$.
\[\frac{1}{2} - \eta = \frac{\frac{1}{2} - \epsilon(Y)}{1 -\epsilon(Y)}\]
\begin{eqnarray*}
\eta & = \frac{1}{2}-\frac{\frac{1}{2} - \epsilon(Y)}{1 -\epsilon(Y)}\\
     & = \frac{(1 \epsilon(Y)) - 2(\frac{1}{2}-\epsilon(Y))}{2(1-\epsilon(Y)}\\
     & = & \frac{1-\epsilon(Y)-1 + 2\epsilon(Y)}{2(1-\epsilon(Y))}\\
     & = & \frac{\epsilon(Y)}{2(1-\epsilon(Y))}\\
\end{eqnarray*}

\item or $\fc{Y}\leq \frac{1}{2}$. Let us write $\fc{Y}=\frac{1}{2}-\epsilon(Y)$,
for some $0 \leq \epsilon(Y) < \frac{1}{2}$. We then get that
\begin{eqnarray*}
\fc{X} & = & \frac{1}{1 + 2(\frac{1}{2} - \epsilon(Y)} \\
  & = & \frac{1}{1 + 1 - 2\epsilon(Y)}\\
  & = & \frac{1}{2(1 - \epsilon(Y))}\\
  & = & \frac{1}{2} + \eta
\end{eqnarray*}
\begin{eqnarray*}
\eta & = & \frac{1}{2(1 - \epsilon(Y))}-\frac{1}{2}\\
  & = & \frac{1-1+\epsilon(Y)}{2(1 - \epsilon(Y))}\\
  & = & \frac{\epsilon(Y)}{2(1 - \epsilon(Y))}\\
\end{eqnarray*}
\end{itemize}
Where are we now? We saw that if we start from $\fc{Y} = \frac{1}{2}
\pm \epsilon(Y)$ and $Y \rightarrow X$ ($Y$ attacks $X$ as in a
cycle), then $\fc{X}=\frac{1}{2} \pm \eta$, where $\eta$ is in the
other direction and
\[\eta=\frac{\epsilon(Y)}{2(1 - \epsilon(Y))}.\]
Let us now assume a cycle
\[X=Z_1 \leftarrow Z_2 \leftarrow \ldots \leftarrow Z_{n}=X\]
Assume $Z_1=\frac{1}{2}\pm \epsilon$. What would the value of
$Z_k$ be?

We claim that
\[ Z_k = \frac{1}{2} \pm \eta_k\]
where
\[ \eta_k = \frac{\epsilon}{2(2^k-(2^k-1)\epsilon)}\]

The proof is by induction. Let $X = Z_k$, then $Y = Z_{k+1}$, and then
\begin{eqnarray*}
\eta_{k+1} & = & \frac{\eta_k}{2(1 - \eta_k)}\\
          & = & \frac{\frac{\epsilon}{2(2^k-(2^k-1)\epsilon)}}{2(1-
                               \frac{\epsilon}{2(2^k-(2^k-1)\epsilon)})}\\
          & = & \frac{\frac{\epsilon}{2(2^k-(2^k-1)\epsilon)}}%
                     {2\big(\frac{2(2^k-(2^k-1)\epsilon -\epsilon)}%
                                 {2(2^k-(2^k+1)\epsilon)}\big)}\\
          & = & \frac{\epsilon}%
                     {2\big(2^{k+1}-2^{k+1}\epsilon+2\epsilon -\epsilon\big)}\\
          & = & \frac{\epsilon}%
                     {2(2^{k+1}-(2^{k+1}-1)\epsilon)}
\end{eqnarray*}
So the recursion works. Now if we have a loop, we get
\[Z_n=Z_1\]
So $\eta_n=\eta_1$ and thus
\[\eta=\frac{\eta}{2(2^{k+1}-(2^{k+1}-1)\epsilon)}\]
If we divide by $\eta$ ($\neq 0$), we get
\[1=\frac{1}{2(2^{k+1}-(2^{k+1}-1)\epsilon)}\]
It is easy to see that only $\epsilon=\frac{1}{2}$ solves the equation. This means that 
$V^c_{e}(Z_i)=\frac{1}{2}$, for all $Z_i$.
\end{enumerate}
\end{proof}
\end{theorem}

\begin{remark}\label{rem:1.1}
Ordinarily we cannot guarantee that $Z_{i+1}$ is the node in
$Att(Z_i)$ with maximum value for all $k^\prime>k$, we need
to find a subsequence. This is done as follows: we start with a node $X$
and since there are a finite number of nodes attacking it (the network is finite),
there exists a subsequence such that there is a single attacker whose
$V^c_{k^\prime}$ value is the maximum for all $k^\prime$ in the subsequence.
We can assume it is $Z_2$. This $Z_2$ is not unique, there may be other choices.
Let $Z^{\alpha_2}_2$ be one arbitrary such choice. Repeating
this consideration now for $Z^{\alpha_2}_2$ and for the subsequence thus obtained, we
get a $Z^{\alpha_3}_3$ and a further subsequence of the subsequence and so on. Eventually,
we get a final subsequence (which depends on the choices of $Z^{\alpha_i}_i$) $V^c_{k+r_1}$,$V^c_{k+r_1+r_2},
\ldots$, such that $Z^{\alpha_{i+1}}_{i+1}$ is the node in $Att(Z^{\alpha_i}_i)$ with maximum value and $0 <
V^c_{k+r_1+r_2+\ldots+r_m}(Z^{\alpha_i}_{i+1}) < 1$, for each $m$. 
\end{remark}

\begin{remark}
We use a similar argument to the one in Remark~\ref{rem:1.1} to show that if a subsequence
$V^c_{k+j}(X)$ converges to $V^c_{e}(X)$, then it can be further refined to a subsequence
$V^c_{s_i}$ such that $V^c_{s_i}(Y)$ converges for all $Y$. The reason
is that the number of such $Y$ is finite (since $S$ is finite). We can then successively
refine the sequence $V^c_{k+j}(X)$ into subsequences for which $V^c_{k+j}(Y)$ also converges.
Therefore, Theorem~\ref{th:convergence-to-half}, can be used to show that the convergent
sequence $V^c_{k+j}$ converges to $\frac{1}{2}$. We can therefore further conclude that every
convergent subsequence of $V_{k+m}(X)$ converges to $\frac{1}{2}$. The next lemma shows
that the sequence $V_{k+m}(X)$ itself converges to $\frac{1}{2}$. 
\end{remark}

\begin{lemma}\label{lem:subseqn-converge}
Let $\alpha=\alpha_1, \alpha_2, \alpha_3, \ldots$, be an infinite sequence of values in $[0,1]$.
If every convergent subsequence of $\alpha$  converges to $\frac{1}{2}$, then $\lim_{i\rightarrow
\infty}\alpha_i=\frac{1}{2}$.

\begin{proof}
For every $0 <\varepsilon < \frac{1}{2}$, $[\frac{1}{2}-\varepsilon,\frac{1}{2}+ \varepsilon]$ only 
a finite number of $\alpha_i$'s are in $[0,\frac{1}{2}-\varepsilon] \cup [\frac{1}{2}+\varepsilon,1]$. 
Otherwise, say $[0,\frac{1}{2}-\varepsilon]$ has an infinite number of $\alpha_i$'s. Then since 
$[0,\frac{1}{2}-\varepsilon]$ is a closed interval with an infinite number of values in it, there
would exist an infinite convergent subsequence of $\alpha$ in it that does not converge to 
$\frac{1}{2}$.

Therefore, we have shown that for every $0 <\varepsilon < \frac{1}{2}$, $\varepsilon$ small,
there exists a number $m$ such that for every $n > m$, $(\frac{1}{2}-\alpha_n) \in [\frac{1}{2}-
\varepsilon,\frac{1}{2}+ \varepsilon]$, that is $\lim_{i\rightarrow
\infty}\alpha_i=\frac{1}{2}$.
\end{proof}
\end{lemma}

Theorem~\ref{th:convergence-one} asserts what the limit values of the 
nodes whose values of the attackers are known at the stable iteration $k$.
Theorem~\ref{th:convergence-two} asserts the same in terms of the {\em limit}
values of the attackers.

\begin{theorem}\label{th:convergence-two}~
\begin{enumerate}
\item If $\max_{Y\in Att(X)} \{V_e(Y)\} =0$, then $V_e(X)=1$.
\item If $\max_{Y\in Att(X)} \{V_e(Y)\} =1$, then $V_e(X)=0$.
\end{enumerate}

\medskip
\begin{proof}
Note that $lim_{j\rightarrow \infty}\{V_{j+1}(X)\}=lim_{j\rightarrow \infty}\{V_{j}(X)\}$.
\begin{enumerate}
\item If $\max_{Y \in Att(X)} \{V_e(Y)\} =0$, then we have that
\begin{eqnarray*}
V_e(X) & = &
        (1 - V_e(X))\cdot \min\left\{\frac{1}{2},1\right\}
        + V_e(X) \cdot \max \left\{\frac{1}{2},1\right\}
          \nonumber\\
        V_{e}(X) & = & (1 - V_e(X))\cdot\frac{1}{2}
         + V_e(X)\nonumber\\
        2\cdot V_{e}(X) & = & 1 - V_{e}(X) + 2\cdot V_{e}(X)\nonumber\\
        V_{e}(X) & = & 1 \nonumber
\end{eqnarray*}
So if the equilibrium values of all attackers of $X$ is $0$, then the
equilibrium value of $X$ is $1$.

\item If $\max_{Y \in Att(X)} \{V_e(Y)\} =1$, then we have that
\begin{eqnarray*}
V_e(X) & = &
    (1 - V_e(X))\cdot \min\left\{\frac{1}{2},0\right\} + V_e(X) \cdot \max \left\{\frac{1}{2},0\right\}\nonumber\\
V_e(X) & = & \frac{V_e(X)}{2} \nonumber\\
V_e(X) & = & 0 \nonumber
\end{eqnarray*}
So if the equilibrium value of any of the attackers of $X$ is $1$, then the
equilibrium value of $X$ is $0$.
\end{enumerate}
\end{proof}
\end{theorem}

\begin{theorem}
\label{th:legal-to-stable}
Let \tuple{S,R} be an argumentation network and \eq{T} its GR system of 
equations. If the assignment $V_0:S\longmapsto U$ is legal then the
sequence $V_0$, $V_1$, $V_2$, \ldots, where each $V_i$, $i>0$, is generated
by \eq{T}, is stable at iteration $0$.

\begin{proof}
Suppose $V_0$ is legal. Then if $V_0(X)=0$, then there exists $Y \in Att(X)$
such that $V_0(Y)=1$. Therefore $V_1(X)=\min\left\{\nicefrac{1}{2},0\right\}=0$.
If $V_0(X)=1$, then for all $Y \in Att(X)$, $V_0(Y)=0$, and hence
$max_{Y\in Att(X)}V_0(Y)=0$. Therefore, $V_1(X)=\max\left\{\nicefrac{1}{2},1\right\}=1$.

The stability of the crisp values then follows from 
Theorem~\ref{th:in-out-stability} and since $0<V_0(X)<1$, then by 
Theorem~\ref{th:maximality} (case 3), so does the stability of the 
remaining non-crisp values.
\end{proof}
\end{theorem}

\begin{proposition}\label{prop:agree-on-attacks} Let \tuple{S,R} be an
argumentation network; \eq{T} its GR system of equations and $V_e$ a function
with the equilibrium values of the nodes in $S$ calculated according to the
Gabbay-Rodrigues Iteration Schema. Let $\lambda$ be a legal labelling function.

Take any $X \in S$. If $\lambda$ and $V_e$ agree on the values of all nodes in
$Att(X)$, then $\lambda$ and $V_e$ agree on the value of $X$.

\begin{proof} There are three cases
to consider. Proofs of cases 1. and 2. are similar to the proofs of cases
1. and 2. of Theorem~\ref{th:convergence-one}.
\begin{enumerate}
\item $\max_{Y\in Att(X)}\{V_e(Y)\}=0$, then for all $Y \in
Att(X)$, $V_e(Y)=0$. It follows that $V_e(X)=
\sum_{k=1}^{\infty}\frac{1}{2^k} + \lim_{j\rightarrow\infty}\frac{V_{k}(X)}{2^j} =
1 +0=1$. Since $V_e$ and $\lambda$ agree
with each other on the values of all nodes in $Att(X)$, we have
that for all $Y \in Att(X)$, $\lambda(Y)=\exc$ and since $\lambda$ is legal,
$\lambda(X)=\inc$, and hence $\lambda$ and $V_e$ agree with each other with
respect to the value of $X$ as well.
\item $\max_{Y\in Att(X)}\{V_e(Y)\}=1$, then there exists $Y \in
Att(X)$, such that $V_e(Y)=1$. It follows that $V_e(X)=\lim_{j\rightarrow
\infty}\frac{V_e(X)}{2^j}=0$. Since $V_e$ and $\lambda$ agree
with each other on the values of all nodes in $Att(X)$, we have
that $\lambda(Y)=\inc$ and since $\lambda$ is legal, $\lambda(X)=\exc$.
Hence $\lambda$ and $V_e$ agree with each other with respect to the
value of $X$ as well.
\item $\max_{Y\in Att(X)}\{V_e(Y)\}=\frac{1}{2}$, then there exists $Y \in
Att(X)$, such that $V_e(Y)=\frac{1}{2}$ (and hence $\lambda(Y)=\und$) and
for no $Y \in Att(X)$, $V_e(Y)=1$ (and hence for no $Y \in Att(X)$, $\lambda(Y)
=\inc$). It follows that
\begin{align*}
V_e(X)= & \frac{1-V_e(X)}{2}+\frac{V_e(X)}{2}\\
2\cdot V_e(X) = & 1 \\
V_e(X) = & \frac{1}{2}
\end{align*}

Since $\lambda$ is legal, $\lambda(X)=\und$, and
hence $\lambda$ and $V_e$ agree with each other with respect to the
value of $X$.
\end{enumerate}
\end{proof}
\end{proposition}

And now to the main theorem of this section, which explains 
the equilibrium values of all nodes and shows their relationship to
Caminada and Pigozzi's down-admissible/up-complete constructions.
A down-admissible labelling is obtained after a series of {\em 
contraction operations} as defined below.

\begin{definition}[\cite{Caminada-Pigozzi:11}] Let $\lambda$ be a
labelling of an argumentation network \tuple{S,R}. A {\em contraction
sequence from $\lambda$} is a sequence of labellings $[\lambda_1=\lambda,
\ldots \lambda_k]$ such that
\begin{enumerate}
\item For each $i\in\{1,\ldots,k-1\}$, $\lambda_{i+1}=\lambda_{i}-\{(X,\inc),
(X,\exc)\}\cup\{(X,\und)\}$, where $X$ is an argument that is illegally
labelled \inc, or illegally labelled \exc\ in $\lambda_j$; and
\item $\lambda_k$ is a labelling without any arguments illegally
labelled \inc\ or illegally labelled \exc.
\end{enumerate}
\end{definition}

Theorem~6 of \cite{Caminada-Pigozzi:11} shows us that
if we successively contract an initial labelling $\lambda$, then
at the end of the contraction sequence $[\lambda_1=\lambda,\lambda_2,\allowbreak
\ldots \lambda_k]$, $\lambda_k$ corresponds to the down-admissible
labelling of $\lambda$, which is the largest admissible labelling that 
is smaller or equal to $\lambda$. 

Not every admissible labelling corresponds to a complete extension. 
However, an admissible labelling can be turned into a labelling that corresponds
to a complete extension by changing the labels of nodes that illegally 
labelled \und, to \inc\ or \exc\ as appropriate. Each such operation
is called an {\em expansion}, and an expansion sequence corresponds
to a list of all such operations:

\begin{definition}[\cite{Caminada-Pigozzi:11}] Let $\lambda$ be an
admissible labelling of the argumentation network \tuple{S,R}. An {\em
expansion sequence from $\lambda$} is a sequence of labellings $[\lambda_1=
\lambda,\ldots \lambda_k]$ such that
\begin{enumerate}
\item For each $i\in\{1,\ldots,k-1\}$,
\[\lambda_{i+1}=\left\{\begin{array}{ll}
\lambda_{i}-\{(X,\und)\}\cup\{(X,\inc)\}, & \w{if $X$ is an argument that
is}\\
\multicolumn{2}{l}{\w{\hspace*{3mm}illegally labelled \und\ in $\lambda_i$ and all its attackers are labelled \exc}}\\
\lambda_{i}-\{(X,\und)\}\cup\{(X,\exc)\}, & \w{if $X$ is an argument that
is}\\
\multicolumn{2}{l}{\w{\hspace*{3mm}illegally labelled \und\ in $\lambda_i$ and it has an attacker labelled
\inc}}\\
\end{array}
\right.\]
\item $\lambda_k$ is a labelling without any arguments illegally
labelled \und.
\end{enumerate}
\end{definition}

Caminada and Pigozzi have shown us that if $[\lambda_1=\lambda,\ldots 
\lambda_k]$ is an expansion sequence,\footnote{Note $\lambda_1$ must be 
admissible.} then $\lambda_k$ is a complete 
labelling and it is the smallest such labelling containing $\lambda$.
We now introduce a few concepts to help us in the proof of our main
theorem.

\begin{definition}
\label{def:agree}
Let \tuple{S,R} be an argumentation network; $V$ be an
assignment of values to the nodes in $S$; and $\lambda$ a labelling
of these nodes. We say that $V$ and $\lambda$ agree with each other
with respect to the value of a node $X$ if and only if the following
conditions hold:

\begin{enumerate}
\item $V(X)=1$ if and only if $\lambda(X)=\inc$
\item $V(X)=0$ if and only if $\lambda(X)=\exc$
\item $V(X)=\nicefrac{1}{2}$ if and only if $\lambda(X)=\und$
\end{enumerate}

We say that $V$ and $\lambda$ agree with each other if they agree with
the values of all nodes in $S$.
\end{definition}

\begin{definition}[Attack tree of a node] Let \tuple{S,R} be a network.
The attack tree $tree(X)$ of a node $X \in S$ is the tree with root $X$ and for
every node $N$ in $Tree(X)$, the children of  $N$ are the nodes in
$Att(N)$.
\end{definition}

\begin{definition}[Path from a node] Let \tuple{S,R} be a network.
Take $X \in S$. A path from $X$ is a sequence of nodes $X=Z_0$, $Z_1$, $Z_2$,
\ldots\ such that each $Z_{i+1}$, $i\geq 0$, is a child of $Z_i$ in the attack tree of $X$.
The set of all paths from a node $X$ is denoted $\Pi(X)$. We allow for a single
node to be a path.
\end{definition}

Using paths, we can define a {\em strongly connected component} (SCC) to be a maximal subset 
$C \subseteq S$, such that for every $X,Y \in C$, there exists a path from $X$ containing $Y$. 


Note that in a SCC $C$ for every path $\pi=Z_0,Z_i,\ldots$ from every node $Z_0
\in C$, there exists a smallest $i(\pi)$ such that for some $r(\pi)$, $Z_{i(\pi)}=Z_{i(\pi)+
r(\pi)}$. $i(\pi) < |C|$.  $i(\pi)$ is the index of the first node in the path $\pi$ that is
involved in a loop, or you can think of it as the minimum distance from the starting
node of the path $\pi$ to a looping node in the path. If $i(\pi)=0$, then $Z_0$
attacks itself. Let us call the {\em loop head} of the path $\pi=Z_0,Z_1,
\ldots$, the node $Z_{i(\pi)}$.

\begin{definition}[$V_{\max}$-paths] Let $Z$ be a node in a SCC $C$ and let
the sequence of values $V_0$, $V_1$, \ldots be stable at iteration $k$. The set of
$V_{\max}$-paths of $Z$ is defined as $V_{\max}$-paths$(Z)=\{\pi=[Z=Z_0,Z_1,\dots]
\in \Pi(Z)\; | \; \ws{for each} Z_{i}, \linebreak V_{k+r}(Z_{i+1})=\max_{Z_{i+1}^{\prime}}\{V_{k+r}(Z^{\prime}_{i+1})\} \allowbreak\w{~for an infinite number of $r$'s}\}$.
\end{definition}

For every $Z\in C$, the set of $V_{\max}$-paths from $Z$ is non-empty (see
Remark~\ref{rem:1.1}).

\begin{definition}[Bar of a node] Let $C$ be a SCC
and take $X\in C$. The {\em bar of $X$} is the set
\[bar(X)=\{ Z \in C \; |\; Z \ws{is the loop head of a path in} 
V_{\max}\w{-paths}(X)\}.\]
\end{definition}

\begin{definition}
Let $\Gamma(X)$ be the set of $V_{\max}$-paths of $X$ and take $U\subseteq C$
a set of points. The bar of $X$ modified by $U$ is defined as 
\[bar(X,U)=\bigcup_{\pi\in \Gamma(X)} \left\{\begin{tabular}{r|ll}
$y$ & \parbox[t]{7cm}{$y$ is the first node in $\pi$ such that either $y$ 
                         is\\ the loop head of $\pi$ or $y \in U$}
\end{tabular}
\right\}\]
\end{definition}

\begin{theorem}
\label{th:main-theorem}
Let \tuple{S,R} be an argumentation network; $V_0$ be an initial
assignment of values to the nodes in $S$; $\lambda_0$ an initial labelling
of these nodes; and $V_0$ and $\lambda_0$ faithful to each other according to
Definition~\ref{def:CP-GR-translation}. Let  $\lambda_{da}$ be the labelling
at the end of a contraction sequence from $\lambda_0$ and $\lambda_{CP}$ the 
labelling at the end of an expansion sequence after $\lambda_{da}$. Let
$k$ be the point at which the sequence $V_0$, $V_1$,\dots becomes
stable and $\f{X}$ the equilibrium value of a node calculated
through the Gabbay-Rodrigues Iteration Schema.
Then $\lambda_{CP}$ and $V_e$ agree with each other according to 
Definition~\ref{def:agree}.

\begin{proof}
The proof is done on induction on the depth of a node $X$. Suppose the depth of $X$ is $0$. There are three main cases to consider.
\begin{itemize}
\item[Case 1:] $X$ is a source node.
By definition, $X$ has no attackers, and hence \linebreak $\max_{Y\in Att(X)}
\allowbreak V_0(Y)= \max_{Y\in Att(X)}V_k(Y)=0$ and then by\linebreak
Theorem~\ref{th:convergence-one}, $\f{X}=1$.

If $\lambda_0(X)=\inc$, then $X$ is legally labelled \inc, $X$ does not
take part in the contraction or expansion sequences and therefore $\cp{X}=
\inc$. If $\lambda_0(X)=\exc$, then $X$ is illegally labelled \exc, and
therefore the label of $X$ is changed to \und\ in the contraction sequence
and since it is illegally labelled \und, then it is subsequently changed
to $\inc$ in the expansion sequence. If $\lambda_0(X)=\und$, then $X$
cannot be contracted, and since it is illegally labelled \und, its label
must be changed to \inc\ during the expansion sequence. In all cases, $\cp{X}=
\inc$, and hence $\lambda_{CP}$ and $V_e$ agree with each other with respect
to the value of $X$.

\item[Case 2:] $X$ is part of a source SCC $C$ and both $V_0\rest C$ and
$\lambda_0\rest C$ are {\em legal assignments} within $C$. Let us partition
$C$ into two components: $C^c$ containing all nodes with crisp values and
$C^u$ containing all nodes with undecided values.

Since $\lambda_0\rest C$ is a legal assignment, and the nodes
in $C^c$ only have values in $\{\inc,\exc\}$, then no nodes in $C^c$ are
illegally labelled and hence their labels are unaffected by the contraction
sequence. Likewise, since no node is labelled undecided in $C^c$, nothing can
be subsequently expanded and $\lambda_{CP}\rest C^c=\lambda_0\rest C^c$. By
construction, the values of all nodes in $C^u$ are \und, and hence these
nodes are not affected by the contraction sequence. Furthermore, they are
all legally labelled undecided and hence the values remain unchanged, and
hence $\lambda_{CP} \rest C=\lambda_0\rest C$.

Since $V_0\rest C$ is a legal assignment, then by
Theorem~\ref{th:legal-to-stable}, it is stable at iteration $0$. As a result,
for all nodes $X \in C^c$, $V_1(X)=V_0(X)$. Hence by
Theorem~\ref{th:in-out-stability}, $\f{X}=V_0(X)$ for all nodes $X \in C^c$,
and then since $\lambda_0$ and $V_0$ are faithful to each other
(Definition~\ref{def:CP-GR-translation}), conditions 1. and 2. of
Definition~\ref{def:agree} are satisfied. We now show that condition 3.
also follows. For all nodes $X \in C^u$, we have that $0 < V_0(X)<
1$. Since $V_0\rest C$ is a legal assignment, then for every $X \in C^u$,
$0 < max_{Y\in Att(X)}\{V_0(Y)\} < 1$.\footnote{This effectively means that
the only possible incoming attacks from $C^c$ are from nodes labelled
\exc. Otherwise, the attacked nodes in $C^u$ should have been labelled
$\exc$ and hence would have been illegally labelled \und.}
Notice that by construction $C^u=C\backslash C^c$. Stage two of case 3 
below shows that for all nodes $X \in C^u$, $\f{X}=\nicefrac{1}{2}$. 
Therefore, condition 3. of
Definition~\ref{def:agree} is also satisfied and as a result, 
$\lambda_{CP}$ and $V_e$ agree with each other with respect to
all nodes in $C$.

\item[case 3:] $X$ is part of a source SCC $C$ and $\lambda_0\rest C$ and
$V_0\rest C$ are not legal assignments.

\noindent{\bf Stage one:}

We know that the sequence of assignments $V_0$, $V_1$,\ldots,
eventually becomes stable at some iteration $k$ and by
Theorem~\ref{th:gr-down-admissible}, $in(V_k)\subseteq in(V_0)$,
$out(V_k) \subseteq out(V_0)$ and $in(V_k)$ is the
largest admissible subset of $in(V_0)$. By Theorem~6 of
\cite{Caminada-Pigozzi:11}, $in(\lambda_{CP})$ is the
largest (and unique) admissible subset of $in(\lambda_0)$
and since $\lambda_0$ and $V_0$ are faithful to each other, we
can conclude that $in(V_k)=in(\lambda_{da})$ and $out(V_k)=out
(\lambda_{da})$.

Note that since the sequence is stable at $k$, $in(V_k) \subseteq
in(V_e)$ and $out(V_k) \subseteq out(V_e)$.

Consider the sequence of expansion operations $e_1$, $e_2$, \ldots,
$e_m$ and the sequence of labellings $\lambda_0=\lambda_{da},\lambda_1,
\lambda_2$,\ldots,$\lambda_m=\lambda_{CP}$, where for each $i>0$,
$\lambda_i$ is obtained from $\lambda_{i-1}$ via the expansion $e_i$.
We show by induction on $m$ that $in(\lambda_{CP})\subseteq in(V_e)$ and
$out(\lambda_{CP})\subseteq out(V_e)$. In a second step, we show that if
$\lambda_{CP}(X)= \und$, then $V_e(X)=\nicefrac{1}{2}$.

Suppose that $e_1$ turns the node $X$ illegally labelled \und\ by
$\lambda_{da}$ into \inc. Then $out(\lambda_1)= out(\lambda_{da})$ and
$in(\lambda_1)= in(\lambda_{da})\cup \{X\}$. Then for all $Y \in Att(X)$,
$\lambda_{da}(X)=\exc$. Therefore, $V_k(Y)=0$ for all $Y \in Att(X)$, and
hence $\max_{Y\in Att(X)}\{V_k(Y)\}=0$. By Theorem~\ref{th:convergence-one},
$V_e(X)=1$ and therefore $X \in in(V_e)$. We set $V_k^{1,out}=out(V_k)$
and $V_k^{1,in}=in(V_k)\cup \{X\}$.

Suppose that $e_1$ turns the node $X$ illegally labelled \und\ by
$\lambda_{da}$ into \exc. Then $in(\lambda_1)=in(\lambda_{da})$ and
$out(\lambda_1)=out(\lambda_{da})\cup \{X\}$. Then there exists $Y
\in Att(X)$ such that $\lambda_{da}(X)=\inc$. Therefore, $V_k(Y)=1$
for some $Y \in Att(X)$, and hence $\max_{Y\in Att(X)}\{V_k(Y)\}=1$.
By Theorem~\ref{th:convergence-one}, $V_e(X)=0$ and therefore
$X \in out(V_e(X))$. We set $V_k^{1,out}=out(V_k)\cup \{X\}$ and
$V_k^{1,in}=in(V_k)$.

Assume that for some $i$, $in(\lambda_{i})=V_k^{i,in}$ and
$out(\lambda_{i})=V_k^{i,out}$. Now consider the ${i+1}$-th
expansion operation $e_{i+1}$.

Suppose that $e_{1+1}$ turns the node $X$ illegally labelled \und\ in
$\lambda_{i}$ into \inc. Then for all $Y \in Att(X)$,
$\lambda_{i}(X)=\exc$. Therefore, $V_e(Y)=0$ for all $Y \in Att(X)$,
and hence $\max_{Y\in Att(X)}\{V_e(Y)\}=0$. By Theorem~\ref{th:convergence-two},
$V_e(X)=1$ and therefore $X \in in(V_e)$. As before, we set $V_k^{i+1,out}=
V_k^{i,out}$ and $V_k^{i+1,in}=in(V_k)\cup \{X\}$.

Suppose that $e_{i+1}$ turns the node $X$ illegally labelled \und\ by
$\lambda_{i}$ into \exc. Then there exists $Y \in Att(X)$ such that
$\lambda_{i}(X)=\inc$. Therefore, $V_e(Y)=1$ for some $Y \in Att(X)$,
and hence $\max_{Y\in Att(X)}\{V_e(Y)\}=1$. By Theorem~\ref{th:convergence-two},
$V_e(X)=0$ and therefore $X \in out(V_e(X))$. Again, we set $V_k^{i+1,out}=
V_k^i\cup \{X\}$ and $V_k^{i+1,in}=V_k^{i,in}$.

By now we know that if $X \in V_k^{m,in}$, then $V_e(X)=1$ and $\lambda_{CP}(X)
=\inc$ and that $X \in V_k^{m,out}$, then $V_e(X)=0$ and $\lambda_{CP}(X)
=\exc$. We ask if there is some $Z \not \in V_k^{m,in}$ such that
$V_e(Z)=1$ or $Z \not \in V_k^{m,out}$ such that $V_e(Z)=0$. The answer is
no as it is explained in stage two below.\\[2ex]
\noindent{\bf Stage two:}

Let us use $C^c$ to denote $(V^{m,in}_k\cup V^{m,out}_k)$ and $C^u$ to denote
$C\backslash C^c$. Suppose $X \in C^u$.

We know that $V_k^{m,in}=in(\lambda_{CP})$ is a complete extension
and that no further expansion operation is possible from $\lambda_{CP}$, therefore
if $X \not \in in(\lambda_{CP})$, then either $\lambda_{CP}(X)=\exc$ and
hence $X \in V_k^{m,out}$, which is not possible, or $\lambda_{CP}(X)=\und$
and legally so. Therefore there exists $Y \in Att(X)$, such
that $\lambda_{CP}(Y)=\und$ and hence $0 < max_{Y\in Att(X)} \{V_e(Y)\} < 1$.

Similarly, if $X \not \in out(\lambda_{CP})$, then either $\lambda_{CP}(X)=
\inc$ and hence $X \in V_k^{m,in}$, which is not possible, or $\lambda_{CP}(X)=
\und$ and legally so. Therefore there exists $Y \in Att(X)$, such that $\lambda_{CP}%
(Y)=\und$ and  hence $0 < \max_{Y\in Att(X)} \{V_e(Y)\} < 1$ and therefore $0 <
V_e(X) < 1$.

So we know that for all $X\in C^u$, $\lambda_{CP}(X)
=\und$ and $0 < V_e(X) < 1$. In what follows, we will show that indeed
for all nodes in $C-C^c$, $V_e(X)=\nicefrac{1}{2}$. Note that since we are in 
a SCC $C$, for all $X \in C^u$, there is an infinite attack
tree with root $X$, in which every branch is of the form $X=Z_0,Z_1,Z_2,\ldots,
Z_k=X$, where for every $i>0$, $(Z_{i+1},Z_i) \in R$. Some of the $Z_i$ are in
$V^{m,out}_k$, but none can be in $V^{m,in}_k$, for that would make $Z_{i-1}$
\exc.

The proof is done by induction on the maximum distance from a
node $X$ in $C^u$ to a loop $Z_1,Z_2,\ldots,
Z_k=Z_1$, where every $Z_i \in C\backslash V_k^m$. There are
infinitely many paths from $X$ in the attack tree of $X$,
but we only need to consider the set $\Gamma(X)$ with all
$V_{\max}$-paths of $X$. Each such path is of the form $\pi(X)=(Z_0=X),Z_1,
\ldots$. Now define the distance of $X$, $\dim{X}$, as the maximum index
$i$ such that for each path $\pi(X)$, $Z_i\in bar(Z,V_k^{m,out})$. This means
that $Z_i$ is the first point in the path $\pi(X)$ which is either a
repetition of a previous point or a point in $V_k^{m,out}$.

If $dim{X}=0$, then $X$ must be attacked by a cycle involving only $X$
(otherwise $X \in V_k^{m,out}$, and then $V_e(X)=0$, a contradiction).
Therefore, we have a cycle that attacks $X$ and which involves $X$ alone.
All attackers in this cycle (i.e., $X$) have maximum value and $0 < V_{k+r}(X)
< 1$ for every $r \geq 0$. By Theorem~\ref{th:convergence-to-half}, the value
of every node in the cycle is $V_e(X)=\nicefrac{1}{2}$. Now the equilibrium value
of the node $X$ attacked by the cycle is calculated by
\begin{eqnarray*}
V_e(X) & = &(1-V_e(X))\cdot \min\left\{\frac{1}{2},\frac{1}{2}\right\} + V_e(X)
\cdot \max\left\{\frac{1}{2},\frac{1}{2}\right\}\\
       & = & \frac{1-V_e(X)}{2} + \frac{V_e(X)}{2}\\
       & = & \frac{1-V_e(X)+V_e(X)}{2}\\
       & = & \frac{1}{2}
\end{eqnarray*}
Now assume that the equilibrium value of all nodes with distance up to $k$
is $\nicefrac{1}{2}$ and consider the node $X$ with distance $k+1$. For all
$Y \in Att(X)$, we have that $\dim{Y}\leq k$. Therefore, either
$Y \in V_k^{m,out}$ in which case $V_e(Y)=0$, or by the inductive
hypothesis $V_e(Y)=\nicefrac{1}{2}$.\footnote{Note that $Att(X)\not\subseteq
V_k^{m,out}$, otherwise $X$ would be illegally labelled \und.}
Therefore we have that $\max{Y\in Att(X)}\allowbreak\{V_e(Y)\}=\nicefrac{1}{2}$
and as before
\begin{eqnarray*}
V_e(X) & = &(1-V_e(X))\cdot \min\left\{\frac{1}{2},\frac{1}{2}\right\} + V_e(X)
\cdot \max\left\{\frac{1}{2},\frac{1}{2}\right\}\\
       & = & \frac{1}{2}
\end{eqnarray*}
To conclude, for all $X \in V^{m,in}_k$, $V_e(X)=0$;
for all $X \in V^{m,out}_k$, $V_e(X)=0$; and for all $X \in C^u$,
$V_e(X)=\nicefrac{1}{2}$. $in(V_e\rest C)$ (resp., $in(\lambda_{CP}\rest C)$) in
this case is the minimal complete extension containing $in(V_k\rest C)$ (resp.,
$in(\lambda_{da}\rest C)$).

\end{itemize}
Assume the theorem holds for all nodes of depth up to $k$. We now show that
it holds for nodes of depth $k+1$.

Define $Known^0_{k+1}=\{X \in S\; | \; depth(X) \leq k\}$ and
$Known^{m+1}_{k+1}=\{X \in S\; | \; depth(X) = {k+1} \ws{and for all} Y \in Att(X),
\; Y \in Known^{m}_{k+1}\}$.

We show that for all $i \geq 0$, we have that $\lambda_{CP}(X)=\f{X}$, for
all $X \in Known^i_{k+1}$. First notice that by induction hypothesis, $\lambda_{CP}(X)=\f{X}$
for all $X \in Known^0_{k+1}$. Now suppose that $\lambda_{CP}(X)=\f{X}$ for all
$X \in Known^i_{k+1}$, then by Proposition~\ref{prop:agree-on-attacks},
$\lambda_{CP}(X)=\f{X}$ for all $X \in Known^{i+1}_{k+1}$. Since the network is
finite, $Known^{e}_{k+1}=Known^{e+1}_{k+1}$, for some $e \geq 0$. Define
$C^{u}_{k+1}=\{X \in S\; | \; depth(X) = {k+1} \}\;\backslash\; Known^{e}_{k+1}$.

By definition, if there exists $X \in C^{u}_{k+1}$ and $Y \in Att(X)$ such that
$Y \in Known^{e}_{k+1}$, then $\lambda_{CP}(Y)=\exc$ and $V_e(Y)=0$ (otherwise the
value of $X$ would be known). Therefore, we can exclude the nodes in $Known^{e}_{k+1}$
and consider $C^{u}_{k+1}$ in isolation.  $C^{u}_{k+1}$ can therefore be treated as
a network of depth $0$, and the proof will follow exactly from Cases~2 and~3 of
the base of the main induction, and hence for all $X \in C^{u}_{k+1}$, $\f{X}=
\lambda_{CP}(X)$.
\end{proof}
\end{theorem}

\begin{corollary}
Let \tuple{S,R} be an argumentation network and $V_0$ be an initial
assignment of values to the nodes in $S$. Let $\f{X}$ be the equilibrium value
of a node $X$ calculated through the Gabbay-Rodrigues Iteration Schema.
For all nodes $X \in S$, $\f{X}\in \left\{0,\nicefrac{1}{2},1\right\}$.

\begin{proof}
Follows from the possible equilibrium values of all nodes in Theorem~\ref{th:main-theorem}.
\end{proof}
\end{corollary}

\section{Discussion and Worked Examples\label{sec:discussions}}

\begin{figure}
\begin{center}
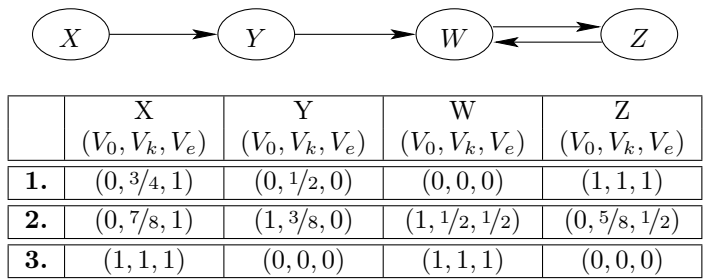 

\bigskip

\begin{tabular}{|c|c|c|c|c|c|} \hline
  & X  & Y  & W & Z\\
  & $(V_0,V_k,V_e)$  & $(V_0,V_k,V_e)$  & $(V_0,V_k,V_e)$ & $(V_0,V_k,V_e)$ \\\hline\hline
{\bf 1.}  & $(0,\nicefrac{3}{4},1)$ & $(0,\hf,0)$ & $(0,0,0)$ & $(1,1,1)$ \\ \hline\hline
{\bf 2.}  & $(0,\nicefrac{7}{8},1)$ & $(1,\nicefrac{3}{8},0)$ & $(1,\hf,\hf)$ & $(0,\nicefrac{5}{8},\hf)$ \\ \hline\hline
{\bf 3.}  & $(1,1,1)$ & $(0,0,0)$ & $(1,1,1)$ & $(0,0,0)$ \\ \hline
\end{tabular}
\end{center}

\caption{Network used in Section~\ref{sec:discussions}.\label{fig:two-networks}}
\end{figure}

Suppose we are given a network such as the one in 
Figure~\ref{fig:two-networks} with some initial values 
to its nodes. The values may or may not correspond to a complete
extension. We can write equations for the network, apply 
the Gabbay-Rodrigues Iteration Schema and obtain extensions
for the network.

For the sake of illustration, we consider three sets of representative
initial values {\bf 1.}, {\bf 2.} and {\bf 3.}. The table in
Figure~\ref{fig:two-networks} shows what
happens when these values are applied to the equations, giving both
the values at the stable point ($V_k$) and at the limit ($V_e$).
The corresponding down-admissible labellings and their resulting up-completion 
according to Caminada-Pigozzi's procedure can be obtained simply by
replacing $0$ with $\exc$, $1$ with $\inc$ and values in $(0,1)$ with \und.

Case {\bf 1.} represents the situation in which the initial values in 
the cycle $W\leftrightarrow Z$ are compatible with an extension and hence
the crisp values are preserved by the calculations. We end up with the complete 
extension $E_{\w{\bf 1}}=\{X,Z\}$. Contrast this with case {\bf 2.}, in which the 
initial values of $W$ and $Z$ are $1$ and $0$, resp. The extension
$E=\{X,W\}$ is also complete but is obtained neither by our procedure
nor by Caminada-Pigozzi's down-admissible/up-complete construction.
This can be explained as follows. The initial illegal value
of $Y$ invalidates the initial acceptance of $W$, turning it
into undecided in the calculation of the down-admissible subset. 
From that point on, the original legal assignments for $W$ and $Z$
can no longer be restored and they both end up as undecided. As
a result, we obtain the complete (but not preferred) extension 
$E_{\w{\bf 2}}=\{X\}$. This interference does not happen
in case {\bf 1.}, because there the interference of the 
undecided value of $Y$ over $W$ is dominated by $Z$'s $1$ value
that keeps $W$'s $0$ value in check (because of the behaviour
of $\max$). As a result, both $W$'s and $Z$'s initial values are
retained.

If however we start with a preferred extension, which is
also complete by definition, we get as a result unchanged initial values 
(cf. Theorem~\ref{th:main-theorem}). Caminada-Pigozzi also 
give the same result because the down-admissible labelling of a labelling 
yielding a preferred extension is the labelling itself and since that 
labelling is also complete, then the up-completion does not change anything
(case {\bf 3.} in the table of Figure~\ref{fig:two-networks}.

We can suggest an enhanced procedure to improve on the results
obtained in case {\bf 2.}, which is outlined below. The procedure
starts with an empty set of crisp values ($Crisp$) and a set of initial 
values to the nodes.

{\small
\begin{enumerate}
\item Calculate the equilibrium values for all nodes using the
iteration schema. \label{step:2}
\item If $\{X\in S \; | \; \f{X} \in \{0,1\}\} \subseteq Crisp$, stop.
The extension is defined in the set $\{X \; | \; \f{X}=1\}$.
Otherwise, set $Crisp = Crisp \cup \{X\in S \; | \; \f{X} \in \{0,1\}\}$ and
proceed to step~\ref{step:5}.
\item For every $X \in \{X \; | \; \f{X} \in \{0,1\}\}$, set $V_0=\f{X}$ and
leave $V_0(X)$ as before for the remaining nodes.\label{step:5}
\item Repeat from \ref{step:2}.
\end{enumerate}
}

The above procedure is {\em sound}, since at each run the equilibrium
values computed yield a complete extension. Note that re-using some of the 
original values does not affect soundness. If they cannot be used to generate
a larger extension, they will just converge to $\hf$. 
The procedure also {\em terminates} as long as the original network 
$S$ is finite, since a new iteration is invoked only when new crisp values 
are generated and this is bound by $|S|$. 

If we apply the procedure to Case {\bf 2.} above, in the first
run we will get $\f{X}=1$, $\f{Y}=0$, $\f{W}=\f{Z}=\hf$. 
Hence, $Crisp=\{X,Y\}$. We then run it once more, this time with initial
values $V_0(X)=1$, $V_0(Y)=0$, $V_0(W)=1$ and $V_0(Z)=0$. This will
stabilise immediately at these values and then $Crisp=\{X,Y,W,Z\}$.
In the third run, no new crisp values are generated, so we stop with 
extension $\{X,W\}$, which is a preferred extension (see case {\bf 3.} 
above). This is closer to the original values, because the 
preference of $W$ over $Z$ is preserved.

Obviously, the procedure can also
be applied  using Caminada-Pigozzi's construction instead of the Gabbay-Rodrigues
Iteration Schema of step 1. above.

\subsection{Worked Examples with Cycles}

\begin{figure}
\begin{center}
\begin{tabular}{ccc}
(L) & & (R)\\[3ex]
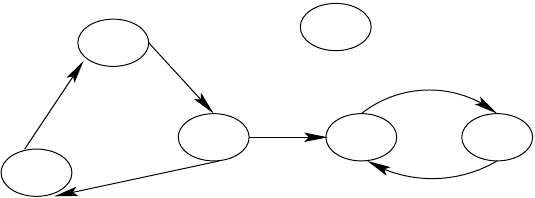 & ~~~ & 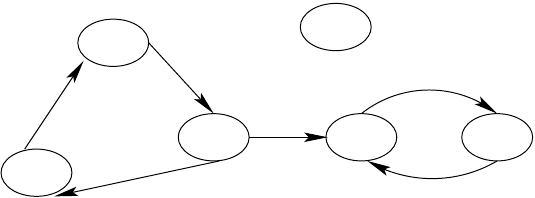
\end{tabular}
\\[3ex]
\noindent\begin{tabular}{|c|c|c|c|c|}
\hline
  & {\bf L1.}      & {\bf L2}          & {\bf R1.}      & {\bf R2} \\                      
  & $(V_0,V_k,V_e)$ & $(V_0,V_k,V_e)$   & $(V_0,V_k,V_e)$ & $(V_0,V_k,V_e)$ \\           
\hline\hline                         
X & $(0,0,0)$      & $(1,0.430,\hf)$  & $(1,1,1)$      & $(0,0,0)$ \\                  
Y & $(1,1,1)$      & $(0,0.516,\hf)$  & $(0,0,0)$      & $(1,1,1)$ \\                  
A & $(0,\hf,\hf)$  & $(1,0.516,\hf)$  & $(1,0.438,0)$  & $(0,0.562,\hf)$ \\            
B & $(1,0.266,\hf)$& $(0,\hf,\hf)$    & $(1,0.062,0)$  & $(0,\hf,\hf)$   \\            
C & $(0,0.562,\hf)$& $(0,0.430,\hf)$  & $(0,0.734,1)$  & $(1,0.266,\hf)$ \\            
Z & $(0,0.938,1)$  & $(\hf,0.992,1)$  & $(0,0.938,1)$  & $(\hf,0.969,1)$ \\            
(S,E)  & (3,58)    & (5,58)           &   (3,76)       & (3,58)          \\ \hline
\end{tabular}

\end{center}
\caption{Equilibrium and stable values of nodes involved in 
cycles.\label{fig:worked-examples}}
\end{figure}

The table in Figure~\ref{fig:worked-examples} displays initial, stable and 
equilibrium values $(V_0,V_k,\allowbreak V_e)$ for all nodes in the networks (L) and (R).
The last row of the table indicates the
iteration in which the stable values were reached and the equilibrium values 
approximated (S,E).
Obviously the equilibrium values are an approximation. We set our
tolerance as $10^{-19}$, the upper bound of the relative error due to 
rounding in the calculations in our $64$-bit machine.\footnote{Effectively 
this means that if the maximum variation in node values between
two successive iterations is smaller than $10^{-19}$, we cannot be sure 
it is not simply the result of a rounding error due to the precision 
of the computer. At that point we assume we have reached the limit of what 
can be accurately calculated.}
Independent nodes, such as $Z$ in the networks above always converge to $1$
independently of their initial values. This also happens to all nodes
whose values of the attackers all converge to $0$. Cases (L) and (R)
explore different scenarios involving cycles. The odd cycle in (L)
attacks the even cycle $X \leftrightarrow Y$ and the even cycle
in (R) attacks the odd cycle $A\rightarrow B \rightarrow C\rightarrow A$.
We start with (L), which contains an odd cycle attacking an even cycle.
The values in the odd cycle in this case will converge to $\hf$
independently of their initial values. This may or may not have an
effect on nodes that are attacked by any of the nodes in the cycle.
We start with an initial valid configuration for $X$ and $Y$ in both 
(L1) and (L2). The end results will differ though as explained
next. If $X$ starts with $0$ and $Y$ with $1$ (L1), then the
interference of the undecidedness of $B$ over $X$ is dominated 
by the $Y$'s value of $1$ and the initial values of both $X$ and $Y$
persist. However, if $X$ starts with $1$ and
$Y$ with $0$, the undecidedness of $B$ will then ``contaminate''
the $X$--$Y$ loop. It will force $X$ to become
undecided, which in turn makes $Y$ also become undecided.
As a result, all of the values will converge to $\hf$
apart from $Z$'s, which as we said is independent and will converge
to $1$ (L2). 

Now let us look at (R) in which the even cycle attacks the odd
one. (R1) and (R2) contain different initial valid
configurations for the even cycle. This time the nodes
in the even cycle are independent of external values and their
original values remain. If $X$ starts with $1$, it remains
with $1$ and this in turn breaks the odd cycle. The attacked
node $B$ is forced to converge to $0$, forcing $C$ to converge 
to $1$ and $A$ to converge to $0$ (independently of their 
initial values). An initial value of $0$ for $X$ cannot break
the odd cycle and its values will converge 
to $\hf$ independently of their initial values
(R2).

\section{Comparisons with other work\label{sec:comparisons}}
This section compares our framework with other techniques that 
deal with initial values.
Our discussions so far and the use of the Gabbay-Rodrigues Iteration
Schema were in the context of the equational approach to
an argumentation network when we are given some initial values.
Our problem was to find a solution to the system of equations that
was ``close'' to these initial values.

Two important concepts which are directly related to the work presented 
in this paper were proposed in \cite{Caminada-Pigozzi:11}, which addressed
the problem of finding an extension of an argumentation network given an 
initial labelling of its arguments. Their procedure works in two steps. Firstly,
they calculate the downward-admissible labelling of the original labelling,
which essentially consists of an admissible labelling whose crisp part is 
maximally included in the original labelling. This is done by a procedure which 
at each step, turns an illegally labelled argument from \inc\ or \exc\ 
into \und\ until no illegal crisp values remain. They called this
step a {\em contraction sequence} and it is similar to what our schema 
does to the sequence of value assignments until it becomes stable, except that 
at each iteration our schema may contract more than one node 
simultaneously, whereas theirs contracts only one node per iteration.
More importantly, their procedure is {\em non-deterministic}: it selects an 
illegally labelled node for contraction, but this requires searching for
such nodes. Hence there is an implicit cost involved in it. Even though the 
search can be optimised, it renders the overall cost of the procedure in terms
of steps higher than ours, which is truly bounded by $|S|$. Now, given an admissible 
labelling, a complete 
extension is constructed by turning nodes that are illegally labelled 
\und\ into \inc\ or \exc\  as appropriate. They call this step
an {\em expansion} and its counterpart in our procedure is the calculation
of the limit values of the sequence. Obviously, in a computer program, 
we can only approximate these limit values. In our implementation,
we stop the iterations when we can no longer guarantee the accuracy 
of the calculations without introducing rounding errors due to the
limitations of the processor. This happens in linear time too 
(see Figure~\ref{fig:worked-examples}).
In practice, the limit values can be guessed much earlier as the
iteration values can be seen to be converging towards one of the 
three values $0$, $\hf$ and $1$.

We stress that neither are we limited to the discreet values
$\exc$, $\inc$ and $\und$, nor to the \eqmax\ equation used
in the iteration schema and this allows the application of
the schema in the calculation of extensions given different
semantics (see Section~\ref{sec:conclusions}).

One can take a different approach to the one above, especially if one is 
not using any equations. One can take the view that given a network with initial
values, we should give an iteration formula that will stabilise on
some limit final values. This approach is a bit risky. One needs to
explain where the initial values come from and what is the meaning
of the iteration formula.  One also needs to check whether or not the
iteration formula is sound relative to the network's extensions in Dung's
sense. In other words, if the initial values correspond to an acceptable
Dung extension, does the iteration formula yield a result which does
not correspond to a Dung extension?  We begin with the work of
Pereira \etal\  \cite{pereira-et-al:11}, which does not take any 
equational approach but simply iterates on the values of the nodes. 
We examine in detail what they do.

In what follows, \tuple{S,R} is an acyclic argumentation network
and $f:S \longmapsto U$ is a function assigning initial 
values to the nodes in $S$. 

\newcommand{\adepname}{a-depth}
\newcommand{\adep}[1]{\w{\adepname$(#1)$}}

\begin{definition}
Consider the sequence $\alpha_0(X), \alpha_1(X), \ldots, \alpha_i(X),
\ldots$, where
\begin{eqnarray*}
\alpha_0(X) & = & f(X)\\ \alpha_i(X) & = & \alpha_{i-1}(X) +
\min\left\{f(X), 1- \max\limits_{Y\in Att(X)}\alpha_{i-1}(Y)\right\}\\
\end{eqnarray*} and let 
\[\alpha(X) =  \lim\limits_{i\imp\infty} \frac{1}{2}\alpha_i + \frac{1}{2}
   \min\big(f(X),1 - \max\limits_{Y \in Att(X)} \alpha_i(Y)\big)\]
\end{definition}

\begin{definition}The {\em attack depth of a node $X$} of 
an acyclic argumentation network, in symbols $\adep{X}$, is defined
recursively as
\[\adep{X} = \left\{ \begin{array}{ll}
0, & \wR{if} Att(X) = \varnothing \\ \big(\max\limits_{Y \in Att(X)}
\adep{Y}\big)+1, & \w{otherwise}
\end{array}
\right.
\]
\end{definition}

The function \adepname\ is well-defined, because there are no cycles
in \tuple{S,R}.

\begin{definition}\label{def-propag-serena} Given initial values for the 
nodes of an acyclic network, the function $\beta:S \longmapsto U$ provides 
a means of calculating fixed-point values for all nodes as follows.
\[
   \beta(X) = \left\{ \begin{array}{ll}
                     f(X), & \wR{if} \adep{X} = 0 \\
    \min\left\{f(X),1 - \max\limits_{Y \in Att(X)} \beta(Y)\right\}, & \w{otherwise}
                      \end{array}
\right.
\]
\end{definition}

\begin{theorem} $\alpha(X)=\beta(X)$ for all $X\in S$.

\begin{proof} The proof is done by induction on the depth of a node.

\noindent Base cases: (Depth $0$) Let $X$ be an argument node of depth
$0$. By definition, $X$ has no attacks. It follows that
\begin{eqnarray*}
\alpha_0(X) & = & f(X) \\ \alpha_1(X) & = & \frac{1}{2}\alpha_0(X) +
\frac{1}{2}\min\left\{f(X),1 - \max_{Y \in Att(X)}\alpha_0(Y)\right\} \\ & = &
\frac{1}{2}f(X) + \frac{1}{2}f(X)\\ & = & f(X)\\ \alpha_2(X) & = &
\frac{1}{2}\alpha_1(X) + \frac{1}{2}\min\left\{f(X),1 - \max_{Y \in
  Att(X)}\alpha_1(Y)\right\} \\ & = & \frac{1}{2}f(X) +
\frac{1}{2}f(X)=f(X)\\ \alpha(X) & = & \lim\limits_{i\imp
  \infty}\left\{\frac{1}{2}\alpha_i + \frac{1}{2}f(X)\right\}\\ \alpha(X)
& = & f(X) = \beta(X)\\
\end{eqnarray*}

\allowdisplaybreaks
\noindent (Depth $1$) Let $X$ be an argument node of depth $1$. 
By definition, all nodes $Y$ attacking $X$ have depth $0$. For all such
nodes $f(Y)=\alpha_0(Y)=\alpha_1(Y)=\alpha_i(Y)= \ldots = \alpha(Y) = \beta(Y)$.
\begin{eqnarray*}
\alpha_0(X) & = & f(X) \\
\alpha_1(X) & = & \frac{1}{2}\alpha_0(X) + \frac{1}{2}\min\left\{f(X),1 - \max_{Y \in Att(X)} \alpha_0(Y)\right\} \\
& = & \frac{1}{2}f(X) + \frac{1}{2}\min\left\{f(X),1 - \max_{Y \in Att(X)}\beta(Y)\right\}\\
\alpha_2(X) & = & \frac{1}{2}\left(\frac{1}{2}f(X) + \frac{1}{2}\min\left\{f(X),1 - \max_{Y \in Att(X)}\beta(Y)\right\}\right) + \\
            &   & \frac{1}{2}\min\left\{f(X),1 - \max_{Y \in Att(X)}\beta(Y)\right\}\\
            & = & \frac{1}{2^2}f(X) + \frac{1}{2^2}\min\left\{f(X),1 - \max_{Y \in Att(X)}\beta(Y)\right\} + \\
            &   & \frac{1}{2}\min\left\{f(X),1 - \max_{Y \in Att(X)}\beta(Y)\right\}\\
\alpha_i(X) & = & \frac{1}{2^i}f(X) + \sum_{i=1}^{t}\frac{1}{2^i}\cdot\min\left\{f(X),1 - \max_{Y \in Att(X)}\beta(Y)\right\}\\
            & = & \frac{1}{2^i}f(X) + \left(1 - \frac{1}{2^i}\right)\cdot \min\left\{f(X),1 - \max_{Y \in Att(X)}\beta(Y)\right\}\\
\alpha(X) & = & \lim\limits_{i \imp \infty} \alpha_i(X) \\
          & = & \lim\limits_{i \imp \infty} \frac{1}{2^i}f(X) + \left(1 - \frac{1}{2^i}\right)\cdot\min\left\{f(X),1 - \max_{Y \in Att(X)}\beta(Y)\right\}\\
          & = & \min\left\{f(X),1 - \max_{Y \in Att(X)}\beta(Y)\right\}\\
          & = & \beta(X)\\
\end{eqnarray*}

Assume that the theorem holds for nodes with attack depth up to $k$ and let $X$
be an argument node whose attack depth is $k+1$. We have that
\begin{eqnarray*}
\alpha_0(X) & = & f(X) \\
\alpha_1(X) & = & \frac{1}{2}\alpha_0(X) + \frac{1}{2}\min\left\{f(X),1 - \max_{Y \in Att(X)} \alpha_0(Y)\right\} \\
            & = & \frac{1}{2}f(X) + \\
            &   & \frac{1}{2}\min\left\{f(X),1 - \max_{Y \in Att(X)}\alpha_0(Y)\right\}\\
\alpha_2(X) & = & \frac{1}{2}\left(\frac{1}{2}f(X) + \frac{1}{2}\min\left\{f(X),1 - \max_{Y \in Att(X)}\alpha_0(Y)\right\}\right) +\\
            &   & \frac{1}{2}\min\left\{f(X),1 - \max_{Y \in Att(X)}\alpha_1(Y)\right\}\\
            & = & \frac{1}{2^2}f(X) + \frac{1}{2^2}\min\left\{f(X),1 - \max_{Y \in Att(X)}\alpha_0(Y)\right\} + \\
            &   & \frac{1}{2}\min\left\{f(X),1 - \max_{Y \in Att(X)}\alpha_1(Y)\right\}\\
\alpha_3(X) & = & \frac{1}{2}\left(\frac{1}{2^2}f(X) + \frac{1}{2^2}\min\left\{f(X),1 - \max_{Y \in Att(X)}\alpha_0(Y)\right\}\right. + \\
            &   & \left.\frac{1}{2}\min\left\{f(X),1 - \max_{Y \in Att(X)}\alpha_1(Y)\right\}\right) +\\
            &   & \frac{1}{2}\min\left\{f(X),1 - \max_{Y \in Att(X)}\alpha_2(Y)\right\}\\
            & = & \frac{1}{2^3}f(X) + \frac{1}{2^3}\min\left\{f(X),1 - \max_{Y \in Att(X)}\alpha_0(Y)\right\} + \\
            &   & \frac{1}{2^2}\min\left\{f(X),1 - \max_{Y \in Att(X)}\alpha_1(Y)\right\}+ \\
            &   & \frac{1}{2}\min\left\{f(X),1 - \max_{Y \in Att(X)}\alpha_2(Y)\right\}\\
\alpha_i(X) & = & \frac{1}{2^i}f(X) + \frac{1}{2^{i-0}}\min\left\{f(X),1 - \max_{Y \in Att(X)}\alpha_0(Y)\right\} + \\
            &   & \frac{1}{2^{i-1}}\min\left\{f(X),1 - \max_{Y \in Att(X)}\alpha_1(Y)\right\}+ \ldots  +\\
            &   & \frac{1}{2^{1}}\min\left\{f(X),1 - \max_{Y \in Att(X)}\alpha_{i-1}(Y)\right\}\\
\alpha_{i+1}(X) & = & \frac{1}{2^i}f(X) + \sum_{i=1}^{i}\frac{1}{2^i}\cdot \min\left\{f(X),1 - \max_{Y \in Att(X)}\alpha_i(Y)\right\}\\
            & = & \frac{1}{2^i}f(X) + \left(1 - \frac{1}{2^i}\right)\cdot \min\left\{f(X),1 - \max_{Y \in Att(X)}\alpha_i(Y)\right\}\\
\alpha(X) & = & \lim\limits_{i \imp \infty} \frac{1}{2^i}f(X) + \left(1 - \frac{1}{2^i}\right)\cdot \min\left\{f(X),1 - \max_{Y \in Att(X)}\alpha_i(Y)\right\}\\
          & = & \lim\limits_{i \imp \infty}\left(1 - \frac{1}{2^i}\right)\min\left\{f(X),1 - \max_{Y \in Att(X)}\alpha_i(Y)\right\}\\
& = & \lim\limits_{i \imp \infty}\min\left\{f(X),1 - \max_{Y \in Att(X)}\alpha_i(Y)\right\}\\
& = & \min\left\{f(X),1 - \max_{Y \in Att(X)}\lim\limits_{i \imp \infty}\alpha_i(Y)\right\}\\
\alpha(X) & = &\min\left\{f(X),1 - \max_{Y \in Att(X)}\alpha(Y)\right\}\\
\end{eqnarray*}
But the attack depth of the nodes $Y \in Att(X)$ is no higher than $k$. By the induction 
hypothesis we have that $\alpha(Y)=\beta(Y)$ for all $Y\in Att(X)$ and hence
\begin{eqnarray*}
\alpha(X) & = & \min\left\{f(X),1 - \max_{Y \in Att(X)}\beta(Y)\right\} = \beta(X)
\end{eqnarray*}
\end{proof}
\end{theorem}

The theorem above shows that when there are no cycles, for any node $X$,
the sequence $\alpha_i(X)$ converges to the value $\beta(X)$, which can 
be calculated by considering the tree with root $X$ and propagating
values from the leaves to the root according to Definition~\ref{def-propag-serena}.

One can argue that the procedure is not sound with respect to
admissibility. In particular, the algorithm does not turn arbitrary
initial values into admissible ones. If we give initial value $0$ to a node 
which should not be labelled \exc, the algorithm does not correct the node's
value and it remains illegally \exc. Likewise, if we start with a 
two-node cycle $A \leftrightarrow B$ and provide initial values to $A$ and $B$ 
that correspond to a complete extension, say $A=1$, $B=0$, in the limit we get 
values $A=\frac{1}{2}$ and $B=0$. Ideally, the initial values should remain 
the same as in the Gabbay-Rodrigues Iteration Schema (and indeed Caminada
and Pigozzi's down-admissible/up-complete construction).

\section{Conclusions and Future Research\label{sec:conclusions}}

This paper investigated aspects concerned with argumentation networks where the 
arguments are provided with initial values. 
We are aware that assigning values to nodes and propagating values through 
the network has been independently investigated before as in, e.g.,
\cite{grad-arg:2005,besnard-hunter:2001}. However, our approach is different
because we see a network as a generator for equations whose solutions
generalise the concept of extensions of the network. 

There are advantages to
using equations to calculate extensions in this way as numerical values arise
naturally in many applications where argumentation systems are used and the 
behaviour of the node interactions can be described naturally 
using equations. In addition, there are many mathematical tools 
to help find solutions to the equations.

The equational approach is general enough to be adapted to particular applications.
For instance, the arguments themselves may be expressed as some proof in a fuzzy 
logic and then the initial values can represent the values of the conclusions 
of the proofs, in the spirit of Prakken's work \cite{prakken:2010}; or they can be
obtained as the result of the merging of several networks, as proposed in
\cite{gabbay-rodrigues:12,gabbay-rodrigues-jlc:13}. 

In this paper, we showed that the equations can be solved through an iterative 
process, as in Newton's method and as such one can regard initial values as  
initial guesses or a desired configuration of the extension. The 
Gabbay-Rodrigues Iteration Schema takes the following {\em generalised} 
form:
\[
V_{i+1}(X) = (1-V_i(X)) \cdot \min\left\{\nicefrac{1}{2},g({\cal N}(X))\right\} +
   V_i(X) \cdot \max\left\{\nicefrac{1}{2},g({\cal N}(X))\right\}
\]

In this paper, we considered the special case where $g$ is $\min$ and ${\cal N}(X)$ 
is the set of complemented values of the nodes in the ``neighbourhood'' of $X$ 
(i.e., the attackers of $X$).\footnote{Note that $1 - \max_{Y\in Att(X)} \{V(Y)\}=\min_{Y\in Att(X)}
\{1-V(Y)\}$.} Other operations can be used for argumentation systems,
whose relationship with the schema is being further investigated. 
One such operation is {\em product}, which unlike $\min$ combines the
strength of the attacks on a node. Another interesting possibility
is to use the schema for {\em abstract dialectical frameworks} (ADFs)
\cite{adf:10}. ADFs require the specification of a possibly unique
type of equation for each node. Consider the ADF with nodes $a$, $b$, 
$c$ and $d$ with $R=\{(a,b),(b,c),(c,c)\}$. The ADF equations are: 
$C_a=\top$, $C_b=a$, $C_c=c \wedge b$ and $C_d=\neg d$. The complete 
models for this ADF are $m_1=(t,t,u,u)$, $m_2=(t,t,t,u)$ and $m_3=(t,t,f,u)$. 
The Gabbay-Rodrigues schema converges to $m_1$ given initial values
$(1,1,\nicefrac{1}{2},\nicefrac{1}{2})$; to $m_2$ given initial values
$(1,1,1,1)$; and to $m_3$ given initial values
$(0,0,0,0)$.

For the case of $\min$, we showed that 
the values generated at each iteration in the schema eventually ``stabilise'' 
by changing illegal crisp values into undecided. This
process will calculate the down-admissible labelling of the initial values,
as in \cite{Caminada-Pigozzi:11}, in time $t$ linear to the set of arguments
($t\leq |S|$).
If we carry on the calculation, the values of the sequence in the limit 
will correspond to a complete extension of the original network. Obviously,
the values corresponding to a legitimate extension are all legal. If they
are given as input, the sequence will immediately stabilise.
In practice, a few iterations are sufficient to indicate what the values 
will converge to in the limit. 
We have also outlined a procedure which can improve on the calculation
above by propagating crisp values and replacing the remaining undecided values
with their initial counterparts after each run of the iterations. This 
procedure terminates when no new crisp values are generated. Original crisp 
values which are compatible with a calculated extension can thus be preserved 
and hence we can end up with a larger complete extension than
the one obtained through a single run. This extension is as compatible 
as possible with the initial values.

\section*{Acknowledgements} 
The authors would like to thank Massimiliano Giacomin, Gabriella Pigozzi, Martin Caminada and Sanjay Modgil for comments and discussions on the topic of this paper.

\bibliographystyle{plain}
\bibliography{ESNAN}

\begin{thebibliography}{10}

\bibitem{sup-att-net:2005}
H.~Barringer, D.~M. Gabbay, and J.~Woods.
\newblock Temporal dynamics of support and attack networks.
\newblock In D.~Hutter and W.~Stephan, editors, {\em Mechanizing Mathematical
  Reasoning}, 2005.
\newblock LNCS, vol. 2605.

\bibitem{besnard-hunter:2001}
P.~Besnard and A.~Hunter.
\newblock A logic-based theory of deductive arguments.
\newblock {\em Artificial Intelligence}, 128(1-2):203 -- 235, 2001.

\bibitem{adf:10}
G.~Brewka and S.~Woltran.
\newblock Abstract dialectical frameworks.
\newblock In {\em Proceedings of the 12th International Conference on the
  Principles of Knowledge Representation and Reasoning: KR'10}, pages 102 --
  111. AAAI Press, 2010.

\bibitem{Caminada:07}
M.~Caminada.
\newblock An algorithm for computing semi-stable semantics.
\newblock In {\em Proceedings of the 9th European Conference on Symbolic and
  Quantitative Approaches to Reasoning with Uncertainty}, ECSQARU '07, pages
  222--234, Berlin, Heidelberg, 2007. Springer-Verlag.

\bibitem{caminada:2011}
M.~Caminada.
\newblock A labelling approach for ideal and stage semantics.
\newblock {\em Argument and Computation}, 2(1):1--21, 2011.

\bibitem{caminada-gabbay:2009}
M.~Caminada and D.~M. Gabbay.
\newblock A logical account of formal argumentation.
\newblock {\em Studia Logica}, 93(2-3):109--145, 2009.

\bibitem{Caminada-Pigozzi:11}
M.~Caminada and G.~Pigozzi.
\newblock On judgment aggregation in abstract argumentation.
\newblock {\em Autonomous Agents and Multi-Agent Systems}, 22(1):64--102, 2011.

\bibitem{grad-arg:2005}
C.~Cayrol and M.-C. Lagasquie-Schiex.
\newblock Graduality in argumentation.
\newblock {\em Journal of Artificial Intelligence Research}, 23:245--297, 2005.

\bibitem{pereira-et-al:11}
C.~da~Costa~Pereira, A.G.B. Tettamanzi, and S.~Villata.
\newblock Changing one's mind: erase or rewind? possibilistic belief revision
  with fuzzy argumentation based on trust.
\newblock In {\em Proceedings of the 22nd International joint conference on
  artificial intelligence : IJCAI'11}, pages 164 -- 171, Menlo Park, 2011. AAAI
  Press.

\bibitem{dung-aaf}
P.~M. Dung.
\newblock On the acceptability of arguments and its fundamental role in
  nonmonotonic reasoning, logic programming and n-person games.
\newblock {\em Artificial Intelligence}, 77:321--357, 1995.

\bibitem{leite-martins-to-appear:2014}
S.~E{\v{g}}ilmez, J.~Leite, and J.~Martins.
\newblock Extending social abstract argumentation with votes on attacks.
\newblock In {\em Proceedings of the 2nd International Workshop on Theory and
  Applications of Formal Argumentation (TAFA'13)}, to appear 2014.

\bibitem{gabbay-rodrigues:532}
D.~Gabbay and O.~Rodrigues.
\newblock Probabilistic argumentation. {A}n equational approach.
\newblock To appear.

\bibitem{dov-eq-short:2011}
D.~M. Gabbay.
\newblock Introducing equational semantics for argumentation networks.
\newblock DOI: 10.1007/978-3-642-22152-1\_2, 2011.

\bibitem{dov-eq:12}
D.~M. Gabbay.
\newblock Equational approach to argumentation networks.
\newblock {\em Argument and Computation}, 3:87--142, 2012.
\newblock DOI: 10.1080/19462166.2012.704398.

\bibitem{gabbay-argumentation:13}
D.~M Gabbay.
\newblock {\em Meta-logical Investigations in Argumentation Networks},
  volume~44 of {\em Studies in Logic: Mathematical Logic and Foundations}.
\newblock College Publications, 2013.
\newblock ISBN: 978-1-84890-103-2.

\bibitem{gabbay-rodrigues-jlc:13}
D.~M. Gabbay and O.~Rodrigues.
\newblock A equational approach to the merging of argumentation networks.
\newblock {\em Journal of Logic and Computation}, 2012.

\bibitem{gabbay-rodrigues:12}
D.~M. Gabbay and O.~Rodrigues.
\newblock A numerical approach to the merging of argumentation networks.
\newblock In M.~Fisher, L.~van~der Torre, M.~Dastani, and G.~Governatori,
  editors, {\em Proceedings of CLIMA XIII}, pages 195--212. Springer-Verlag,
  2012.

\bibitem{Hassell:78}
M.~P. Hassell.
\newblock {\em The Dynamics of Arthropod Predator-Prey Systems}.
\newblock Princeton University Press, 1978.

\bibitem{leite-martins:2011}
J.~Leite and J.~Martins.
\newblock Social abstract argumentation.
\newblock In {\em Proceedings of the 22nd International Joint Conference on
  Artificial Intelligence}, 2011.
\newblock To appear.

\bibitem{prakken:2010}
H.~Prakken.
\newblock An abstract framework for argumentation with structured arguments.
\newblock {\em Argument and Computation}, 1:93--124, 2010.

\bibitem{Suli-Mayers:2003}
E.~S\"{u}li and D.~F. Mayers.
\newblock {\em {An Introduction to Numerical Analysis}}.
\newblock Cambridge University Press, September 2003.

\end{thebibliography}

\newpage

\appendix


\section{Predator-Prey and Argumentation Motivating Case Studies\label{app:examples}}

Let us motivate our ideas through two main examples. Our purpose is to 
make some conceptual distinction about iteration processes.

\begin{example}
\label{example:446-EE1}
Let us look at an example from biology. This is a model by
M. P. Hassell \cite{Hassell:78} of the dynamics of a system with two 
parasitoids ($\bP$ and $\bQ$) and one host ($\bN$). The interactions in
the ecology are depicted in Figure~\ref{fig:449-F953}. The equations modelling
the dynamics are the following (see \cite[p. 295]{sup-att-net:2005}).

\[\begin{array}{l}
\bN_{t+1} =\lambda
\bN_tf_1(\bP_t)f_2(\bQ_t)\\[0.6ex]
\bP_{t+1}=\bN_t[1-f_1(\bP_t)]\\ [0.6ex]
\bQ_{t+1}=\bN_tf_1(\bP_t)[1-f_2(\bQ_t)]\\[1ex]
\end{array}
\]

In the above equations the subscripts $t$ and $t+1$ indicate two successive generations of $\bP$, $\bQ$ and $\bN$; 
$\lambda$ is the finite host rate of increase; and the functions $f_1$ and 
$f_2$ are the probabilities of a host not being found by $\bP_t$ or $\bQ_t$ 
parasitoids, respectively. This model applies to two quite distinct types of 
interaction that are
frequently found in real systems. It applies to cases where $\bP$ acts
first, to be followed by $\bQ$ acting only on the survivors. Such is the
case where a host population with discrete generations is parasitized
at different developmental stages. In addition, it applies to cases
where both $\bP$ and $\bQ$ act together on the same host stage, but the
larvae of $\bP$ always out-compete those of $\bQ$, should multi-parasitism
occur.

The functions $f_1$ and $f_2$ are:
\[\renewcommand{\arraystretch}{2}
\begin{array}{l}
f_1(\bP_t) =\left[ 1+\displaystyle\frac{a_1\bP_t}{k_1}\right]^{-k_1}\\
f_2(\bQ_t) =\left[ 1+\displaystyle\frac{a_2\bQ_t}{k_2}\right]^{-k_2}
\end{array}
\]
where $a_1$, $a_2$, $k_1$ and $k_2$ are constants.

To simplify and later compare the biological model with the 
argumentation model, we put  $k_1=k_2=-1$.

\begin{figure}[h]
\begin{center}
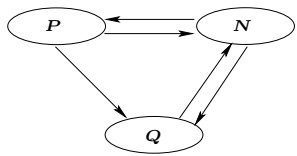
\end{center}
\caption{A sample biological network.\label{fig:449-F953}}
\end{figure}

\noindent This gives
\begin{eqnarray*}
f_1(\bP_t) & = & 1 - a_1\bP_t \\
f_2(\bQ_t) & = & 1 - a_2\bQ_t \\
\end{eqnarray*}
and therefore, the equations are
\medskip

\noindent \begin{tabular}{llll}
$(1,t)$: & $\bN_{t+1}$ & = & $\lambda N_t(1-a_1\bP_t)(1-a_2\bQ_t)$ \\
$(2,t)$: & $\bP_{t+1}$ & = & $a_1 N_t\bP_t$ \\
$(3,t)$: & $\bQ_{t+1}$ & = & $a_2 \bQ_t \bN_t(1-a_1 \bP_t)$ \\
\end{tabular}
\medskip

At a state of equilibrium, we get the following fixpoint equations:
\begin{eqnarray}
\bN & = & \lambda \bN(1-a_1\bP)(1-a_2\bQ)  \label{eq:ex:1}\\
\bP & = & a_1 \bN \bP \label{eq:ex:2}\\
\bQ & = & a_2 \bQ \bN (1-a_1 \bP) \label{eq:ex:3}
\end{eqnarray}
It can be easily seen from the above equations that one of the solutions
is $\bP=\bQ=\bN=0$ (the ``all zero'' solution). If we ignore it, we get 
from~(\ref{eq:ex:2})
that 
\begin{eqnarray}
\bN & = & \frac{1}{a_1}
\end{eqnarray}
and from~(\ref{eq:ex:3}) we get
\begin{eqnarray}
1 & = & a_2\cdot \frac{1}{a_1}(1-a_1 \bP)
\end{eqnarray}
and hence 
\begin{eqnarray*}
a_1 & = & a_2 - a_2 a_1 \bP\\
\bP & = & \frac{a_2 - a_1}{a_1 a_2}
\end{eqnarray*}
From~(\ref{eq:ex:1}), we get
\begin{eqnarray*}
1 & = & \lambda \big(1-\frac{a_1(a_2-a_1)}{a_1 a_2}\big)(1-a_2\bQ) \\
1 & = & \frac{\lambda a_1}{a_2}(1 - a_2 \bQ)
\end{eqnarray*}
so
\begin{eqnarray*}
\frac{a_2}{\lambda a_1} & = & 1 - a_2 \bQ\\
a_2\bQ & = & \frac{\lambda a_1 - a_2}{\lambda a_1}\\
\bQ & = & \frac{\lambda a_1 - a_2}{\lambda a_1 a_2}\\
\end{eqnarray*}
To have a specific example for discussion let $a_1=2$, $a_2=3$,
$\lambda=2$. We get $\bN=0.5$, $\bP=\frac{1}{6}$ and $\bQ=\frac{1}{12}$.
Indeed, substituting these values in the equations we have

\noindent\begin{tabular}{lrcl}
(1) & $1$ & $=$ & $2\left(1-2\cdot \frac{1}{6}\right)\left(1-\frac{3}{12}\right)$ \\
    &     & $=$ & $2 \cdot \frac{2}{3} \cdot \frac{9}{12}$ \\
    &     & $=$ & $2 \cdot \frac{18}{36}$ \\
    &     & $=$ & $1$
\end{tabular}

\noindent\begin{tabular}{lrcl}
(2) & $1$ & $=$ & $2\cdot \frac{1}{2}$\\
    &     & $=$ & $1$ \\
\end{tabular}

\noindent\begin{tabular}{lrcl}
(3) & $1$ & $=$ & $3\cdot \frac{1}{2}\left(1 - \frac{2}{6}\right)$\\
    &     & $=$ & $\frac{3}{2}\cdot \frac{4}{6}$ \\
    &     & $=$ & $1$
\end{tabular}

Let us substitute $a_1$, $a_2$ and $\lambda$ in the equations and
pretend we do not know the solution. We get the equations:

\medskip

\noindent\begin{tabular}{lrcl}
(1*) & $\bN$ & $=$ & $2\bN(1-2\bP)(1-3\bQ)$\\
(2*) & $\bP$ & $=$ & $2\bP \bN$\\
(3*) & $\bQ$ & $=$ & $\frac{3}{2}\bQ(1-2\bP)$\\
\end{tabular}
\medskip

So we have a system of equations modelling a certain ecology.

The equations above give rise to the iteration equations

\medskip
\noindent \begin{tabular}{llll}
$(1*,i)$: & $\bN_{i+1}$ & = & $2 \bN_i(1-2\bP_i)(1-3\bQ_i)$ \\
$(2*,i)$: & $\bP_{i+1}$ & = & $2 \bN_i\bP_i$ \\
$(3*,i)$: & $\bQ_{i+1}$ & = & $\frac{3}{2} \bQ_i(1-2 \bP_i)$ \\
\end{tabular}
\medskip

Let us discuss our options. We have a system of equations involving
\bN, \bP\ and \bQ\ and we want to solve it. We do not know whether there
are solutions.

\medskip

\noindent {\bf Option 1} -- a {\em mathematical view}. Let us just find
a solution. We can guess a candidate solution, use Newton's method
and iterate. Let us do this with the guess $\bN_0=\bP_0=\bQ_0=\frac{1}{2}$
and iterate. These are equations $(1*,i)$, $(2*,i)$ and $(3*,i)$ for $i=1$.

Because the equations come from ecological considerations, the iterations 
are not just a numerical device but also have an evolutionary meaning. 
However, our view is purely mathematical. The corresponding to the
meaning is accidental.

We get
\medskip

\noindent \begin{tabular}{lllll}
$\bN_{1}$ & $=$ & $2\cdot \frac{1}{2}\cdot N_i(1-1)\left(1-\frac{3}{2}\right)$ & $=$ & $0$ \\[1ex]
$\bP_{1}$ & $=$ & $2\cdot \frac{1}{2}\cdot \frac{1}{2}$  & $=$ & $0$ \\[1ex]
$\bQ_{1}$ & $=$ & $\frac{3}{2}\cdot \bQ_i(1-2 \bP_i)$  & $=$ & $0$ \\
\end{tabular}

\medskip

\noindent \begin{tabular}{lll}
$\bN_{2}$ & $=$ & $0$ \\
$\bP_{2}$ & $=$ & $0$ \\
$\bQ_{2}$ & $=$ & $0$ \\
\end{tabular}

\medskip

We converge to the ``all zero'' solution.

\medskip
\noindent {\bf Option 2} -- a {\em semantical view}. We seek a solution motivated
not by mathematics but by the meaning of the equations: by ecological considerations.
So let us adopt the {\em friends of parasites} view and say that we are equal 
and we all have a right to live and so let us seek a steady state of compromise
and living together in tolerance and understanding, namely $\bN_0=\bP_0=\bQ_0=
\frac{1}{2}$. 

Unfortunately using Newton's method leads us, as shown 
above, to the solution $\bP=\bQ=\bN=0$. In biological terms this is not good,
it means everything is dead. So we may need a better iteration schema, a 
schema suitable for the biological interpretation.

We can choose to be selfish and cruel and start with $\bN_0=1$ and $\bP_0=\bQ_0=0$.
This means we aim at full population and no parasites. Iterating the equations
will give us

\medskip

\noindent \begin{tabular}{lllll}
$\bN_{1}$ & $=$ & $2$\\
$\bP_{1}$ & $=$ & $0$\\
$\bQ_{1}$ & $=$ & $0$\\
\end{tabular}

\medskip

\noindent \begin{tabular}{lll}
$\bN_{k}$ & $=$ & $2^k$ \\
$\bP_{k}$ & $=$ & $0$ \\
$\bQ_{k}$ & $=$ & $0$ \\
\end{tabular}

\medskip

This does not lead to a solution. It diverges!

The reader can check that even if the initial values are very close to a solution, 
the method in general will not converge to the solution. 
\end{example}

\begin{remark}
\label{rem:446-r2}
The conclusion we draw from Example~\ref{example:446-EE1} is that we must be 
aware that some iteration processes can be mathematical  only, just possibly leading 
to a mathematical solution but otherwise semantically meaningless, and some may be 
semantically meaningful and useful in the context of the application area from which
the equations arise.

This observation shall become sharper and clearer in the case of our next example
from abstract argumentation.
\end{remark}


\begin{example}
\label{example:446-EE2}
Consider Figure~\ref{fig:449-F953} again but this time as an argumentation 
network where $\bN$, $\bP$, $\bQ$ are arguments. This network has three extensions
$E_1$, $E_2$ and $E_3$, namely

\medskip

\begin{tabular}{rcll}
$E_1$ & = & \bP\ is \inc \\
      & = & \bN\ and \bQ\ are \exc\\ \\
$E_2$ & = & \bN\ is \inc \\
      & = & \bP\ and \bQ\ are \exc\\ \\
$E_3$ & = & \bP, \bN\ and \bQ\ are all \und\\
\end{tabular}

In \cite{dov-eq-short:2011,dov-eq:12,gabbay-argumentation:13}, we showed
how to provide semantics for abstract argumentation in terms of equations. 
These equations are generated according to {\em equation schema}, of which two of
the most significant ones  are $\eqmax$ and $\eqinv$, described next. 

Let $Att(X)=
\{Y_1,\ldots,Y_k\}$ be all the attackers of $X$. Consider $X$,
$Y_1$,\ldots,$Y_k$ as variables ranging over $[0,1]$. Define

\begin{eqnarray*}
G_{max}(Att(X)) & = & 1 - \max \{Y_1,\ldots,Y_k\} \\
G_{inv}(Att(X)) & = & \Pi_{i=1}^{k} (1 - Y_i)\\
\end{eqnarray*}
The equation we write for a node $X$ is
\[ X = G(Att(X)) \eqno \eqr{*} \]
where $G$ can be $G_{max}$ or $G_{inv}$ or some other function. 
We consider $X=1$ to mean $X$ is \inc; $X=0$ to mean 
$X$ is \exc; and $0< X < 1$ to mean that $X$ is \und. The background 
material on the equational approach is given in the next section. It is sufficient 
to say here that $G_{max}$ follows more closely the traditional semantics of 
argumentation networks being only concerned about the highest strength of 
attack to a node. The solutions to the equations using $G_{max}$ correspond 
to the traditional concept of extensions (in Dung's sense) taking the nodes 
with value $1$ in a solution to be the nodes in the extension.

$G_{inv}$ on the other hand is also sensitive to the number of attackers to
a node. For example, assume there are $10$ undecided attackers $Y_i$ of $X$ each
having value $\frac{1}{2}$ (\und), then the value of $X$ becomes
$\frac{1}{2^{10}}$ under $G_{inv}$, while under $G_{max}$, the value of 
$X$ is simply $\frac{1}{2}$. Note that $X$ is nearer to
$0$ (i.e., \exc) in the $G_{inv}$ case!

The $G_{max}$ equations for the network in Figure~\ref{fig:449-F953} are:
\begin{eqnarray}
\bN & = & 1 - \max\{\bP,\bQ\} \label{eq:max-1}\\
\bP & = & 1 - \bN\\
\bQ & = & 1 - \max\{\bP,\bN\} \label{eq:max-3}
\end{eqnarray}
and its $G_{inv}$ equations are:
\begin{eqnarray}
\bN & = & (1 - \bP)(1 - \bQ) \label{eq:inv-1}\\
\bP & = & (1 - \bN) \\
\bQ & = & (1 - \bP)(1 - \bN) \label{eq:inv-3}
\end{eqnarray}

The $G_{max}$ equations have the solutions: $\bN=\bQ=0$ and $\bP=1$ ($E_1$); 
$\bN=1$, $\bP=\bQ=0$ ($E_2$); and $\bN=\bP=\bQ=\frac{1}{2}$ ($E_3$). The $G_{inv}$ 
only accepts the first two solutions with the extension $E_3$ not being possible.%
\footnote{The specific behaviour of $G_{inv}$ is outside of the scope of this paper.
However it is explored in detail in \cite{gabbay-rodrigues:532}.}

Now suppose we actually do not know whether there are solutions or what they would be
and let us consider our options. We have a system of equations involving 
\bN, \bP and \bQ\ and we want to try and solve it.  

\medskip

\noindent {\bf Option 1} -- A {\em mathematical view}. Let us just find
a solution. This is a numerical analysis problem. We can guess a candidate solution; 
use, for instance, Newton's method; and iterate in the hope of converging to a solution. 
\medskip
\noindent {\bf Option 2} -- A {\em semantical view}. We seek a solution motivated
not by mathematics but by the meaning of the equations; by argumentation considerations.
Newton's method may not be adequate here. We want a method which, if we start very near 
a solution, then we get convergence to that desired solution. Here we cannot accept any 
solution. We want solutions which reflect the input. So we need to devise algorithms involving iterations which have a semanical meaning, in addition to the usual mathematical 
properties that the iteration sequences calculated by these algorithms converge. 
This point is important. Suppose we give the following interpretation to the 
network. $100$ voters need to form a committee from amongst three experts $\bP$, $\bQ$ 
and $\bN$ to give an opinion on a crucial issue. All of them vote for
$\bN$ to be included (\inc), none of them want $\bP$ to be included (i.e, they
want $\bP$ to be \exc), and they are equally divided on their support for $\bQ$ 
(\und). There is however an additional information about these candidates which
is of a personal nature of which the voters are not aware. These are represented
by the attack relation in the network, in which $X \rightarrow Y$ means $X$ refuses
to work with $Y$. We thus say that we have a numerical assignment $\bN=1$, $\bP=0$ 
and $\bQ=\frac{1}{2}$ and we now ask what extension (i.e., what committee membership) 
is nearest to this majority vote? At first glance, the reader may think that it is 
extension $E_2$ ($\bN$ is \inc, and $\bP$ and $\bQ$ are \exc), because it agrees
with the wishes of all of the voters that $\bN$ is \inc\ and $\bP$ is \exc.  We would
like our iteration algorithm to give us this result if possible. 

Let us look at what Newton's method would do to these initial values.


We start with initial values $\bN_0=1$, $\bP_0=0$ and $\bQ_0=\frac{1}{2}$ and 
iterate for the case of $G_{max}$ (equations~(\ref{eq:max-1})--(\ref{eq:max-3})). 
We shall see that iterating in this way is not satisfactory. We get

\medskip
\begin{tabular}{lll}
$\bN_1=\frac{1}{2}$, & $\bP_1=0$, & $\bQ_1=0$\\
$\bN_2=1$, & $\bP_2=\frac{1}{2}$, & $\bQ_2=\frac{1}{2}$\\
$\bN_2=\frac{1}{2}$, & $\bP_2=0$, & $\bQ_2=0$\\
\end{tabular}
\medskip

There is no convergence here, so this is not
satisfactory as we do not get an answer for membership (i.e., 
no extension in the argumentation sense). 

Let us now compare with the Gabbay-Rodrigues Iteration Schema for $G_{max}$, 
which is the main subject matter of this paper and is introduced in 
Section~\ref{sec:initial-values}. The schema always yields a solution which 
corresponds to an extension in the argumentation sense.

Let \tuple{S,R} be an argumentation network and $X,Y_i \in 
S$ be considered variables. Let $Att(X)=\{Y_j\}$ ($j \geq 0$) be the attackers
of $X$ and let the equations be $X=G_{max}(Att(X))$.\footnote{$G_{inv}$ can also
be used, with different results.} Let $V_i(X)$ be the value
of $X$ at iteration step $i$. Then the value of $X$ at step $i+1$ is
calculated as
\begin{eqnarray*}
V_{i+1}(X) & = & 
   (1-V_i(X)) \cdot \min\left\{\frac{1}{2},G\left(\{V_i(Y_j)\}\right)\right\} +\\
     &   & 
   V_i(X) \cdot \max\left\{\frac{1}{2},G(\{V_i(Y_j)\})\right\}
\end{eqnarray*}
So for the network in Figure~\ref{fig:449-F953} and $G_{max}$ we get
\begin{eqnarray*}
V_{i+1}(\bN) & = & 
   (1-V_i(\bN)) \cdot \min\left\{\frac{1}{2},1 - \max\{V_i(\bP),V_i(\bQ)\}\right\} +\\
      &   & 
   V_i(\bN) \cdot \max\left\{\frac{1}{2},1 - \max\{V_i(\bP),V_i(\bQ)\}\right\}\\
V_{i+1}(\bP) & = & 
   (1-V_i(\bP)) \cdot \min\left\{\frac{1}{2},1 - V_i(\bN)\right\} +\\
      &   & 
   V_i(\bP) \cdot \max\left\{\frac{1}{2},1 - V_i(\bN)\right\}\\
V_{i+1}(\bQ) & = & 
   (1-V_i(\bQ)) \cdot \min\left\{\frac{1}{2},1 - \max\{V_i(\bP),V_i(\bN)\}\right\} +\\
      &   & 
   V_i(\bQ) \cdot \max\left\{\frac{1}{2},1 - \max\{V_i(\bP),V_i(\bN)\}\right\}
\end{eqnarray*}

Let us now take the initial conditions $V_0(\bN)=1$, $V_0(\bP)=0$ and
$V_0(\bQ)=0$ and calculate the iterations. All values will converge to 
$\frac{1}{2}$. 

The perceptive reader might ask what is the philosophy behind the schema that
led us to the extension $E_3$, rather than to the larger extension $E_2$. The
schema is very sensitive to the undecided values. It acts cautiously in 
considering the votes for $\bN$'s being included, because a proportion
of the voters wanted $\bQ$ to be included but $\bN$ and $\bQ$ cannot
work together. 

\end{example}

\section{Numerical Argumentation Networks\label{sec:nans}}


In \cite{sup-att-net:2005}, the idea of {\em support and attack networks} 
was initially proposed. These networks allow for the
assignment of initial values to the nodes of the graph; the specification 
of a {\em transmission factor} associated with the strength with which an
attack between arguments is carried out; and the higher-level notion of
an attack to an attack. In \cite{gabbay-rodrigues:12}, we showed how some
of these features can be used in the merging of argumentation networks. 
The {\em numerical argumentation networks} we now propose share some
of the features of the support and attack networks, but introduce
a functional approach to the computation of interaction between nodes.




\begin{definition}[Numerical Argumentation Network] A {\em numerical
argumentation network} is a tuple $\tuple{S,R,\fzero,\feq,g,h,\Pi}$, where 
\begin{itemize}
\item $S$ is a set of nodes, representing arguments;
\item $R \subseteq S^2$ is an attack relation, where $(X,Y) \in R$ means 
``$X$ attacks $Y$'';
\item $\fzero:S \longrightarrow U$ is a function assigning initial values 
to the nodes in $S$; 
\item $g$ is a function to combine attacks to a node; 
\item $h$ is a function to combine the initial value of a node with 
the value of its attack;
\item $\Pi$ is an algorithm to compute equilibrium values $\f{X}$, for each
node $X \in S$.
\end{itemize}
\end{definition} 

We assume that $g$ and $h$ are possibly distinct argumentation-friendly 
functions according to Definition~\ref{def:arg-friendly}.
The equilibrium value of a node $X$, $V_e(X)$, is defined as 
$h(V_0(X),g_{Y\in Att(X)}(\{1-\f{Y}\}))$ and computed by the 
algorithm $\Pi$. Since the computation of the equilibrium values of the
nodes takes the values of the attacking nodes into account, in Cayrol and 
Lagasquie-Schiex's terminology, the algorithm $\Pi$ offers a procedure to 
perform an {\em   interaction-based valuation} of the graph $\tuple{S,R}$.  
However, our approach
is more general because the computation is done in terms of equations
satisfying abstract principles.

\begin{figure}[htb]

\medskip

\begin{center}
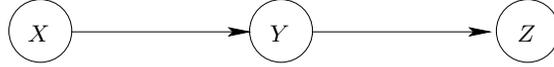
\end{center}

\medskip

\caption{A simple argumentation graph without cycles.\label{fig:seq-attacks}}
\end{figure}

We start our discussion with a simple graph without cycles, such as the
one in Figure~\ref{fig:seq-attacks} to illustrate how numerical argumentation
networks are used in the context of the \df\ functions seen in this paper.

Given initial values $V_0(X)$, $V_0(Y)$, and $V_0(Z)$ for the nodes $X$, $Y$ 
and $Z$, respectively, we want the values of $V_e(X)$, $V_e(Y)$ and $V_e(Z)$ 
to depend on them. Since the node $X$ is not 
attacked by any node, its equilibrium value $V_e(X)$ is defined as $h(V_0(X),
g(\varnothing))=h(V_0(X),1)=V_0(X)$. 
However, the value of $V_e(Y)$ and $V_e(Z)$ depend not only
on their initial values, but also on the equilibrium values of their
attackers. This suggests some notion of {\em directionality} in the 
computation. 

\begin{figure}[htb]
\begin{center}
\hspace*{-4.5cm}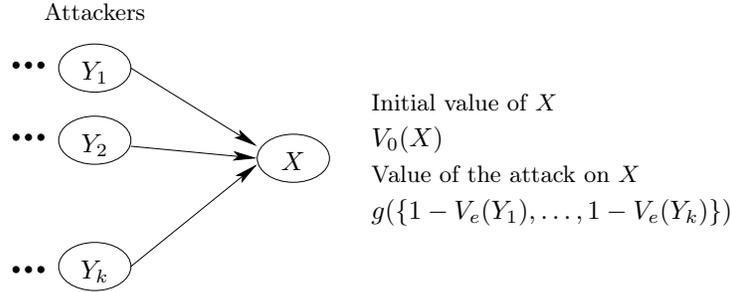
\end{center}
\caption{Attacks to a node and its initial value.\label{fig:multiple-attacks-vo}}
\end{figure}

Now consider a more complex network, in which the node $X$ has a number of
attackers as well as an initial value $V_0(X)$ as depicted in 
Figure~\ref{fig:multiple-attacks-vo}.

\begin{figure}[htb]

\medskip

\begin{center}
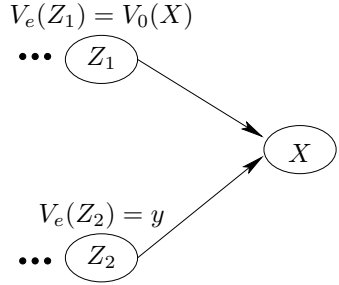
\end{center}

\medskip

\caption{Combination of a node's initial value with its attacks.\label{fig:multiple-attacks-vo-y}}
\end{figure}

We can compute $g(\{1-\f{Y_1},\ldots,1-\f{Y_k}\})=y$, which gives us 
the value of the attack on $X$. The equilibrium value of $X$ is the result 
of combining its initial value $V_0(X)$ with the value of the combined attacks 
on it, so we can pretend we have the interaction depicted in 
Figure~\ref{fig:multiple-attacks-vo-y}.

\noindent and compute $h(V_e(Z_1),V_e(Z_2))$, i.e., $h(V_0(X),
g(\{1-\f{Y_1},\ldots,1-\f{Y_k}\})$.  We get equations of the kind 
\begin{equation}
\f{X}=h(V_0(X),g(\{1-\f{Y_1},\ldots,1-\f{Y_k}\})\label{eq:eq-comb-attac-vo}
\end{equation}
to solve. As we mentioned,  $g$ and $h$ may be different functions, so for
example we could have
$g(\{1-\f{Y_1},\ldots,\allowbreak 1-\f{Y_k}\})=
\min(\{1-\f{Y_1},\ldots,\allowbreak 1-\f{Y_k}\})$ 
and $h(x,y)=x\cdot y$.

\begin{figure}[htb]

\medskip

\begin{center}
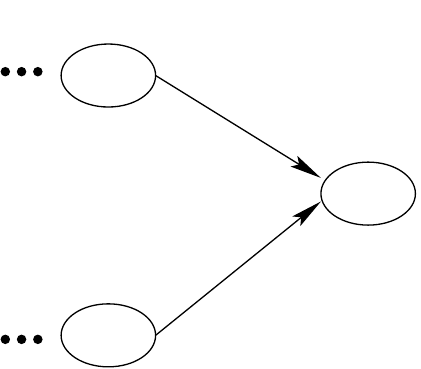
\end{center}

\medskip

\caption{Combining attacks and initial value.\label{fig:multiple-attacks-vo-y-2}}
\end{figure}

When $f$ and $g$ are the same, e.g., $f=g=\min$, we can pretend we
have Figure~\ref{fig:multiple-attacks-vo-y-2}. And then we get
$\f{X}=\min(\{1-(1-V_0(X)),1-\f{Y_1},\ldots,1-\f{Y_k}\})=\min(\{V_0(X),
1-\f{Y_1},\ldots,1-\f{Y_k}\})$. Note that in this situation, the traditional 
equation (without $h$ and initial values) is a special case of 
$\fzero(X)=1$, because $h(1,z) =z$ and then $\f{X}=h(1,g(\{1-\f{Y_1},
\ldots,1-\f{Y_k}\})) =g(\{1-\f{Y_1},\ldots,\allowbreak 1-\f{Y_k}\})$.

We now address another issue. Once we solve equation~(\ref{eq:eq-comb-attac-vo}), 
we get a function \feq\ such that
\[\f{X}=h(V_0(X),g(\{1-Y_1,\ldots,1-Y_k\}))\]
Can we use $\f{X}$ itself as an initial value?

In other words, do we have that equation~(\ref{eq:reuse-vo}) below holds?
\begin{equation}
\f{X}=h(\f{X},g(\{1-Y_1,\ldots,1-Y_k\}))\label{eq:reuse-vo}
\end{equation}

\begin{figure}[htb]

\medskip

\begin{center}
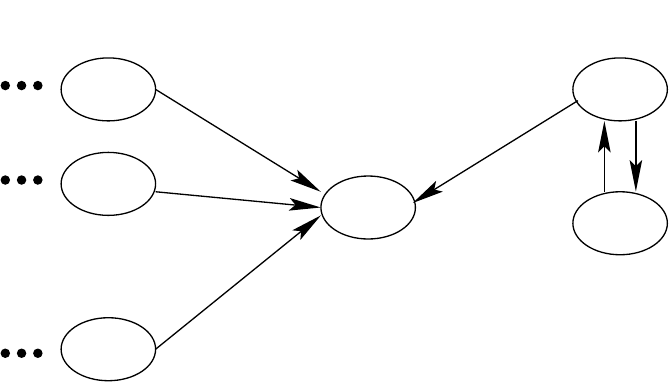
\end{center}

\medskip

\caption{Combining attacks and initial value.\label{fig:new-nodes}}
\end{figure}

The answer is ``no'', because $g$ and $h$ are not necessarily the same
function. In case it is the same function, we have
\begin{eqnarray*}
\f{X} & = & h(\f{X},g(\{1-Y_1,\ldots,1-Y_k\}))\\
      & = & g(\{\f{X},g(\{1-Y_1,\ldots,1-Y_k\})\})\\
      & = & g(\{\f{X},1-Y_1,\ldots,1-Y_k\})\\
      & = & g(\{Z,1-Y_1,\ldots,1-Y_k\})
\end{eqnarray*}
\noindent where $Z$ is the equilibrium value of a new point
attacking $X$, whose value is fixed at $V_0(X)$. We can simulate this
by adding new points $Z_X^1$ and $Z_X^2$ for each $X$ and form the
graph depicited in Figure~\ref{fig:new-nodes}.
All solutions to the cycle $Z^1_X\leftrightarrow Z^2_X$ are of the form
$(\f{Z^1_X}$,$1-\f{Z^1_X})$, which means that $Z^1_X$ can get any value in
$U$ and hence so can its attack on $X$. This can be seen as having the
same effect as giving $X$ a particular initial value in $U$. 


These conditions are satisfied by the 
t-norm $\min$. An attack takes the complement of the value of the attacking 
node to $1$ (co-norm).

We have that
\[\min\limits_{Y\in Att(X)}\{1-\f{Y}\}=1 - \max\limits_{Y\in Att(X)}\{\f{Y}\}\]
giving us our now familiar \eqmax.

The t-norm $\min$ only cares about the strength of the strongest argument. 
In some applications, one could argue that attacks by multiple arguments 
should bear more weight than the value of any of the arguments alone. One 
way of modelling this is by combining attacks via {\em product}.  
\begin{equation}
\label{eq:mode-of-attack}
\displaystyle \prod\limits_{Y \in Att(X)} 
(1-\f{Y})
\end{equation}

Again, if any attacker of an argument has equilibrium value $1$, 
then the value of the product will be $0$. Otherwise, if all attackers of $X$ 
are fully defeated, i.e., if they all have equilibrium value $0$, then the 
value of the product will be $1$. 
Combining the value of attacks in this way was initially proposed in 
\cite{sup-att-net:2005}. 

The expression~(\ref{eq:mode-of-attack}) is equivalent to
\begin{equation}
\displaystyle 1- \displaystyle \curlyvee_{Y \in Att(X)} \f{Y}
\label{eq:lm-eq}
\end{equation}
where $x \curlyvee y=x+y -x.y$ and for $V=\{x_1,\ldots,x_k\}$,
$\curlyvee V=(((x_1 \curlyvee x_2) \curlyvee \ldots) \curlyvee x_k)$.
(\ref{eq:lm-eq}) is the complement of the probabilistic sum t-conorm. 
It is well known that in probability theory, the probabilistic sum expresses 
the probability of
the occurrence of independent events. Since we want to weaken the
value of the attacked node, we take the complement of this sum to $1$.

A network generates a system of equations. If there are cycles 
in the graph, then some of the variables associated with equilibrium values 
will be expressed in terms of each other. We now explore this in a bit
more detail.

Consider the following example.

\begin{figure}[htb]

\medskip

\begin{center}
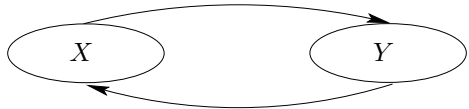
\end{center}

\medskip

\caption{A cycle involving two nodes.\label{fig:two-cycle}}
\end{figure}

Assume that all initial values are $1$, that $g$ and $h$ are product. 
The graph in Figure~\ref{fig:two-cycle} will generate the system of equations
\begin{eqnarray*}
\f{X} & = & 1- \f{Y} \\
\f{Y} & = & 1- \f{X} \\
\end{eqnarray*}
\noindent which has an infinite number of solutions given by the
formula $\f{X}+\f{Y}=1$. A way to arrive at a unique solution to the 
equations is to introduce a constant $\kappa<1$ and analyse the solution 
to the system of equations in the limit $\kappa \rightarrow 1$. This would 
give us

\begin{eqnarray*}
\f{X} & = & \kappa(1- \f{Y}) \\ \f{Y} & = & \kappa(1- \f{X})
\\ \\ \f{X} & = & \kappa - \kappa \f{Y}\\ & = & \kappa -
\kappa(\kappa - \kappa \f{X})\\ & = & \kappa - \kappa^2 + \kappa^2
\f{X} \\ \f{X} - \kappa^2 \f{X} & = & \kappa - \kappa^2 \\ \f{X}(1
- \kappa^2) & = & \kappa - \kappa^2 \\ \f{X} & = & \frac{\kappa(1 -
  \kappa)}{(1-\kappa)(1 + \kappa)} \\ \f{X} & = &
\frac{\kappa}{1+\kappa}
\end{eqnarray*}

Hence, when $\kappa \rightarrow 1$, $\f{X}=\f{Y}=\nicefrac{1}{2}$.
This result explains the implicit introduction of the parameter
$\varepsilon$ to the vote aggregation function proposed by Leite and Martins 
in \cite{leite-martins:2011}.\footnote{We disagree with the reasons for
the introduction of the parameter itself, although technically it is
the reason why the solution converges. A full discussion about this is 
given on Section~\ref{sec:comparisons}.}

\begin{figure}[hbt]

\medskip

\begin{center}
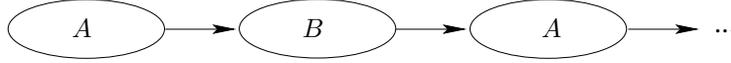
\end{center}

\medskip

\vspace*{-5mm}
\caption{Unravelling the cycle in the network of 
Figure~\ref{fig:two-cycle}.\label{fig:unravelled-cycle}}
\end{figure}

Since the initial values of the two nodes in the network of 
Figure~\ref{fig:two-cycle} are the same, another way of looking at the 
network is by unravelling the cycle starting arbitrarily at one of its nodes, 
say $X$. In our example, this would result in the (infinite) network of 
Figure~\ref{fig:unravelled-cycle}.

If we assume the initial values for $X$ and $Y$ are both $x$, the 
equilibrium value for $X$ could be calculated as

\[ \f{X}=x\cdot (1-(x\cdot (1-(x\cdot ( 1- \ldots))))\]

Now suppose $x=\frac{1}{1+\varepsilon}$, for some $\varepsilon>0$, we have 
that 
\[\f{X}=\frac{1}{1+\varepsilon}\left(1 - \left(\frac{1}{1+\varepsilon}\left(1 - 
\left(\frac{1}{1+\varepsilon}\left(1 - \ldots\right)\right)\right)\right)\right)\]
Thus, in fact, we would be multiplying the initial value
$x=\frac{1}{1+\varepsilon}$ by the number
\[\delta=1 - \left(\frac{1}{1+\varepsilon}\left(1 - \left(\frac{1}{1+\varepsilon}\left(1 - \ldots\right)\right)\right)\right)\]

Let us calculate what the value $\delta$ is. To simplify the 
calculation we set $\alpha=(1+\varepsilon)$, we then get
\[\delta=1 - \left(\frac{1}{\alpha}\left(1 - \left(\frac{1}{\alpha}\left(1 - \ldots\right)\right)\right)\right)\]
If we expand the first multiplication, we get
\begin{eqnarray*}
\displaystyle
\delta&  = & 1 - \left(\frac{1}{\alpha} - \frac{1}{\alpha^2}\left(1- \frac{1}{\alpha}\left(\ldots\right)\right)\right)\\
& = & 1 - \left[\frac{1}{\alpha} - \frac{1}{\alpha^2} + \frac{1}{\alpha^3}\left(1-\frac{1}{\alpha}\left(\ldots\right)\right)\right]\\
& = & 1 - \left[\frac{1}{\alpha} - \frac{1}{\alpha^2} + \frac{1}{\alpha^3} - \frac{1}{\alpha^4}\left(1-\frac{1}{\alpha}(\ldots)\right)\right]\\
& = & 1 -  \left[\left(\frac{\alpha - 1}{\alpha^2}\right) + \left(\frac{\alpha - 1}{\alpha^4}\right) +
\left(\frac{\alpha -1}{\alpha^6}\right) + \ldots \right]
\end{eqnarray*}
The component
\[\left(\frac{\alpha - 1}{\alpha^2}\right) + \left(\frac{\alpha - 1}{\alpha^4}\right) +
\left(\frac{\alpha -1}{\alpha^6}\right) + \ldots\]
can be re-written as
\[\sum\limits_{k=1}^{\infty} \left(\alpha-1\right)\left(\frac{1}{\alpha^2}\right)^k\] 
which is the same as
\[\sum\limits_{k=0}^{\infty} (\alpha-1)\left(\frac{1}{\alpha^2}\right)^k - (\alpha -1)\]

The first component in the main subtraction above is the sum of a geometric
series with common ratio $\frac{1}{\alpha^2}$ and scale factor
$\alpha-1$. Now note that the ratio $\frac{1}{\alpha^2} < 1$, since
$\alpha = 1+\varepsilon> 1$, and hence
\[\sum\limits_{k=0}^{\infty} (\alpha-1)\left(\frac{1}{\alpha^2}\right)^k =
  \frac{(\alpha - 1)}{1-\frac{1}{\alpha^2}}=
\frac{\alpha^2(\alpha-1)}{\alpha^2 -1}\] 

The subtraction can therefore be re-written as
\begin{eqnarray*}
\displaystyle
&     & \frac{\alpha^2(\alpha-1)}{\alpha^2 -1}-(\alpha - 1)\\
& = & \frac{\alpha^2(\alpha-1)-(\alpha^2-1)(\alpha - 1)}{\alpha^2 -1}\\
& = & \frac{(\alpha-1)(\alpha^2 -\alpha^2+ 1)}{\alpha^2 -1}\\
& = & \frac{\alpha}{\alpha^2-1}
\end{eqnarray*}
Remember that $\alpha=1+\varepsilon$, hence 
\begin{eqnarray*}
\displaystyle
\frac{\alpha}{\alpha^2-1} & = &
\frac{1+\varepsilon-1}{(1+\varepsilon)(1+\varepsilon)-1}\\ & = &
\frac{\varepsilon}{\varepsilon^2 +2\varepsilon +1 -1}\\ & = &
\frac{\varepsilon}{\varepsilon(\varepsilon+2)}\\ & = & \frac{1}{\varepsilon+2}
\end{eqnarray*}
Therefore,
\[ \delta = \left(1- \frac{1}{\varepsilon+2}\right)\]
and hence in the limit $\varepsilon \rightarrow 0$, we get
\[ \f{X}=\lim\limits_{\varepsilon \to 0} \frac{1}{1+\varepsilon}
   \left(1- \frac{1}{\varepsilon+2}\right)=\frac{1}{2}\]
as expected.

\begin{figure}[hbt]

\medskip

\begin{center}
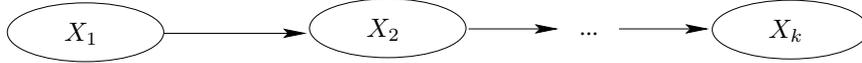
\end{center}

\medskip

\vspace*{-5mm}
\caption{Sequence of attacks.\label{fig:attack-sequence}}
\end{figure}

If we just have an acyclic sequence of attacks such as the one in 
Figure~\ref{fig:attack-sequence}, we can analyse what happens with the 
equilibrium values of each node, given a fixed initial value $v$ for all
nodes (again we consider $f$ as product).

From the network in Figure~\ref{fig:attack-sequence}, we get that
$\f{X_1}=v$, $\f{X_2}=v\cdot (1-v)$, $\f{X_3}=v\cdot (1-(v \cdot
(1-v)))$, and so forth. If $v=1$, then $\f{X_1}=1$, $\f{X_2}=0$,
$\f{X_3}=1$,\ldots.  The values alternate between $0$ and $1$,
agreeing with Dung's original semantics as expected. If $v=0$, then
$\f{X_i}=0$ for all $0 \leq i\leq k$. This is a consequence of the
fact, that by using $g$, the equilibrium value depends on the node's 
initial value and if this is $0$, so is the equilibrium value of the
node when $g$ is product. Similarly, if the initial values of all nodes
is $\frac{1}{2}$, we get $\f{X_1}=\frac{1}{2}$, $\f{X_2}=\frac{1}{4}$,
$\f{X_3}=\frac{3}{8}$, \ldots.

Contrast the calculation of the equilibrium values above with that of
Besnard and Hunter \cite{besnard-hunter:2001}, in which the values are 
calculated by a so-called {\em categoriser} function. In their paper, the
given example of such a function was the {\boldmath$h$}{\bf-categoriser} $h$, 
defined as 
\[
h(X) = \left\{ 
\begin{tabular}{ll} $1$, & if $Att(X)=\varnothing$ \\ 
$\frac{1}{1+\sum\limits_{Y\in Att(X)}h(Y)}$, & otherwise
\end{tabular}
\right. 
\]

Assuming initial value $v=1$ in the example above, we would have that 
$h(X_1)=1$, $h(X_2)=\frac{1}{2}$, $h(X_3)=\frac{2}{3}$, and so forth.
This obviously does not agree with Dung's interpretation. 

\begin{figure}[hbt]

\medskip

\begin{center}
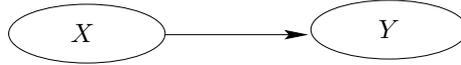
\end{center}

\medskip

\vspace*{-5mm}
\caption{Attack by a node of same initial value.\label{fig:attacked-by-same}}
\end{figure}

The effect on the equilibrium value of a node calculated using $g$  and $h$
as product, when the node is attacked by a single node of same initial value 
is now discussed. This is the scenario depicted in Figure~\ref{fig:attacked-by-same}.

If we assume that $X$ and $Y$ get initial value $x$, we have that
since $X$ has no attacking arguments, $\f{X}=x\cdot (1-0)=x$.  
We then have

\begin{tabular}{l@{\hspace{0.15cm}}l@{\hspace{0.15cm}}l@{\hspace{0.15cm}}l@{\hspace{0.15cm}}l}
$\f{X}$ & $=$ &  $x$  &\\
$\f{Y}$ & $=$ &  $x(1-\f{X})$ & $=$ &  $x-x^2$\\
\end{tabular}

\medskip

\begin{figure}[hbt]
\begin{center}
\includegraphics[width=11cm]{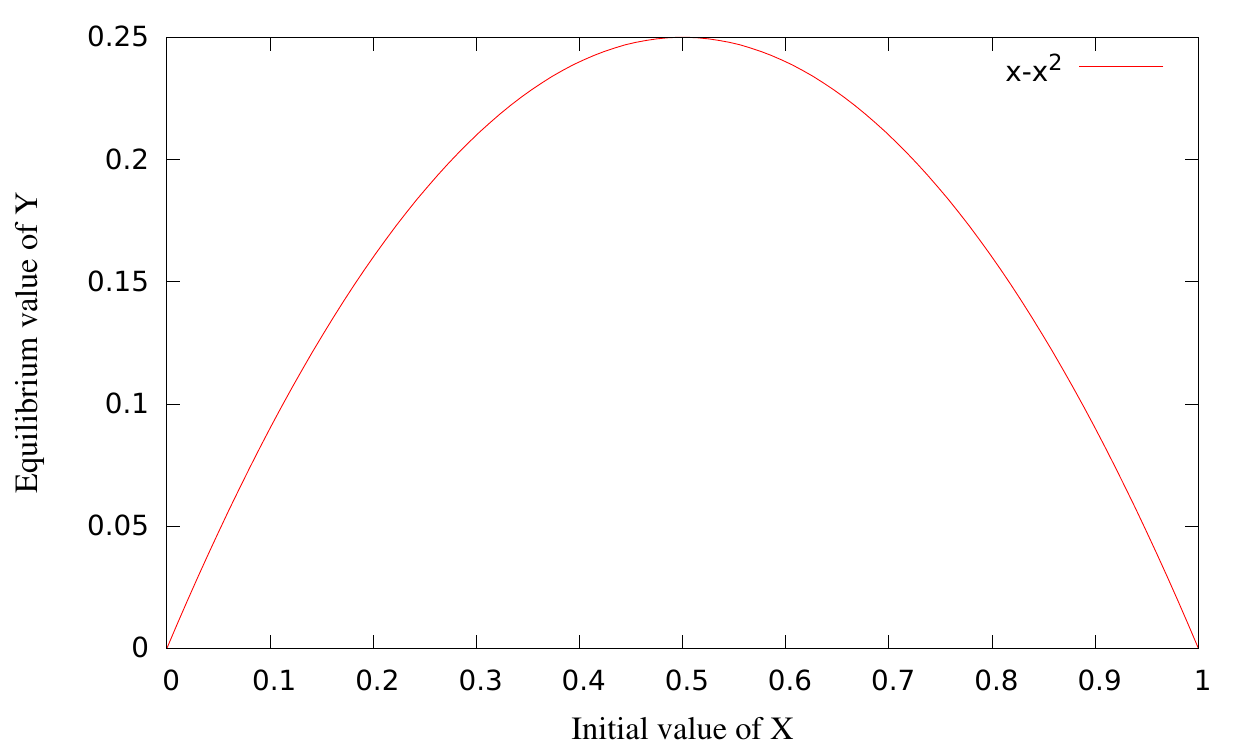}
\end{center}
\caption{Attack by a single node of same initial value.
\label{fig:equiv-attacks-all}}
\end{figure}

If $X$ gets initial value $1$, then it gets equilibrium value $1$
and since it attacks $Y$, its equilibrium value is $0$, as 
expected.\footnote{This equilibrium value would be $0$ independently
of the initial value of $Y$ in this case, because we retain Dung's 
semantics in the trivial cases.}  On the other hand, if $X$ and $Y$ get
initial value $0$, then $Y$'s equilibrium value will also be $0$. If 
$X$ and $Y$ get initial value $\frac{1}{2}$, then the attack by $X$ 
on $Y$ is not sufficiently strong to annihilate $Y$'s initial value
completely. In fact, it only brings it down by $50\%$, i.e., giving
it equilibrium value $\frac{1}{4}$. This is the maximum weakening 
that an attack by an equally strong argument can inflict on $Y$
using product. The full range of values under these circumstances
is illustrated by Figure~\ref{fig:equiv-attacks-all}.

\subsection{Comparisons with Social Abstract Argumentation Networks}

In \cite{leite-martins:2011}, Leite and Martins proposed {\em social
  abstract argumentation frameworks} which can be seen as an extension
of Dung's {\em abstract argumentation frameworks} to allow the
representation of information about votes to arguments. This work was
subsequently extended in \cite{leite-martins-to-appear:2014} to handle 
votes on attacks too.

The motivation
for these networks is to provide a means to calculate the result of the
interaction between arguments using approval and disapproval ratings
from users of news forums. The idea is that when a user sees an
argument, she may approve it, disapprove it, or simply abstain from
expressing an opinion. Since the arguments relate to each other through 
an attack relation (not necessarily known to the users), the votes 
themselves are not sufficient to provide an
overall picture of the discussion.  An interesting feature of these
environments is therefore their intrinsic informal nature in the sense
that in practice it is possible that voters vote for multiple arguments
in the debate and also that users may be unware of
conflicts between the arguments.

One immediate concern is the provision of an appropriate semantics 
which can give an interpretation to the votes capturing the intuition of 
the voting process. The semantics must take into account both the 
interactions between the arguments as well as the votes originally 
cast for them.

We now introduce E{\v{g}}ilmez \etal's work \cite{leite-martins-to-appear:2014},
which is an extension to \cite{leite-martins:2011} so we can compare it with our 
methodology.\footnote{Note that \cite{leite-martins:2011}
  were not aware (and did not quote) \cite{sup-att-net:2005}, which
  was six years earlier. Thus, the only new contribution in \cite{sup-att-net:2005}
  was how they determine the initial values and the connection with
  voting.}

\begin{definition} \cite{leite-martins-to-appear:2014}
A social abstract argumentation framework is a tuple \linebreak $\tuple{\c{S},
\c{R}, V_S, V_R}$, where \c{S} is a set of arguments; $\c{R} : \c{S}
\times \c{S}$ is a binary attack relation between arguments; and
$V_S:\c{S} \longrightarrow {\mathbb N} \times {\mathbb N}$ and 
$V_R:\c{R} \longrightarrow {\mathbb N} \times {\mathbb N}$ are functions
mapping arguments and attacks to tuples $\langle v^+,v^-\rangle$ representing
the total of approval and disapproval votes received by each.
\end{definition}

In order to provide a semantical interpretation, E{\v{g}}ilmez \etal\  
introduce the concept of a {\em semantic framework} presented below.

\begin{definition}\cite{leite-martins-to-appear:2014} A {\em social abstract 
argumentation semantic framework} is a tuple $\langle L, \tau,
  \curlywedge, \curlyvee, \neg \rangle$, where
\begin{itemize}
\item $L$ is a totally ordered set with top and bottom elements $\top$
  and $\bot$, respectively
\item $\tau:\mathbb{N} \times \mathbb{N} \longrightarrow L$ is a vote
  aggregation function that computes the {\em social support} of
  arguments and attacks
\item $\curlywedge_S, \curlywedge_R: L \times L \longrightarrow L$; $\curlyvee: L
  \times L \longrightarrow L$; and $\neg: L \longrightarrow L$ are
  algebraic operations on $L$
\end{itemize}
\end{definition}

The operations $\tau$, $\curlywedge$, $\curlyvee$ and $\neg$ are used
to calculate the overall strength of the arguments and attacks based on their
initial votes. For the voting scenario considered in \cite{leite-martins-to-appear:2014}, 
the so-called {\em product semantics} was given. In this semantics, $L$ is $U$ 
(i.e., the interval $[0,1]$); $\curlywedge_S$ and $\curlywedge_R$ are both
the {\em product t-norm} $\curlywedge$, where $x \curlywedge y = x . y$; 
$\curlyvee$ is its associated t-conorm, i.e., $x \curlyvee y = 1-(1-x).(1-y)=
x+y - x.y$; $\neg x = 1- x$; and $\tau$ is one of a family of operations 
$\tau_\varepsilon$ defined as follows:

\begin{definition}\label{simple-ag}[Initial support for attacks and arguments] 
Let $X$ be an argument and $V_S(X)=\tuple{p,m}$.
\[ \tau_{\varepsilon}(X)= \frac{p}{p+m+\varepsilon} \]
where $\varepsilon > 0$.

The initial support value for an attack $(X,Y)$ is calculated identically,
except that we use $V_R\big((X,Y)\big)$ instead of $V_S(X)$.
\end{definition}

One can regard $\tau_\varepsilon$ and the operation that calculates the 
initial social support value for arguments and attacks.  However, one adverse 
effect of calculating the initial support in this way is that it fails to put the 
votes in context, so an argument for which a single supporting vote is cast can get 
social support close to $1$ (depending on what the value of $\epsilon$ is).\footnote{$\varepsilon$ 
cannot be $0$, because this would render $\tau_\varepsilon$ ill defined for components 
with no votes.}

The semantics of a social abstract framework is then defined by a {\em
  social model} presented below.

\begin{definition}\cite{leite-martins-to-appear:2014} Let $F$ be a social abstract 
argumentation framework and $\c{T}=\langle L,\tau,\curlywedge_S, \curlywedge_R, 
\curlyvee, \neg\rangle$ a semantic framework. A social model of $F$ under semantics 
\c{T} is a total mapping $M: \c{S} \longrightarrow L$ such that for every 
$X \in \c{S}$ \[M(X) = \tau(X) \curlywedge \neg \curlyvee_{Y_i \in Att(X)} \{\tau\big((Y_i,X)\big) 
\curlywedge M(Y_i)\}\]
\end{definition}

Note that if $\curlywedge$ is product t-norm and $\curlyvee$ is its t-conorm, 
as in \cite{leite-martins-to-appear:2014}, then
\begin{eqnarray*}
M(X) & = & \tau(X) \curlywedge \neg \curlyvee_{Y_i \in Att(X)} \{\tau\left((Y_i,X)\right) \curlywedge M(Y_i)\}\\
     & = & \tau(X) \cdot \left(1 - \left(1 - \prod_{Y_i \in Att(X)} \left(1 - \tau((Y_i,X)) \cdot M(Y_i)\right) \right)\right)\\
     & = & \tau(X) \cdot \prod_{Y_i \in Att(X)} \left(1 - \tau((Y_i,X)) \cdot M(Y_i)\right)\\
\end{eqnarray*}
Contrast $M(X)$ with the equilibrium value of $X$, $V_e(X)$ as we proposed it in 
\cite[Definition~5]{gabbay-rodrigues:12}:
\[V_e(X)=V_0(X) \cdot \prod_{Y_i \in Att(X)}\left( 1 - \xi\left((Y_i,X)\right)V_e(Y_i)\right)\]

The calculation is exactly the same, except that we compute initial support
differently as discussed next. We emphasise that the notion of the strength of 
attack already existed since \cite{sup-att-net:2005}.

As Leite \etal\  initially pointed out in \cite{leite-martins:2011}, there 
are difficulties with the vote aggregation function $\tau$. At first, the 
constant $\varepsilon$ was introduced to avoid the existence of infinite
models. For example, consider the network

\medskip

\begin{center}
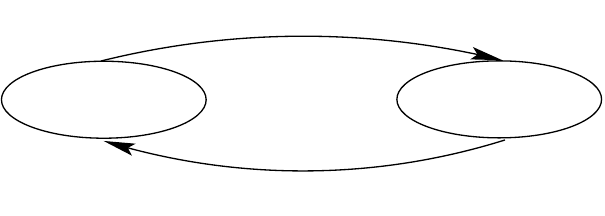
\end{center}

\medskip

And assume that $V_S(X)=V_S(Y)=\tuple{x,0}$. Then we have that
$\tau_0(X)=\tau_0(Y)=1$ and hence any model $M$ satisfying the
equation $M(X)=1-M(Y)$ is a social model of the network.

However, if the social support uses a very small value for $\varepsilon$ 
that is nevertheless greater than $0$, we get the following situation.

\begin{eqnarray*}
M(X) & = & \frac{1}{1+\varepsilon}(1 -M(Y)) \\
M(Y) & = & \frac{1}{1+\varepsilon}(1 -M(X)) \\
\end{eqnarray*}

If we substitute one value for the other, we get that
\begin{eqnarray*}
M(X) & = & \frac{1}{1+\varepsilon}\left(1  - \frac{1}{1+\varepsilon}(1 -M(X))
\right)\\
& = & \frac{1}{1+\varepsilon}\left( \frac{1 + \varepsilon - 1 + M(X)}{1 + 
\varepsilon} \right) \\
& = & \frac{1}{1+\varepsilon}\left( \frac{\varepsilon+ M(X)}{1 + \varepsilon} 
\right) \\
& = & \frac{\varepsilon+ M(X)}{(1 + \varepsilon)^2} \\
M(X) (1 + \varepsilon)^2 & = & \varepsilon + M(X) \\
M(X)(1 + \varepsilon)^2 - M(X) & = & \varepsilon \\
M(X) & = & \frac{\varepsilon}{(1+\varepsilon)^2 -1} \\
     & = & \frac{\varepsilon}{2\varepsilon + \varepsilon^2} \\
     & = & \frac{1}{2 + \varepsilon}
\end{eqnarray*}

\noindent and hence $\lim_{\varepsilon  \to 0} M(X) = \frac{1}{2}=M(Y)$, which provides
a unique solution.

In our opinion, there is a methodological problem and a {\em technical} one.
The value $\varepsilon > 0$ solves the technical problem, which is the convergence
to a single model. However, methodologically speaking, the objective of $\tau$ is to
calculate initial support for components and in that respect, the constant $\varepsilon$
has no part to play. This situation does not arise in \cite{gabbay-rodrigues:12,%
gabbay-rodrigues-jlc:13}, because the social support function there is normalised with 
respect to the total number of argumentation networks being merged. We hope we have 
shed some light into the technicalities of finding solutions to the equations 
throughout this paper.


A more difficult problem is the exaggerated role played by terminal arguments with 
little support, as shown below. Consider the following example:

\medskip

\begin{center}
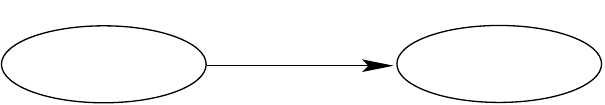
\end{center}

\medskip

\noindent and assume that $V_S(X)=\tuple{1,0}$ and $V_S(Y)=\tuple{99,0}$. 
According to Definition~\ref{simple-ag}, $\tau_0(X) =1$. Since $X$ is
a terminal argument, $M(X)=1(1-0)=1$ and hence $M(Y)=\tau_0(Y)(1-
\tau\left((X,Y)\right)\cdot M(X))=\tau_0(Y)\left(1-\tau\left((X,Y)\right)\right)$. 
Hence, the fate of $Y$ depends on how strongly the attack from $X$ is 
supported.\footnote{The main motivation for the introduction of the
weights on attacks in \cite{leite-martins-to-appear:2014}.} 
Although this technically solves the problem, it mixes the
two issues, because a voter must vote for an argument as well as for
its attacks, if they are to have any effect and an argument can get
very high initial support even if it is voted only by a very small number
of voters.\footnote{High values of $\tau$ should correspond to
high level of initial support.}

\end{document}